\documentclass[lettersize, journal]{IEEEtran}

\usepackage{amsmath}
\usepackage{amsfonts}
\usepackage{multirow}
\usepackage{lineno,hyperref}
\usepackage{amssymb}

\usepackage{mathrsfs}
\usepackage{graphicx}
\usepackage{epstopdf}
\usepackage{rotating}
\usepackage{subfigure}
\usepackage{color}
\usepackage{algorithm}
\usepackage{mathrsfs}
\usepackage{amsthm}
\usepackage{cite}
\usepackage{array}

\usepackage{algpseudocode} 

\newtheorem{define}{Definition}

\graphicspath{{figs/}}

\title{Blind Deconvolution for Color Images Using Normalized Quaternion Kernels}

\author{Yuming Yang\thanks{Yuming Yang is with the School of Mathematical Sciences, Key Laboratory of Intelligent Computing and Applications (Ministry of Education),
Tongji University, Shanghai 200092, China (E-mail: 2111162@tongji.edu.cn).},
	Michael K. Ng\thanks{Michael K. Ng is with the Department of Mathematics, Hong Kong Baptist
		University, Hong Kong (E-mail: michael-ng@hkbu.edu.hk).
		M. Ng is supported by the GDSTC: Guangdong and Hong Kong Universities ``1+1+1'' Joint Research Collaboration Scheme UICR0800008-24, National Key Research and Development Program of China under Grant 2024YFE0202900, RGC GRF 12300125 and Joint NSFC and RGC N-HKU769/21.},
		Zhigang Jia\thanks{Zhigang Jia is with the School of Mathematics and Statistics \& RIIS, Jiangsu Normal University, Xuzhou 221116, China (E-mail: zhgjia@jsnu.edu.cn).
		Z. Jia is supported by the National Key R\&D program of China
		(No. 2023YFA1010101);
		the National Natural Science Foundation of China under grants 12171210, 12090011, 61971234 and 11771188;
		the ``QingLan'' Project for Colleges and Universities of Jiangsu Province (Young and middleaged academic leaders);
		the Major Projects of Universities in Jiangsu Province (No. 21KJA110001);
		the Natural Science Foundation of Fujian Province of China grants 2022J01378.},
		Wei Wang\thanks{Wei Wang is with the School of Mathematical Sciences, Key Laboratory of Intelligent Computing and Applications (Ministry of Education), Tongji University, Shanghai 200092, China (E-mail: wangw@tongji.edu.cn). 
		W. Wang is supported by Natural Science Foundation of Shanghai (22ZR1465300).}
		\thanks{(Corresponding author: Wei Wang.)}}

\begin{document}
	\maketitle
	
	\begin{abstract}
		In this work, we address the challenging problem of blind deconvolution for color images.
		Existing methods often convert color images to grayscale or process each color channel separately, which overlooking the relationships between color channels.
		To handle this issue, we formulate a novel quaternion fidelity term designed specifically for color image blind deconvolution.
		This fidelity term leverages the properties of quaternion convolution kernel, which consists of four kernels:
		one that functions similarly to a non-negative convolution kernel to capture the overall blur, and three additional convolution kernels without constraints corresponding to red, green and blue channels respectively model their unknown interdependencies. 
		In order to preserve image intensity, we propose to use the normalized quaternion kernel in the blind deconvolution process. 
		Extensive experiments on real datasets of blurred color images show that the proposed method effectively removes artifacts and significantly improves deblurring effect, demonstrating its potential as a powerful tool for color image deconvolution.

	\end{abstract}
	

	\thispagestyle{plain}
	\markboth{}{Q}
	
	\section{Introduction}
	Blind image deblurring is a fundamental and challenging task in image processing and computer vision\cite{1974An}. It aims to recover a sharp latent image and the corresponding blur kernel from a single blurred image without any prior knowledge of the blur process. This problem is of great importance as it serves as a preprocessing step for various high-level vision tasks such as object detection, recognition, and scene understanding. Due to the imperfections in image acquisition, such as camera shake, motion during exposure, or defocusing, images can be significantly blurred, which degrades the quality of the captured images and affects the performance of subsequent vision tasks.
	
	For a space-invariant blurring problem, the blurring model is usually established mathematically as
	\begin{equation}
		f = k \star u +n
		\label{equ1}
	\end{equation}
	where $\star$ is the convolution operator, $f, u, k,$ and $n$ denote the blurred image, sharp latent image, blur kernel, and noise, respectively. 
	The blind blurring requires the simultaneous recovery of the sharp latent image $u$ and the blur kernel $k$, given only the blurred image $f$. 
	Since multiple pairs of $u$ and $k$ can yield the same blurred image $f$, this problem is a classic example of an ill-posed problem.
	To address the ill-posed nature of blind deblurring, it is essential to incorporate appropriate prior information, thereby regularizing the problem to render it well-posed. 
	
	Within the Bayesian statistical framework, the Maximum A Posteriori (MAP) estimation has emerged as a conventional technique for estimating the latent image and blur kernel. 
	This approach effectively addresses the ill-posed nature of the blind deblurring problem, thereby yielding more stable and reasonable deblurring results.
	Chan and wong \cite{1998TVDB} proposed first modern and influential blind deblurring method, which is regularized by the total variation (TV) image prior.
	After years of intensive research \cite{FMD2009,2011Levin,L01,Panl0,REVISEMAP}, state-of-the-art MAP-based approaches have reached a consensus regarding image regularization for blind deblurring: the $L_0$ + X paradigm are widely advocated.
	Within this paradigm, the image gradient is regularized via the $L_0$ norm to aggressively purse salient edges, whereas X denotes a flexible portfolio of image priors, such as the dark channel\cite{DCP2018}, local features\cite{2019Chen,2022LIU,2024LGP}, surface-aware\cite{SA2021}, superpixel segmentation prior\cite{2022Luo}.

	It is worth noting that Fergus et al.\cite{2006Fergus} point out that a direct application of the MAP-based approaches may fail on blind deblurring problems. 
	Levin et al.\cite{2011Levin} further showed that any algorithm trying to minimize the same energy failed because the desired solution had a higher energy than the blur-free solution.
	Krishnan and Fergus \cite{2011Krishnan} proposed $L_1$/$L_2$ regularization giving the lowest cost for the sharp image. 
	Rameshan et al.\cite{2012Rameshan} analyzed that the failure of the MAP estimation in Levin et al \cite{2011Levin} is due to the assumption of the uniform prior for the blur kernel. 
	Perrone et al.\cite{TVBD2014} explained and proved that delayed scaling (normalization) of the blur kernel in iterative steps is essential for algorithm convergence, ensuring the MAP model achieves desirable results.
	Marti et al.\cite{2021MAP} revisited the original criticism \cite{2011Levin} of the MAP approach for blind image deblurring. They show through conceptual analysis and experiments that this criticism is unwarranted, and the MAP approach can prefer sharp images over blurry ones. All the above mentioned methods use the mathematical model of (\ref{equ1}) directly to establish the fidelity term, e.g., $\|k \star u - f\|_2^2$, which is reasonable and natural for grayscale images.
	For color images, the most common operation is to first convert the RGB (red, green and blue) input to grayscale version, then estimate the blur kernel by using only the grayscale version and apply the derived kernel uniformly to RGB channels. Another one is to consider RGB channels separately for color image deblurring.
	However, these approaches are not ideal for color images without considering the relationships among color channels.

	Color description relies on color spaces, which meet the need for color language and assign specific values to colors. 
	In the late 1920s, Wright\cite{wright} and Guild\cite{guild} independently conducted a series of experiments on human sight and led to the CIE RGB color space, quantifying all visible colors using red, green, and blue primaries. 
	Tristimulus values, the measures of primary stimuli causing a color sensation on the retina, are uniquely determined when matching a color to a color system. 
	It was shown in \cite{wright,guild} that there is a relationship between tristimulus values and wavelengths in the RGB color space across the visible spectrum.
	In particular, the tristimulus values referring to red, green and blue colors appear in the whole visual spectrum.
	These results indicate that it may not be sufficient to use the same blur kernel for the red, green and blue channels. 
	On the other hand, Košík et al. \cite{Kosik2025} captured the blurred image of a single white pixel under fully controlled conditions and separated the RGB channels, observing that the three channel-wise blurs are highly similar yet subtly distinct.
	Hence, the blur across the RGB channels is inherently interactive and we aspire to devise a principled representation that faithfully captures and exploits this interaction.

	Quaternion have gained traction in color image processing. In \cite{2025CSTV}, jia et al. derived a quaternion convolution operator $\mathbf{Q}$ when studying cross-channel deblurring problem.
	They regard $\mathbf{Q}$ as the splitting factor of $\mathbf{\tilde{K}}$ (the convolution kernel for the cross-channel deblurring problem, composed of $3\times 3$ ordinary convolution kernels) to enhance the data fidelity term.
	We believe that this blur operator $\mathbf{Q}$ can effectively capture the interactions between color channels, which is suitable for applications in color image deblurring.
	Per the definition of quaternions, the quaternion convolution operator $\mathbf{Q}$ comprises four convolution operators. 
	One describes the consistent blur across the three channels, which support the assumption that the RGB blurs should be intrinsically linked. 
	The other three capture the interactions between the channels. 
	This makes the quaternion convolution kernel a natural and logical choice for such operations.
	
	The main contributions of this work are summarized as follows.
	First, we formulate a novel quaternion data fidelity term using the quaternion convolution kernel, applied to color image blind deconvolution. 
	Second, we propose to use the normalized quaternion kernel in the blind deconvolution process in order to preserve image intensity. 
	Third, we develop an efficient model that integrates image gradient sparsity with quaternion convolution kernel. 
	We emphasize that the proposed quaternion convolution kernel can efficiently combine or generalize other methods with various priors to improve the deblurring effect.	
	Extensive experiments on blurred color images show that the proposed model effectively removes artifacts and significantly improves the visual quality of the restored results, demonstrating its potential as a powerful tool for color image deconvolution.

	This paper is organized as follows. In Section II, we recall preliminary information about {the current MAP-based blind deconvolution methods} and quaternions. 
	In Section III, we propose a novel normalized quaternion convolution kernel and the corresponding quaternion deconvolution model. 
	In Section IV, we present an effective algorithm to solve the proposed model. 
	In Section V, we show numerical examples to illustrate the superiority of the proposed quaternion deconvolution model.
	In Section VI, we give the concluding remarks.
	
	\section{Preliminaries}
	\subsection{$L_0$ + X paradigm}
	
	In recent years, MAP-based approaches structured around the $L_0$ + X paradigm have coalesced into the prevailing consensus for blind image deblurring. 
	We provides a concise exposition of the paradigm in this section.
	Xu et. al.\cite{L01} proposed a blind image deconvolution method which used a piecewise function to approximate the $L_0$-norm to model the distribution of image gradients. 
	In \cite{Panl0}, Pan et. al. proposed a simple yet effective $L_0$-regularized prior based on intensity and gradient and achieved high-quality restoration results.
	These works have contributed to the $L_0$ + X paradigm, which is combined with various image priors \cite{DCP2018, 2019Chen, SA2021, 2022LIU, 2024LGP}. 
	Specifically, this paradigm can be represented in the following format, 
	\begin{equation}
		\min_{u,k} \|k\star u - f \|_2^2 + \lambda \|\nabla u\|_0 + \gamma \|k\|_2^2 + \mu \phi(u),
		\label{L0X}
	\end{equation}
	where the first term is the data fidelity term, the second term constrains the $L_0$ norm of the image gradient, the third term is corresponding to the normalization of the blur kernel, and the forth term represents the auxiliary constraint given by image priors. These prior items can usually discriminate sharp and blurred images, leading to the improvement of the deblurring effect.
	In \cite{DCP2018}, Pan et. al. introduced the dark channel prior through an interesting observation that the dark channel of blurred images is less sparse, which can be represented by 
	$$\phi(u) = \| D(u) \|_0,$$
	where $D(u)$ denotes the dark channel of the image $u$.
	In \cite{SA2021}, Liu et. al. proposed surface-aware prior based on the intrinsic geometrical consideration to smooth the artifacts in the intermediate latent image.
	The surface-aware prior is represented as
	$$\phi(u)=\sum_{i=1}^m\sum_{j=1}^n[\sqrt{1+\|D_{i,j}u\|_2^2}+\delta(\|D_{i,j}^xu\|_0+\|D_{i,j}^yu\|_0)],$$ 
	where $\delta$ is a positive parameter.
	
	\subsection{Quaternions}
	
	In this section, we briefly review the basic concepts of quaternions. Let $\mathbb{Q} = \{a_0 + a_1\mathbf{i} + a_2\mathbf{j} + a_3\mathbf{k} \ | \ a_0, a_1, a_2, a_3 \in \mathbb{R}\}$ denote
	the quaternion skew-field, $\mathbb{Q}^n$ represent the set of n-dimensional
	quaternion vectors, and $\mathbb{Q}^{m\times n}$ denote the set of $m\times n$ quaternion
	matrices \cite{quaternion1866}, where three imaginary units $\mathbf{i, j, k}$ satisfy
	
	$$\mathbf{i}^2 = \mathbf{j}^2 = \mathbf{k}^2 = \mathbf{ijk} = -1.$$
	

	{Given a quaternion matrix $$\mathbf{A}=A_0+A_1\mathbf{i}+A_2\mathbf{j}+A_3\mathbf{k}\in\mathbb{Q} ^{m\times n},$$ where $A_0, A_1, A_2, A_3\in \mathbb{R}^{m\times n}.$ }
	The conjugate transpose of quaternion matrix $\mathbf{A}$
	 is defined as 
	 $$\mathbf{A} ^* = A_0^T- A_1^T\mathbf{i} - A_2^T\mathbf{j} - A_3^T\mathbf{k}.$$
	According to \cite{2021Jia}, a homeomorphic mapping $\Re$ will be defined from quaternion matrices, vectors, scalars or operators to structured real matrices or operators,
	$$\begin{gathered}\Re(\mathbf{A})=\left[\begin{array}{rrrr}A_0&-A_1&-A_2&-A_3\\A_1&A_0&-A_3&A_2\\A_2&A_3&A_0&-A_1\\A_3&-A_2&A_1&A_0\end{array}\right].\end{gathered}$$
	The inverse mapping of $\Re$ on the structured real matrices is defined by $\Re ^{- 1}( \Re ( \mathbf{A} ) ) = \mathbf{A}$. 
	Let $\Re _c( \mathbf{A} )$ denote the first column of $\Re(\mathbf{A}).$ The inverse mapping of $\Re_c$ is similarly $\begin{aligned}\text{defined by }\Re_c^{-1}(\Re_c(\mathbf{A}))=\mathbf{A}.\end{aligned}$
	
	We then introduce the measurement of quaternion vectors and matrices. 
	Given a quaternion 
	$$\mathbf{a} = a_0 + a_1\mathbf{i} + a_2\mathbf{j} + a_3\mathbf{k},$$ 
	the modulus of $\mathbf{a}$ is given by 
	$$|\mathbf{a}| = \sqrt{ a_0^2+ a_1^2+ a_2^2+ a_3^2},$$
	The absolute values of a quaternion vector $\mathbf{v} = [\mathbf{v}_i]\in\mathbb{Q}^n$ and a quaternion matrix $\mathbf{A} = [\mathbf{a}_{ij}]\in\mathbb{Q}^{m\times n}$ are
	$$|\mathbf{v}| = [|\mathbf{v}_i|]\in\mathbb{R}^n\ \mathrm{~and~}\ |\mathbf{A}| = [|\mathbf{a}_{ij}|]\in\mathbb{R}^{m\times n}.$$
	The next step is to define the quaternion vector (or matrix) norms, which are functions from quaternion vectors (or matrices) to nonnegative real numbers.
	\begin{define}
		Let $p \geq 1$. The $p$-norm of $\mathbf{v} \in \mathbb{Q}^n$ is 
		$$\|\mathbf{v}\|_p = \left( \sum_{i=1}^n |\mathbf{v}_i|^p \right)^{\frac{1}{p}}.$$ 
		The $p$-norm and $F$-norm of $\mathbf{A} \in \mathbb{Q}^{m \times n}$ are
		$$\|\mathbf{A}\|_p = \max_{\mathbf{x} \in \mathbb{Q}^n \backslash \{0\}} \frac{\|\mathbf{A} \mathbf{x}\|_p}{\|\mathbf{x}\|_p}, \ \ \|\mathbf{A}\|_F = \left( \sum_{i=1}^m \sum_{j=1}^n |\mathbf{a}_{ij}|^2 \right)^{\frac{1}{2}}.$$
	\end{define}
	By Definition 1, we can easily derive that 
	$$\|\mathbf{v}\|_2 = \frac{1}{2} |\Re(\mathbf{v})|_2,\ \ \|\mathbf{A}\|_2 = |\Re(\mathbf{A})|_2$$ 
	and 
	$$\|\mathbf{A}\|_F = |\Re(\mathbf{A})|_F = \frac{1}{2} |\Re(\mathbf{A})|_2.$$
	
	A color image in RGB color space can be represent as a pure imaginary quaternion function
	$$\mathbf{u} = u_1(x,y)\mathbf{i}+u_2(x,y)\mathbf{j}+u_3(x,y)\mathbf{k},$$ 
	where $(x, y)$ denotes the position of a color pixel in a given region $\Omega$ and three real bivariate functions $u_i(x, y)(i = 1, 2, 3)$ denote pixel values of red, green and blue channels, respectively. 
	In the next section, the bold letter $\mathbf{u}$ denotes a quaternion and the ordinary letter $u$ denotes the corresponding matrix form, which has the following correspondence
	$$\tau(\mathbf{u}(x,y)) := u(x,y) = \begin{bmatrix}
		0&u_1(x,y) & u_2(x,y) & u_3(x,y)
	\end{bmatrix}^T,$$
	where $\tau$ is a transformation map.
	
	\section{Quaternion deconvolution model}
	
	In this section, we will introduce the quaternion convolution kernel and present the proposed quaternion deconvolution model based on normalized quaternion convolution kernel.
	
	\subsection{Quaternion convolution operator and the proposed data-fitting term}
	
	Suppose $Q_0,Q_1,Q_2,Q_3,Q_4$ are the common real convolution kernels, embedding them in a quaternion paradigm, then, the quaternion convolution kernel is defined as $\mathbf{Q} = Q_0 + Q_1\mathbf{i} + Q_2\mathbf{j} + Q_3\mathbf{k}$. 
	Based on quaternion multiplication, we can define the following quaternion convolution operator.
	\begin{define}
		A quaternion convolution operator on a color image $\mathbf{u}(x, y)$ is defined by
		\begin{equation*}
			\begin{aligned}
				&\mathbf{Q}\circledast \mathbf{u} \\
			=	& -(Q_{1}\star u_{1}(x,y)+Q_{2}\star u_{2}(x,y)+Q_{3}\star u_{3}(x,y)) \\
				& +(Q_0\star u_1(x,y)-Q_3\star u_2(x,y)+Q_2\star u_3(x,y))\mathbf{i} \\
				& +(Q_3\star u_1(x,y)+Q_0\star u_2(x,y)-Q_1\star u_3(x,y))\mathbf{j} \\
				& +(-Q_{2}\star u_{1}(x,y)+Q_{1}\star u_{2}(x,y)+Q_{0}\star u_{3}(x,y))\mathbf{k},
			\end{aligned}
		\end{equation*}
		where $\star$ represent the convolution operator.
	\end{define}
	By the above definition, we observe that $Q_0$ acts uniformly on $u_1, u_2$, and $u_3$, which is similar to the conventional blur kernel $k$ providing an overall blur effect. 
	In contrast, $Q_1, Q_2$, and $Q_3$ capture the differences between the green and blue, red and blue, red and green channels, respectively. 
	This precisely characterizes the correlations between the RGB channel blur kernels.
	For ease of representation and analysis, we will use the following equivalent matrix representation of $\mathbf{Q}\circledast \mathbf{u},$
	$$\begin{aligned}
		\tau (\mathbf{Q}\circledast \mathbf{u}) &= \Re(\mathbf{Q})\circledast u\\
		 &= \begin{bmatrix}
			Q_0 & -Q_1 & -Q_2 & -Q_3\\
			Q_1 & Q_0 & -Q_3 & Q_2\\
			Q_2 & Q_3 & Q_0 & -Q_1\\
			Q_3 & -Q_2 & Q_1 & Q_0
		\end{bmatrix}\circledast \begin{bmatrix}
			0 \\ u_1\\u_2\\u_3
		\end{bmatrix}\\
	\end{aligned},$$
	which has the equivalent vector form
	$$\begin{aligned}
		T_{\mathbf{Q}}\vec{u} = \begin{bmatrix}
			T_{Q_0} & -T_{Q_1} & -T_{Q_2} & -T_{Q_3}\\
			T_{Q_1} & T_{Q_0} & -T_{Q_3} & T_{Q_2}\\
			T_{Q_2} & T_{Q_3} & T_{Q_0} & -T_{Q_1}\\
			T_{Q_3} & -T_{Q_2} & T_{Q_1} & T_{Q_0}
		\end{bmatrix}\begin{bmatrix}
			0 \\ \vec{u}_1\\\vec{u}_2\\\vec{u}_3
		\end{bmatrix},
	\end{aligned}$$
	where $ T_{Q_i}$ is the matrix form corresponding to the point spread function (PSF) $Q_i$, $T_{\mathbf{Q}}$ denote the whole matrix of $\mathbf{Q}$ and $\vec{u}_i$ denote the vector forms of $u_i$.
	
	Inspired by the above Definition 2, we propose a novel data fitting term $\|\mathbf{Q}\circledast \mathbf{u} - \mathbf{f} \|_2^2$ based on the quaternion convolution operator,
	where $\mathbf{f}$ is the quaternion representation of the input blurred image. The proposed model will be formulated based on this novel data fitting term.

	\subsection{Color image deconvolution based on quaternion kernel}
	We then embed the proposed quaternion data-fitting term into the $L_0$ + X paradigm and present the proposed color image deconvolution model.
	We first consider the baseline case with no additional image prior imposed, i.e., $\phi(u) = 0$. The resulting model is given as follows, 
	\begin{equation}
			\min_{\mathbf{u},\mathbf{Q}} \|\mathbf{Q}\circledast \mathbf{u} - \mathbf{f} \|_2^2 + \lambda \|\nabla u\|_0 + \gamma \|\mathbf{Q}\|_2^2,
			\label{Q_L0}
	\end{equation}
	where $\lambda$ and $\gamma$ are weight parameters.
	In the proposed model, the first term is the quaternion data-fitting term enforcing that the convolution of the recovered image with the quaternion blur kernel closely matches the observed data. The second term inherited from the $L_0$ + X paradigm is corresponding to the $L_0$ norm of the image gradient, effectively promoting prominent edge structures. The third component applies regularization to the blur kernel, mitigating the ill-posedness encountered during the estimation.
	Due to the simplicity and efficiency, the Gaussian prior has been widely used for blurring kernels, which is also known as Tikhonov regularization.
	
	In the subsequent derivations and analyses, we focus exclusively on model (\ref{Q_L0}). We further observe that the image prior $\phi(u)$ is independent of the blur kernel in the $L_0$ + X paradigm. Consequently, the proposed quaternion convolution kernel can be easily extended to $L_0$ + X paradigm equipped with any image prior. Two illustrative instantiations are presented below.
	\begin{itemize}
		\item Quaternion deconvolution model with dark-channel prior \cite{DCP2018}
		\begin{equation}
			\min_{\mathbf{u},\mathbf{Q}} \|\mathbf{Q}\circledast \mathbf{u} - \mathbf{f} \|_2^2 + \lambda \|\nabla u\|_0 + \gamma \|\mathbf{Q}\|_2^2 + \mu \|D(u)\|_0.
			\label{QDCP}
		\end{equation}
		\item Quaternion deconvolution model with surface-aware prior \cite{SA2021}
		\begin{equation}
			\min_{\mathbf{u},\mathbf{Q}} \|\mathbf{Q}\circledast \mathbf{u} - \mathbf{f} \|_2^2 + \lambda \|\nabla u\|_0 + \gamma \|\mathbf{Q}\|_2^2 + \mu \phi(u),
			\label{QSA}
		\end{equation}
		where $\phi(u)$ is {as mentioned in section II-A}.
	\end{itemize}
	We remark here that the above models (\ref{QDCP}) and (\ref{QSA}) can be solved by using similar methods introduced in \cite{DCP2018} and \cite{SA2021}.
		
	\subsection{Normalization of the quaternion convolution kernel}
	
	\begin{figure}[b]
		\centering
		\includegraphics*[width = 0.25\linewidth]{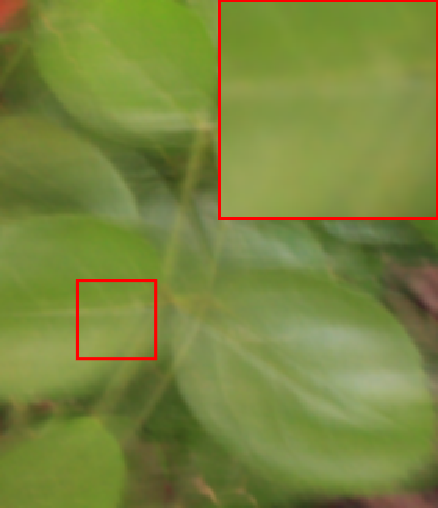}
		\includegraphics*[width = 0.25\linewidth]{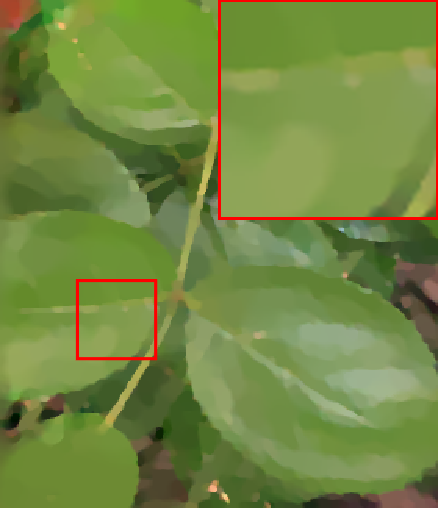}
		\includegraphics*[width = 0.25\linewidth]{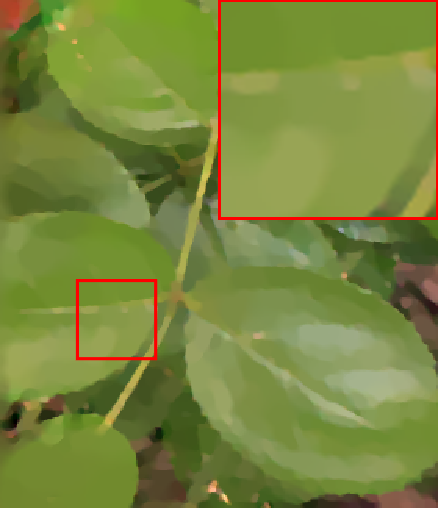}\\\vspace{1mm}
		\includegraphics*[width = 0.25\linewidth]{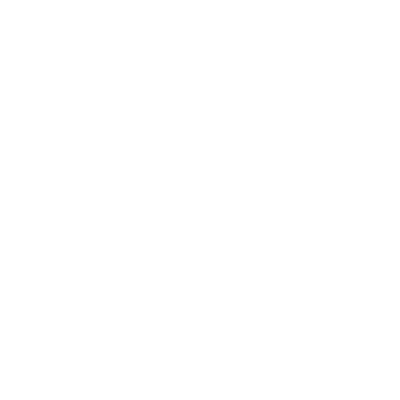}
		\includegraphics*[width = 0.25\linewidth]{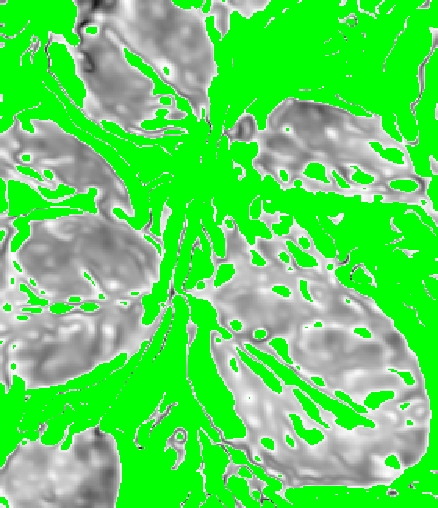}
		\includegraphics*[width = 0.25\linewidth]{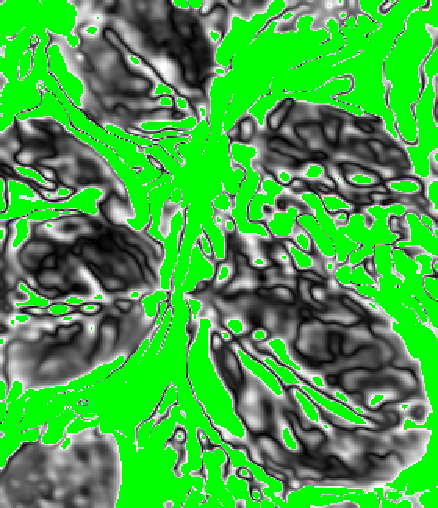}\\
		\hspace{24mm}45322/8.59\hspace{7mm}36352/6.93\\
		\caption{Top row (left to right): blurred image, the restored result by using $\|\mathbf{Q}\|_1 = 1$, and the restored result by using the proposed normalization (upper-right inset: the corresponding zoom-in parts).
		Bottom row: the corresponding spatial distributions of S-CIELAB color error.
		The green regions indicate the pixels with color errors exceeding 5 units.
		The SCIELAB/CIEDE2000 values are given below the spatial distribution images.}
		\label{fig1}
	\end{figure}
	The delayed scaling (normalization) during the blur kernel iteration step is fundamental to the convergence of the algorithm \cite{TVBD2014}, which ensures that the model achieves the desired results.
	For the common real convolution kernel $k$, it is natural and reasonable to use the regularization $k>0$ and $\|k\|_1 = 1$ as constraints.
	This is because the fact that the blur kernel can be viewed as a probability distribution that represents the weight of each pixel's contribution to the surrounding pixels. 
	A fundamental property of a probability distribution is that it sums to 1, which indicates that the weights sum to 1.
	However, $\|\mathbf{Q}\|_1 = \|Q_0\|_1 + \|Q_1\|_1 +\|Q_2\|_1 +\|Q_3\|_1 =1$ is not a natural and reasonable normalization in the quaternion convolution case.
	Since quaternion convolution involves different channels which are not equivalent in the convolution process, which means that the sum of $\mathbf{Q}$ is no longer 1.
		
	Based on the above analysis and observations, we keep the nonnegative condition of $Q_0$ and reuse the equation $\mathbf{Q}\circledast \mathbf{u} = \mathbf{f}$ to give a novel normalization of the quaternion convolution kernel.
	Assume that there exists $\mathbf{t} = [t_0\ t_1\ t_2\ t_3] \in \mathbb{R}^4$ such that
	$(t_0Q_0+t_1Q_1\mathbf{i}+t_2Q_2\mathbf{j}+t_3Q_3\mathbf{k})\circledast \mathbf{u} = \mathbf{f}$, i. e.,
	$$\begin{bmatrix}
	 t_0 Q_0 & -t_3Q_3 & t_2Q_2\\
	 t_3Q_3 & t_0Q_0 & -t_1Q_1\\
	 -t_2Q_2 & t_1Q_1 & t_0Q_0
	\end{bmatrix}
	\circledast \begin{bmatrix}
		 u_1\\u_2\\u_3
	\end{bmatrix} = \begin{bmatrix}
		f_1\\f_2\\f_3
	\end{bmatrix}.$$
	The above equation can be converted into the following equivalent version,
	\begin{equation}
		A\begin{bmatrix}
			t_0& t_1& t_2& t_3
		\end{bmatrix}^T =  \begin{bmatrix} ||f_1||_1& ||f_2||_1& ||f_3||_1\end{bmatrix}^T
		\label{norma}
	\end{equation}
	where 
	 $${\footnotesize  A = \begin{bmatrix}
		||Q_0\star u_1||_1&0&||Q_2\star u_3||_1&-||Q_3\star u_2||_1\\
		||Q_0\star u_2||_1&-||Q_1\star u_3||_1&0&-||Q_3\star u_1||_1\\
		||Q_0\star u_3||_1&||Q_1\star u_2||_1&-||Q_2\star u_1||_1&0
	\end{bmatrix}}.$$
	We use the Moore-Penrose generalized inverse to directly solve the above system of equations in practice.
	Therefore, we derive a principled normalization scheme for the quaternion convolution kernel,
	\begin{equation}
		\tilde{\mathbf{Q}} = t_0Q_0 +t_1Q_1\mathbf{i}+t_2Q_2\mathbf{j}+t_3Q_3\mathbf{k},
		\label{norma2}
	\end{equation}
	where $\tilde{\mathbf{Q}} $ denotes the normalized quaternion convolution kernel.
		
	In Figure \ref{fig1}, we give an example to demonstrate the efficacy of the proposed normalization scheme in correcting hue shifts. We observe that the restored result shows severe color deviations by using the constraint $\|\mathbf{Q}\|_1 = 1$. However, the restored result has very good hue preserving effect by using the proposed normalization scheme.		
	We quantify the color deviations between the restored result and the blurred image by computing the S-CIELAB and CIEDE2000 metrics, which are reported beneath each figure (SCIELAB / CIEDE2000).	
	Noting that S-CIELAB color metric \cite{scie} includes a spatial processing step and is
	useful and efficient for measuring color reproduction errors of digital images,
	and CIEDE2000 color-difference formula \cite{CIEde} accurately models human visual sensitivity to chromatic discrepancies and is widely adopted for quantifying perceptual dissimilarity between two color. Smaller SCIELAB and CIEDE2000 values indicate more accurate color representation and smaller perceptual color difference.
	The restored results are compared with the blurred input, and the quantitative results in SCIELAB and CIEDE2000 metircs confirm that the proposed normalization scheme effectively mitigates hue deviation.
	In addition, we display the spatial distributions of the S-CIELAB color error in Figure \ref{fig1}, which is consistent with the quantitative results. 
	Further experiments on normalization are provided in Section V-A.
	
	\section{Numerical algorithm}
	
	Recall that the proposed model with normalized quaternion convolution kernel are given as follows,
	\begin{equation}\label{mainQ}
		\begin{aligned}
			&\min_{\mathbf{t}, \mathbf{u},\mathbf{Q}} \|\mathbf{Q}\circledast \mathbf{u} - \mathbf{f} \|_2^2 + \lambda \|\nabla u\|_0 + \gamma \|\mathbf{Q}\|_2^2,\\ 
			&{\mathrm{s. t.}}  (t_0Q_0+t_1Q_1\mathbf{i}+t_2Q_2\mathbf{j}+t_3Q_3\mathbf{k})\circledast \mathbf{u} = \mathbf{f}
		\end{aligned}
	\end{equation}	
	We solve model (\ref{mainQ}) by using alternating scheme, i.e., solving one variable by fixing the
	others in the corresponding minimization problem, which is widely used in deconvolution problem \cite{Panl0, DCP2018, SA2021}.
		
	Specifically, for fixed $\mathbf{Q}$, the latent result $\mathbf{u}$ will be solved by using
	\begin{equation}\label{sub_u}
		\min_{\mathbf{u}}\|\mathbf{Q}\circledast \mathbf{u} - \mathbf{f} \|_2^2 + \lambda \|\nabla u\|_0.
	\end{equation}
	For fixed $\mathbf{u}$, the quaternion kernel $\mathbf{Q}$ will be solved by using 
	\begin{equation}\label{sub_Q}
		\min_{\mathbf{Q}}\|\mathbf{Q}\circledast \mathbf{u} - \mathbf{f} \|_2^2 + \gamma \|\mathbf{Q}\|_2^2.
	\end{equation}
	The updating of $\mathbf{t}$ is then given by solving equation (\ref{norma}). 
		
	We summarize the proposed alternating scheme to solve (\ref{mainQ}) in the following Algorithm \ref{alg}.
	\begin{algorithm}[h]
		\caption{Quaternion deconvolution}
		\label{alg}
		\textbf{Input:}  blurred image $f$, parameters $\lambda,\gamma$ and initial blur kernel $\mathbf{Q}$;\\
		\textbf{Output:} blur kernel $\mathbf{Q}$ and latent image $\mathbf{u}$;
		\begin{algorithmic}[0]          
			\While{not converged}
		\State update $\mathbf{u}$ by solving (\ref{sub_u});
		\State update $\mathbf{Q}$ by solving (\ref{sub_Q});
		\State calculate $\mathbf{t}$ by solving (\ref{norma});
		\State obtain the normalized $\tilde{\mathbf{Q}}$ by (\ref{norma2});
		\State set $\mathbf{Q} = \tilde{\mathbf{Q}}$ for the next iteration.
		\EndWhile
		\end{algorithmic}
	\end{algorithm}
	
	\subsection{Estimating the latent image $\mathbf{u}$}
	
	We first rewrite the quaternion formulation of subproblem (\ref{sub_u}) into an equivalent matrix-vector version, 
	\begin{equation*}
		\min_{u}\|\tau (\mathbf{Q}\circledast \mathbf{u})  - f \|_2^2 + \lambda \|\nabla u\|_0.
	\end{equation*}
	We make use of the half-quadratic splitting approach \cite{L02} to solve the $L_0$ minimization.
	Specifically, we introduce the auxiliary variables $v$ corresponding to image gradients in the horizontal and vertical directions. 
	The objective function (\ref{sub_u}) can be rewritten as
	\begin{equation}\label{sub_u1}
		\min_{u,v} \|\tau (\mathbf{Q}\circledast \mathbf{u})  - f \|_2^2 + \lambda \|v\|_0 + \beta \|\nabla u -v \|_2^2.
	\end{equation}
	We then consider the following two split problems,
	\begin{equation}\label{sub_2v}
		\min_{v}\lambda \|v\|_0  + \beta \|\nabla u -v \|_2^2
	\end{equation}
	and
	\begin{equation}\label{sub_2u}
		\min_{u}\|\tau (\mathbf{Q}\circledast \mathbf{u}) - f \|_2^2 + \beta \|\nabla u -v \|_2^2.
	\end{equation}
	The solution of (\ref{sub_2v}) is 
	\begin{equation}
		v = \left\{\begin{matrix} 
			\nabla u,&\|\nabla u\|^2\ge \frac{\lambda}{\beta}\\
			0,& \text{otherwise}.
		\end{matrix}\right.
		\label{sg}
	\end{equation}
	Noting that the input image $\mathbf{u}$ is a color image with three channels, we apply the hard-threshold rule of (\ref{sg}) to the channel-wise average to prevent color artifacts at the edges.
	For the subproblem (\ref{sub_2u}), we consider the vector form and rewrite the subproblem as
	\begin{equation*}
		\min_{u}\|T_{\mathbf{Q}} \vec{u} - \vec{f} \|_2^2 + \beta \|D_x \vec{u}  - \vec{v}  \|_2^2+ \beta \|D_y \vec{u}  - \vec{v}  \|_2^2.
	\end{equation*}
	where $D_x$ and $D_y$ are the matrices corresponding to the linear operator $\nabla$.
	Specifically, $D_x = I_4 \otimes I_n \otimes d_m$ and $D_y = I_4 \otimes d_n \otimes I_m$, $m$ and $n$ are the height and width of the input image, $I_n$ denotes the identity matrix of size $n\times n$ and 
	$$d_m = \begin{bmatrix}
		-1 & 1 & & \\
		& \ddots & \ddots & \\
		& & -1 & 1 \\
		1& & & -1
		\end{bmatrix} \in \mathbb{R}^{m\times m}.
		$$
	By calculating the Euler-Lagrange equation, we obtain,
	$$  T_{\mathbf{Q}}^*(T_{\mathbf{Q}} \vec{u} - \vec{f}) +\beta D_x^* (D_x \vec{u} -\vec{v}) +\beta D_y^* (D_y \vec{u} -\vec{v})= 0.$$
	Therefore, we get
	\begin{equation}
		(T_{\mathbf{Q}}^*T_{\mathbf{Q}}  +\beta D_x^* D_x  +\beta D_y^* D_y )\vec{u} = T_{\mathbf{Q}}^*\vec{f}+ \beta (D_x^*+ D_y^*) \vec{v}.
		\label{su}
	\end{equation}
	Under the periodic boundary conditions on $D_X$ and $D_y$, all block matrices $ \{T_{Q_i}^*T_{Q_j}, i, j = 0, 1, 2, 3\}$ in $T_{\mathbf{Q}}^*T_{\mathbf{Q}}$ are block circulant matrices with circulant blocks. Therefore, each block in the coefficient matrix of (\ref{su}) can be diagonalized by the discrete Fourier transform $\mathcal{F}$.
	Applying $\mathcal{F}$ to both sides of (\ref{su}) yields
	\begin{equation}
		\begin{bmatrix}
			\Lambda_{1,1} & \Lambda_{1,2} & \Lambda_{1,3} & \Lambda_{1,4}\\ 
			\Lambda_{2,1} & \Lambda_{2,2} & \Lambda_{2,3} & \Lambda_{2,4}\\ 
			\Lambda_{3,1} & \Lambda_{3,2} & \Lambda_{3,3} & \Lambda_{3,4}\\ 
			\Lambda_{4,1} & \Lambda_{4,2} & \Lambda_{4,3} & \Lambda_{4,4}\\ 
		\end{bmatrix}\begin{bmatrix}
			\mathcal{F}(u_0)\\ \mathcal{F}(u_1)\\ \mathcal{F}(u_2)\\ \mathcal{F}(u_3)
			\end{bmatrix} = \begin{bmatrix}
			b_0\\ b_1\\ b_2\\ b_3
			\end{bmatrix} 
	\end{equation}
	where $\Lambda_{i, j}, i, j = 1, \cdots, 4$ are all diagonal matrices and
	$${\footnotesize \begin{aligned}
		\begin{bmatrix}
			b_0\\ b_1\\ b_2\\ b_3
		\end{bmatrix} = \begin{bmatrix}
			\mathcal{F}(T_{Q_0}) & -\mathcal{F}(T_{Q_1}) & -\mathcal{F}(T_{Q_2}) & -\mathcal{F}(T_{Q_3})\\
			\mathcal{F}(T_{Q_1}) & \mathcal{F}(T_{Q_0}) & -\mathcal{F}(T_{Q_3}) & \mathcal{F}(T_{Q_2})\\
			\mathcal{F}(T_{Q_2}) & \mathcal{F}(T_{Q_3}) & \mathcal{F}(T_{Q_0}) & -\mathcal{F}(T_{Q_1})\\
			\mathcal{F}(T_{Q_3}) & -\mathcal{F}(T_{Q_2}) & \mathcal{F}(T_{Q_1}) & \mathcal{F}(T_{Q_0})
		\end{bmatrix}^*\begin{bmatrix}
			0\\ \mathcal{F}(f_1)\\ \mathcal{F}(f_2)\\ \mathcal{F}(f_3)
		\end{bmatrix}  &\\
		+ \beta (\mathcal{F}(D_x)^*+\mathcal{F}(D_y)^*) \begin{bmatrix}
			0\\ v_1\\ v_2\\ v_3
		\end{bmatrix}&,
	\end{aligned}}$$
	where $\mathcal{F}(T_{Q_i})$ is the diagonal matrix formed by the Fourier transform of the convolution matrix $Q_i$ under the periodic boundary conditions.

	A direct solution of the resulting $4N \times 4N$ linear system would be prohibitively expensive. 
	However, the block-circulant structure induced by the convolution kernel can be diagonalized by Fast Fourier transform (FFT), so each pixel can be regarded as an independent $4 \times 4$ linear system.
	Each small system can be solved in O(1) time, giving an overall complexity of O($N$), where $N = m\times n$ is number of pixels.
	Because the $4 \times 4$ blocks have the same pattern and are completely independent, the computation is highly parallel and approaches O(1) wall-clock time under ideal parallelization.	
	After solving the linear system given in \eqref{su}, we recover the desired latent image $\mathbf{u}$ by applying the inverse FFT to $\mathcal{F}(u_1),\mathcal{F}(u_2)$ and $\mathcal{F}(u_3)$.
	Therefore, the solution of (\ref{sub_u}) can be summarized in Algorithm \ref{alg2}.
	\begin{algorithm}[h]
		\caption{Algorithm for solving (\ref{sub_u})}
		\label{alg2}
		\textbf{Input:}  blurred image $f$, kernel $\mathbf{Q}$, parameter $\lambda$ and initial parameter $\beta$;\\
		\textbf{Output:} the latent image $\mathbf{u}$;
		\begin{algorithmic}[0]      
			\While{$\beta<\beta_{max}$}
			\State solve $g$ by using (\ref{sg});
			\State solve $u$ by using (\ref{su}) for each pixel;
			\State update $\beta$ by $\beta\leftarrow 2\beta$.
			\EndWhile
		\end{algorithmic}
	\end{algorithm}

	\subsection{Estimating the quaternion convolution kernel $\mathbf{Q}$}
	
	Through extensive discussions and applications\cite{FMD2009, DCP2018, SA2021}, gradient-based kernel estimation exploits the richer structural information embedded in image gradients, yielding a more efficient and accurate blur kernel estimating.
	Therefore, we replace $\mathbf{u}$ with $\nabla \mathbf{u}$ in (\ref{sub_Q}) and estimate the quaternion blur kernel $\mathbf{Q}$ by
	\begin{equation*}
		\min_{\mathbf{Q}}\|\mathbf{Q}\circledast \nabla \mathbf{u} - \nabla \mathbf{f} \|_2^2 + \gamma \|\mathbf{Q}\|_2^2.
	\end{equation*}
	The above problem can be rewritten in matrix form, yielding the linear system $AQ=b$, where $A$ can be partitioned into $4 \times 4$ blocks. 
	Each block can be handled by FFT as in \cite{FMD2009}. 
	The system is then solved by using the conjugate-gradient method to obtain the desired kernel $Q$.
	
	Following the state-of-the-art methods, the proposed kernel estimation is performed in a coarse-to-fine manner via an image pyramid \cite{FMD2009}, which effectively alleviates the risk of converging to a local minima caused by the non-convex optimization landscape.
	Specifically, an image pyramid $\{f_1, f_2,\dots , f_L\}$ is constructed from the blurred image f, where $f_1 = f$ and $f_L$ denotes the coarsest down-sampled version. Estimation of the blur kernel and the latent image begins at the coarsest level $L$, and the resulting kernel $\mathbf{Q}_l$ is then up-sampled and passed to level $l-1$ to serve as an initialization for the next finer scale.
	
	\subsection{Incorporating additional image priors}
	
	To incorporate additional image priors, we simply augments model (\ref{sub_u}) by adding additional image priors,
	$$\min_{u}\|\mathbf{Q}\circledast \mathbf{u} - \mathbf{f} \|_2^2 + \lambda \|\nabla u\|_0 + \mu \phi(u),$$	
	where $\phi(u)$ denote the additional image prior, {e.g., the dark-channel prior (\ref{QDCP}) and the surface-aware prior (\ref{QSA}).}
	In general, the above model can be solved by using the half-quadratic splitting technique, specifically, we split it into three sub-problems,
	\begin{itemize}
		\item the sub-problem that involves $L_0$ norm (\ref{sub_2v}), \item the sub-problem that involves the additional image prior
		\begin{equation*}
			\min_{g} \mu \phi(g) + \alpha \|u - g \|_2^2,
		\end{equation*}
		\item the sub-problem that involves the convolution kernel
		$$\min_{\mathbf{u}}\|\Re(\mathbf{Q}\circledast \mathbf{u}) - f \|_2^2 + \beta \|\nabla u -v \|_2^2 + \alpha \|u - g \|_2^2.$$
	\end{itemize} 
	The sub-problem corresponding to the additional image prior can be solved by using similar methods as introduced in \cite{DCP2018} for dark channel prior, \cite{SA2021} for surface-aware prior, etc.

	The sub-problem corresponding to the convolution kernel can be solved by solving the following equation which is similar to (\ref{su}) 
	\begin{equation*}
		\begin{aligned}
			(T_{\mathbf{Q}}^*T_{\mathbf{Q}}  +\beta D_x^* D_x   +&\beta D_y^* D_y +\alpha )\vec{u}\\ &= T_{\mathbf{Q}}^*\vec{f}+ \beta (D_x^* + D_y^*) \vec{v} + \alpha \vec{g},
		\end{aligned}
	\end{equation*}
	and can also be decomposed into $N$ simple linear systems.

	Moreover, we note that existing MAP-based blind image deconvolution methods usually estimate the blur kernel on the grayscale version of the image and then apply this kernel in a non-blind model to obtain the final sharp result, which inspires us to introduce additional image priors in the proposed model to improve the deconvolution effect.

	\section{Experimental Results}

	In this section, we present the numerical results to demonstrate the effectiveness of the proposed quaternion deconvolution model.
	In most of the following experiments, we benchmark the traditional convolution kernel against the proposed quaternion kernel in the $L_0$ + X framework under three different settings, 
	\begin{itemize}
		\item [(1)] No additional prior: the deconvolution model with consistent convolution kernel (CCK) which is consistent for RGB channels given by 
		(\ref{L0X}), and the proposed deconvolution model with quaternion convolution kernel (QCK) given by (\ref{Q_L0}).
		\item [(2)] Dark-channel prior: the conventional model (dark-channel) is derived by adding the surface-aware prior to (\ref{L0X}) with the dark-channel prior, while the quaternion model (dark-channel+QCK) is defined by (\ref{QDCP}).
		\item [(3)] Surface-aware prior: the conventional model (surface-aware) is derived by adding the surface-aware prior to (\ref{L0X}), and the quaternion model (surface-aware+QCK) is specified by (\ref{QSA}).
	\end{itemize}
	We remark here that the results of the dark-channel model \cite{DCP2018} and the surface-aware model \cite{SA2021} are taken directly from the authors' official implementations or reproduced by using the parameters reported in the original papers. For general settings, we fix $\gamma = 2,\beta_{max} = 2^3$ and set the maximum number of outer iterations to 50.
	For the parameters $\lambda$ and $\mu$ introduced by the additional image priors, we make adjustments in the interval [0.004, 0.02]. The proposed algorithm is implemented in MATLAB R2024a.

	\subsection{The role of the normalized quaternion convolution kernel}
	
	\begin{figure}[b]
		\centering
		\includegraphics*[width = 0.19\linewidth]{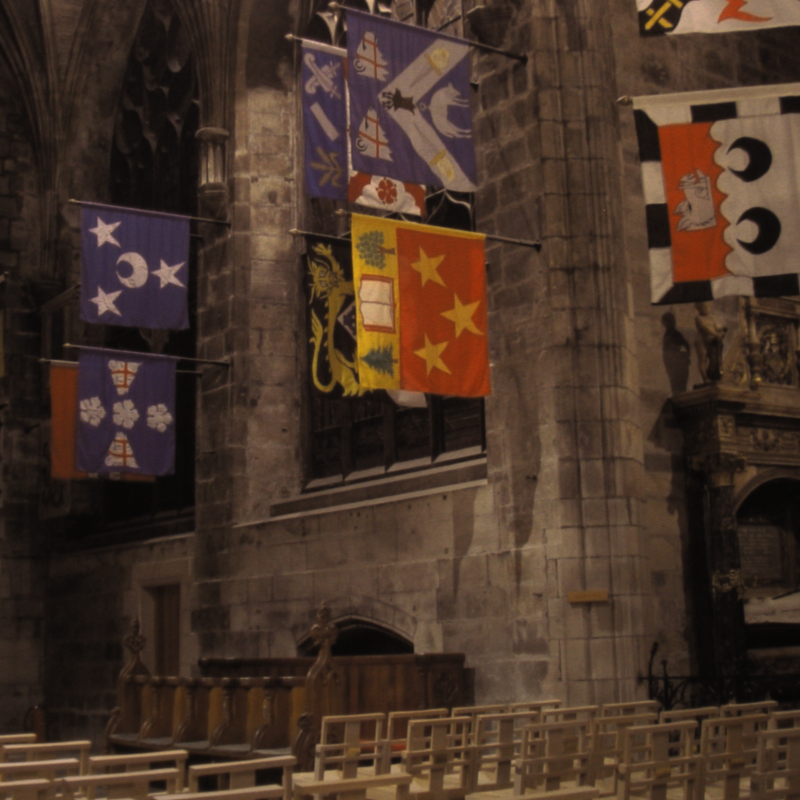}
		\includegraphics*[width = 0.19\linewidth]{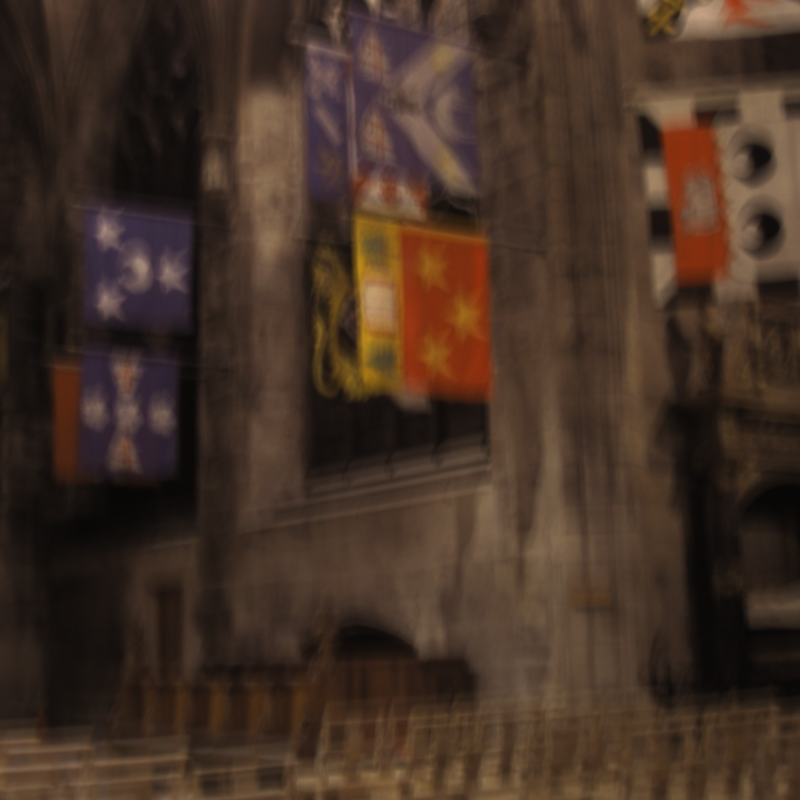}
		\includegraphics*[width = 0.19\linewidth]{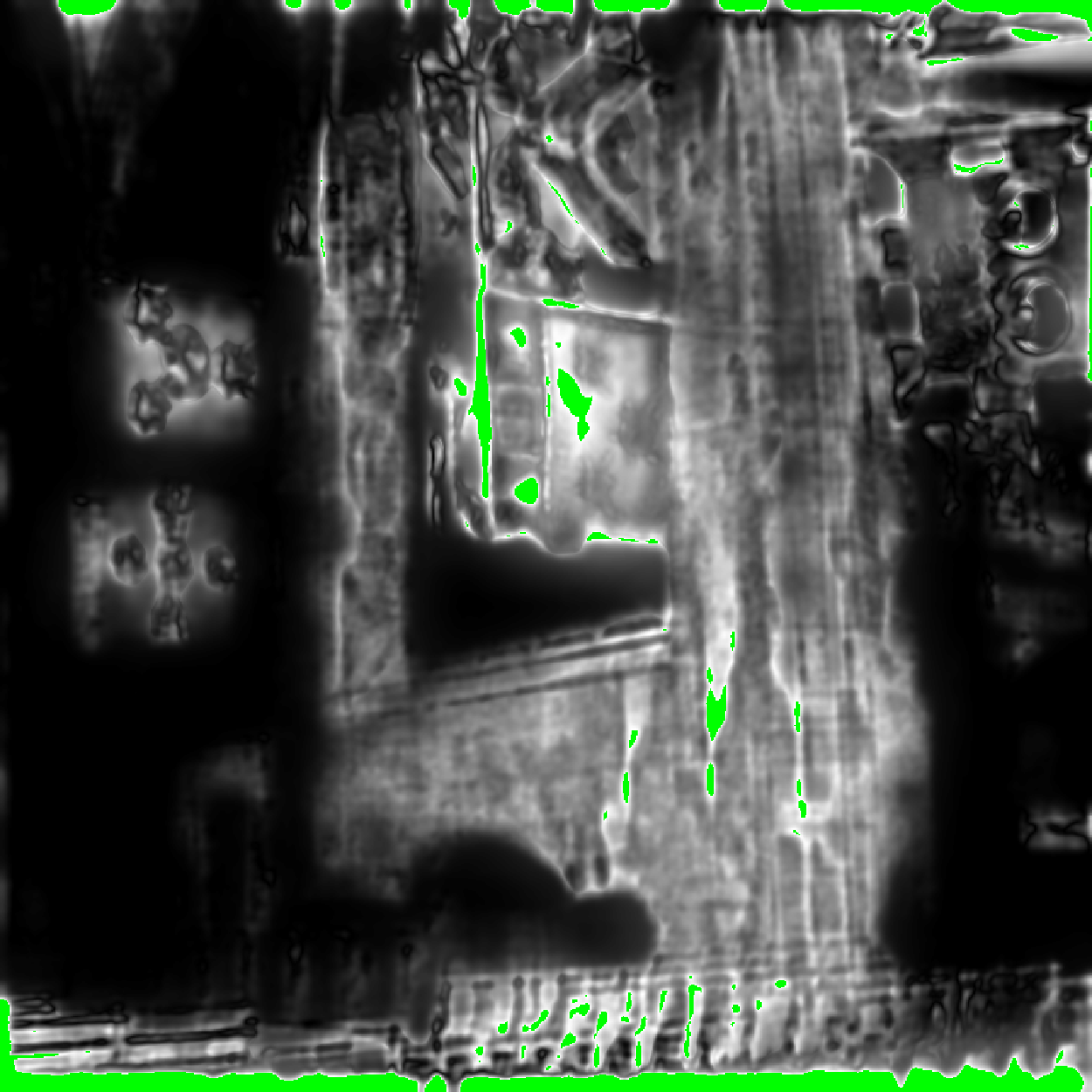}
		\includegraphics*[width = 0.19\linewidth]{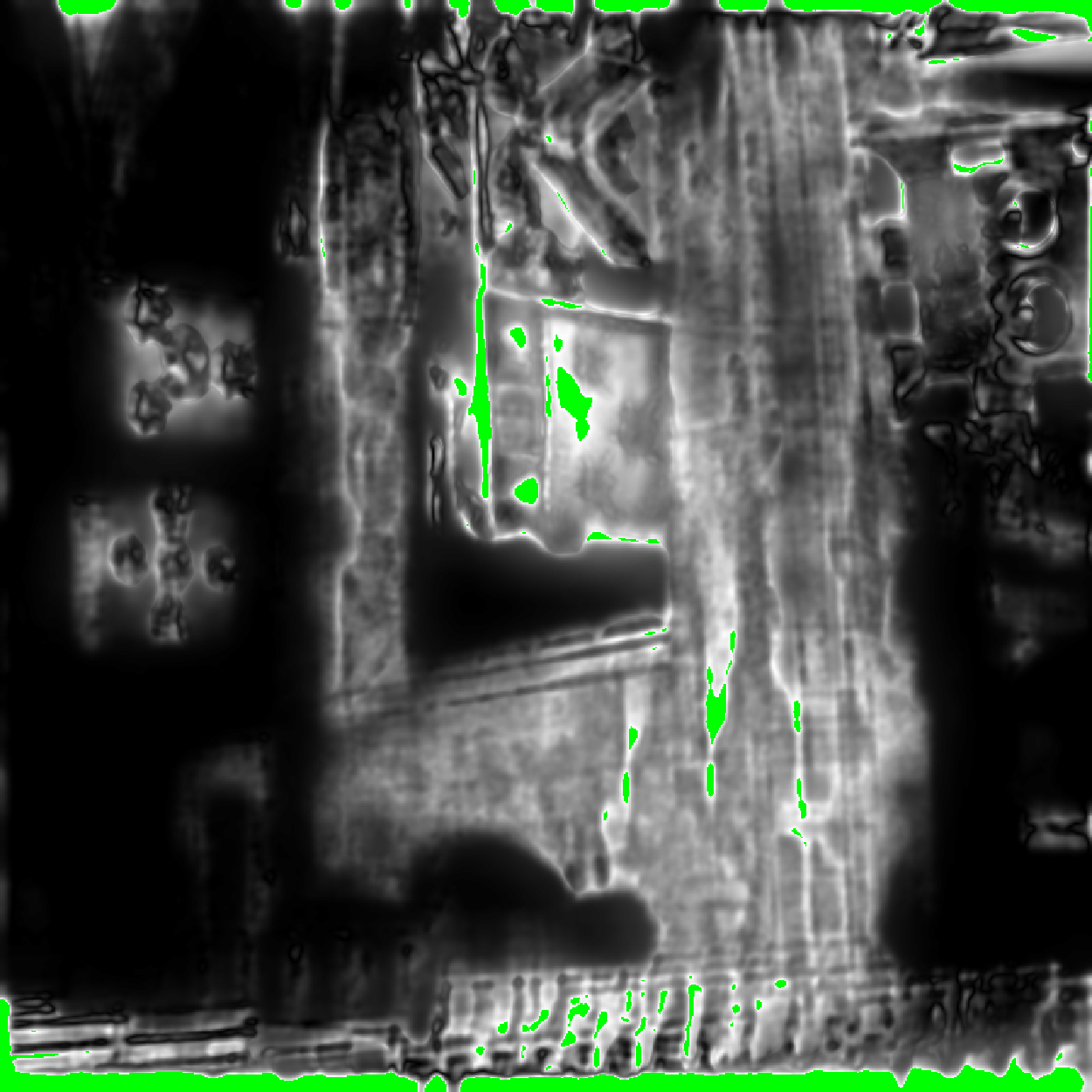}
		\includegraphics*[width = 0.19\linewidth]{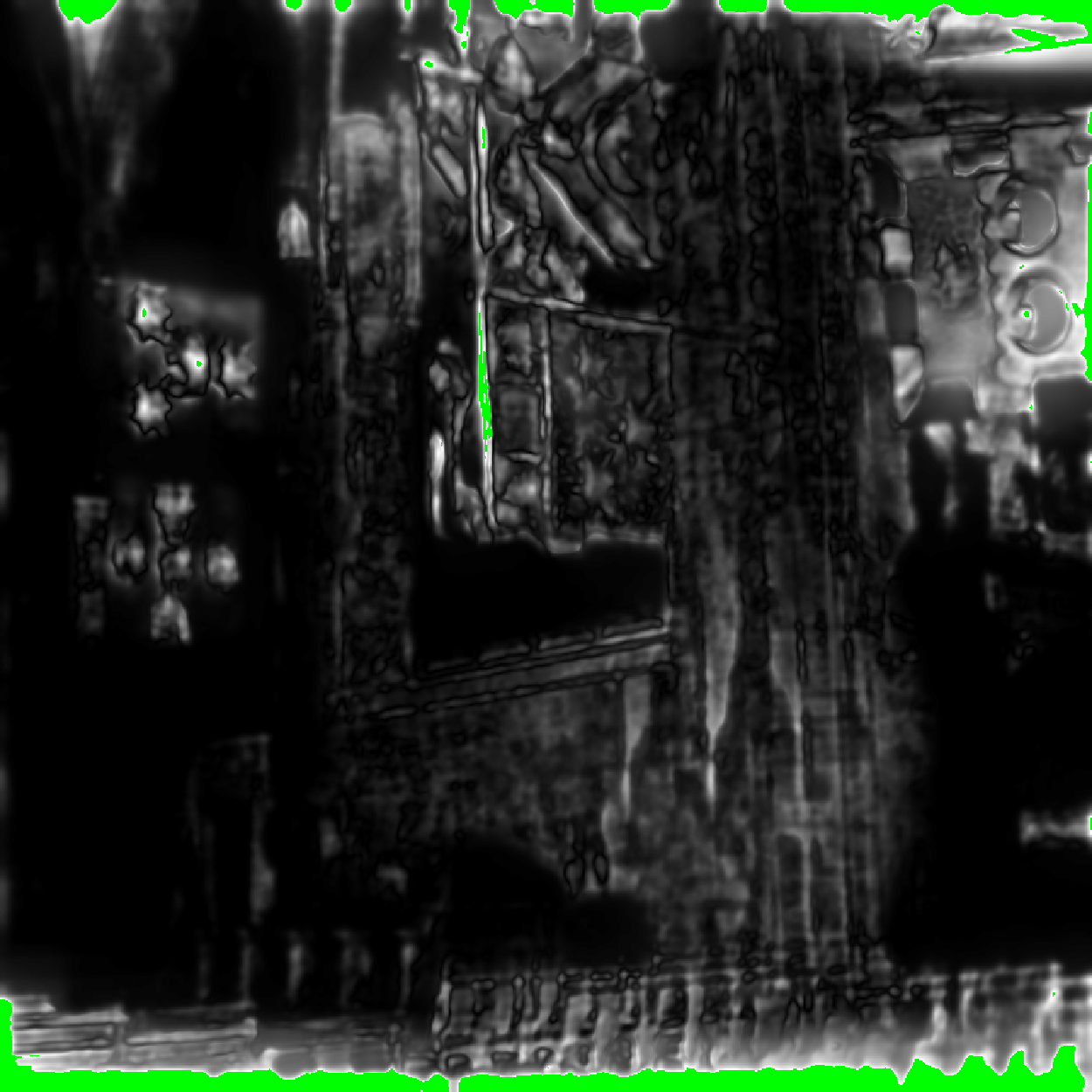}\\\vspace{-1mm}
		\subfigure[sharp]{\includegraphics*[width = 0.19\linewidth]{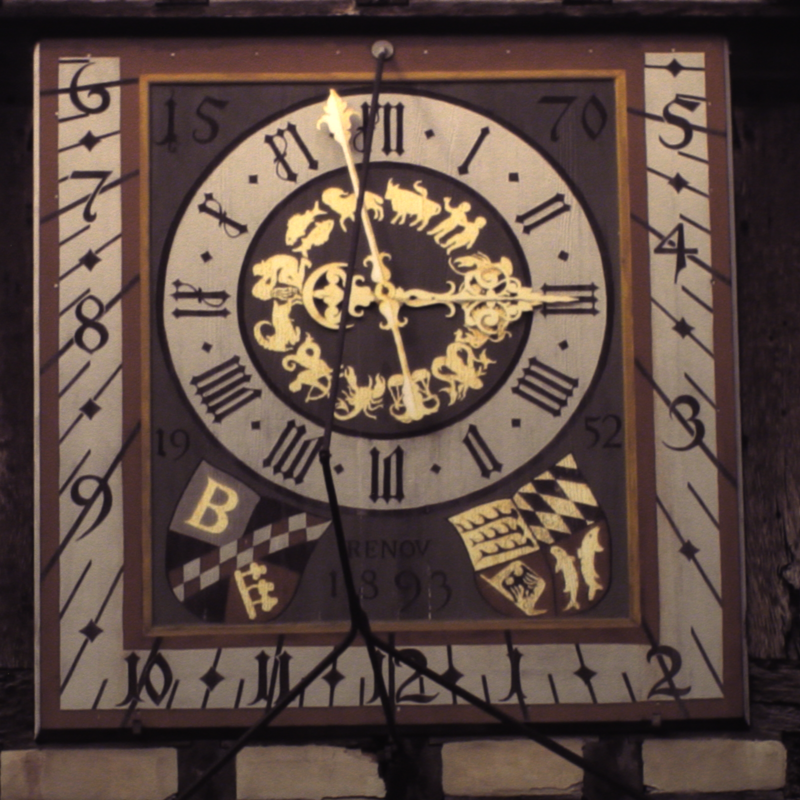}}
		\subfigure[blurred]{\includegraphics*[width = 0.19\linewidth]{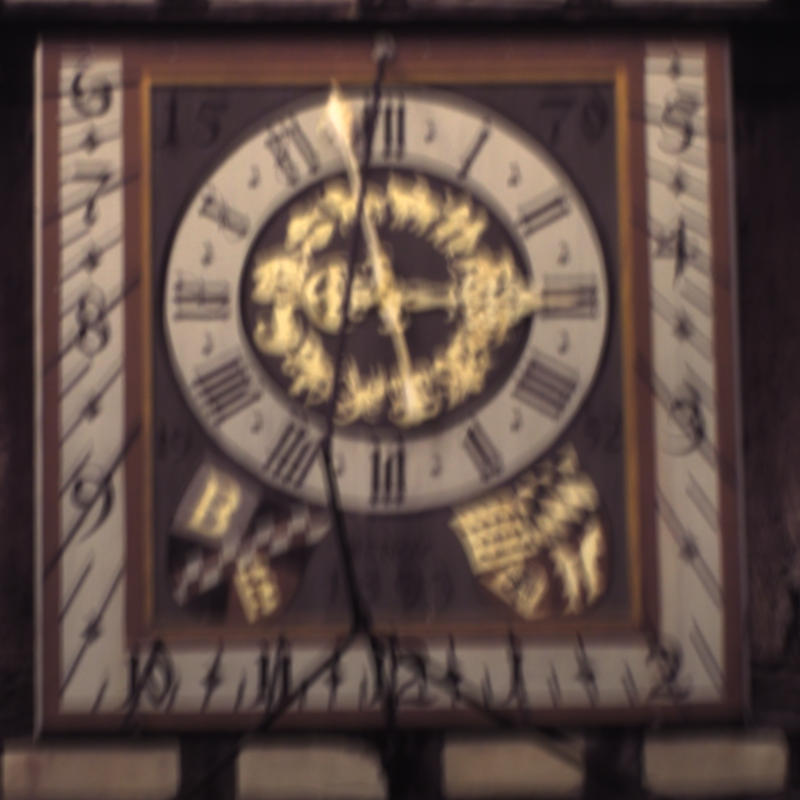}}
		\subfigure[CCK with normalized $k$]{\includegraphics*[width = 0.19\linewidth]{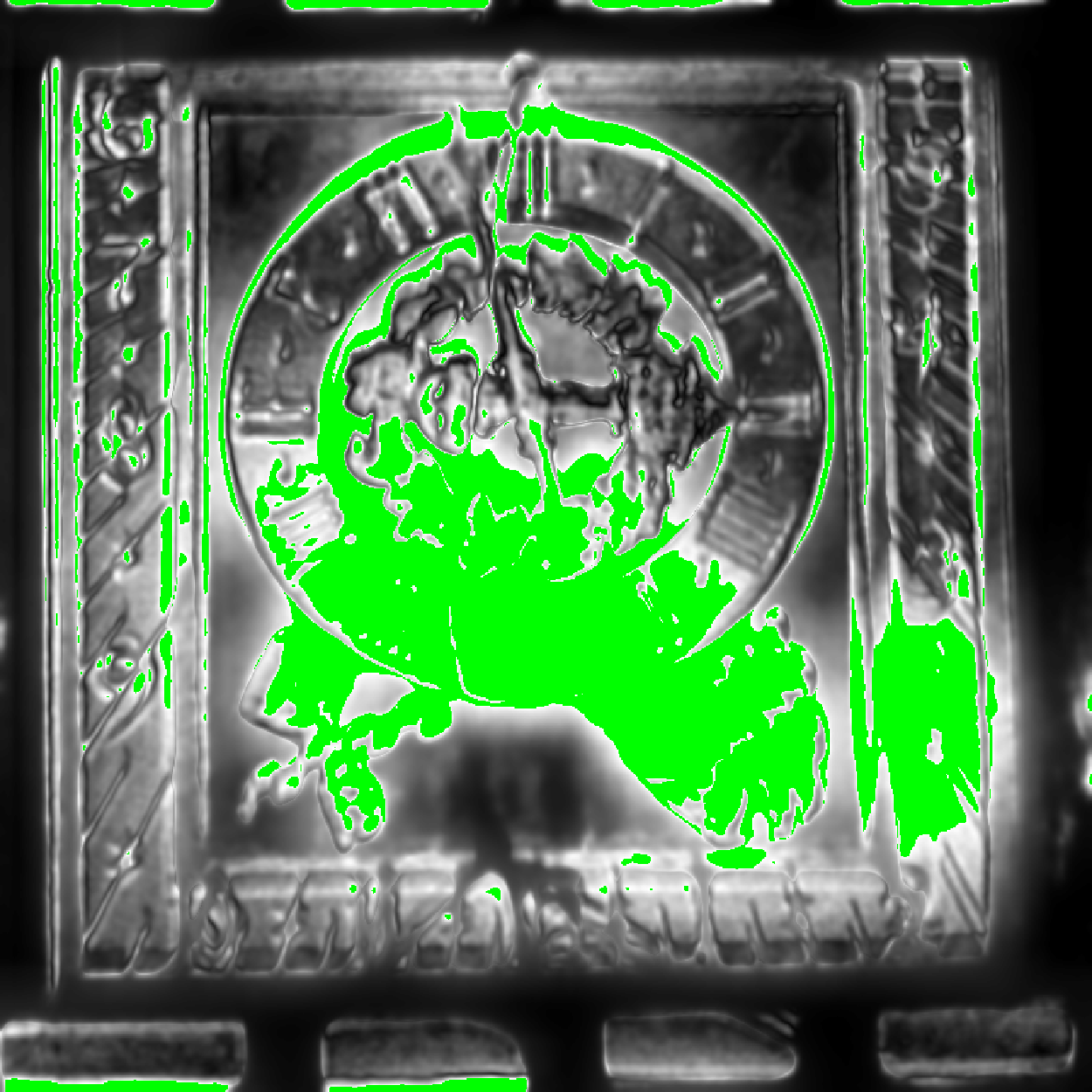}}
		\subfigure[QCK with     $\|\mathbf{Q}\|_1=1$]{\includegraphics*[width = 0.19\linewidth]{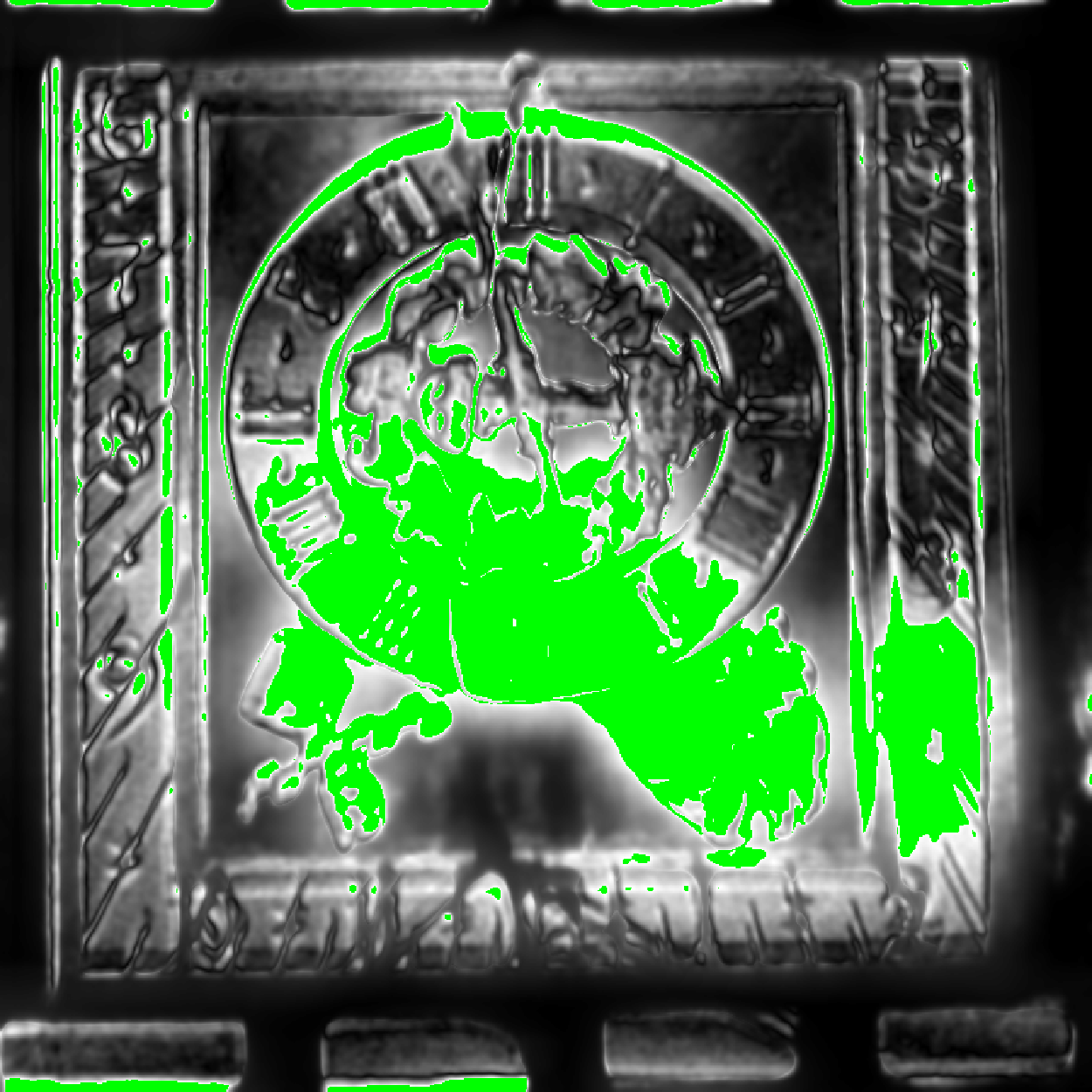}}
		\subfigure[QCK with (\ref{norma2})]{\includegraphics*[width = 0.19\linewidth]{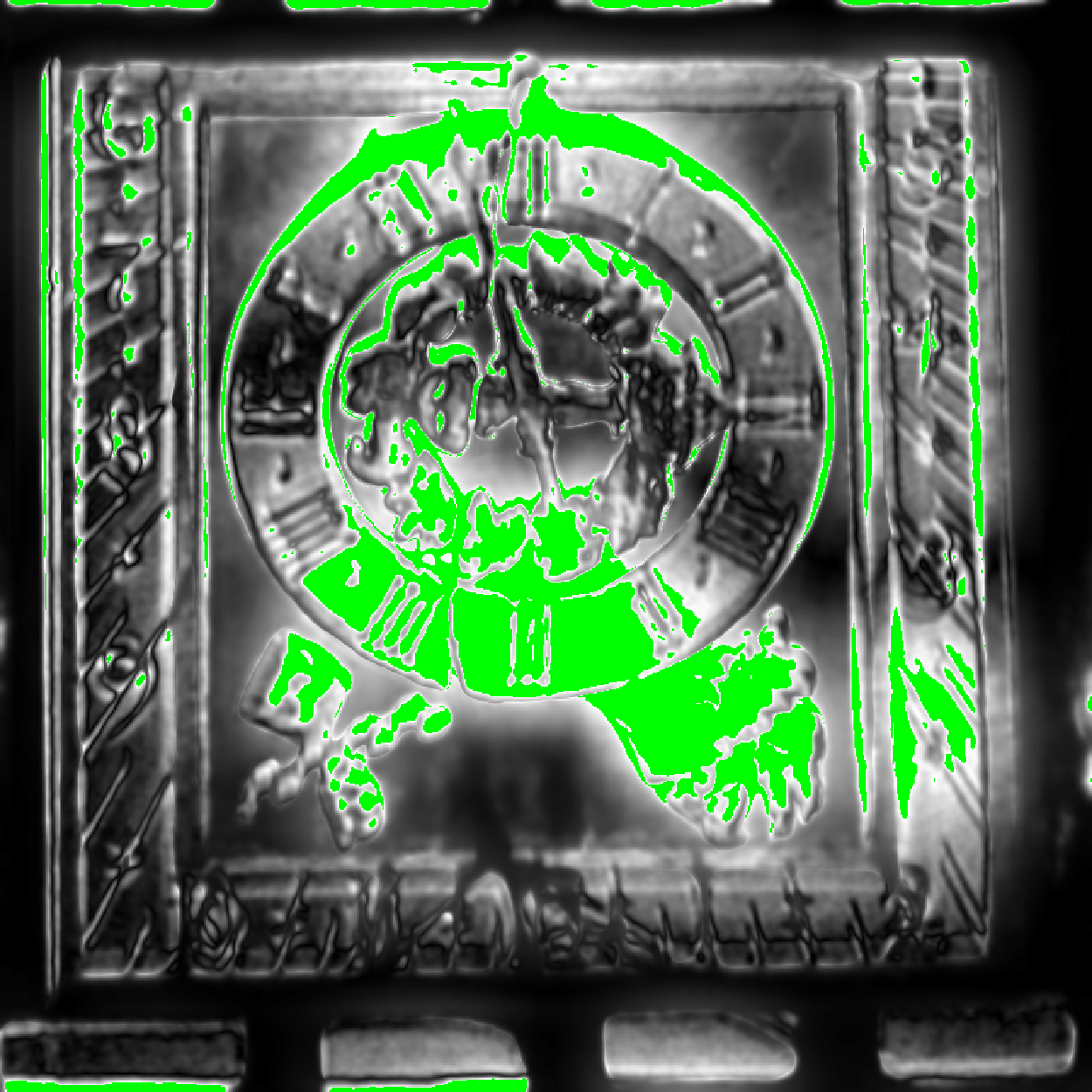}}\\
		\caption{First two: two images from K{\"o}hler dataset; Last three: the spatial distributions of S-CIELAB color error by using different models in the kernel experiments.
			The green regions indicate the pixels with color errors exceeding 5 units, and the number of pixels are 12410, 11675 and 8843 for the first row, 102675, 98809, and 67194 for the second row.}
		\label{fig-inverse}
	\end{figure}
	
	In this subsection, we conduct kernel experiments to compare the proposed normalized quaternion convolution kernel (QCK) with the traditional consistent conventional kernel (CCK). 
	Given a blurred image and its corresponding sharp ground truth, we can estimate the convolution kernel by using different model. 
	For the CCK case, we apply the following estimation model, 
	$$\min_{k}\|k\star I - B \|_2^2 + \gamma \|k\|_2^2, $$
	and adopt the most common normalization method by setting $k>0$ and $\|k\|_1 = 1$.
	For the QCK case, we solve model (\ref{sub_Q}) and apply the two different normalization strategies,
	\begin{itemize}
		\item Normalize the quaternion convolution kernel directly by setting $\|\mathbf{Q}\|_1 = 1$.
		\item Normalize the quaternion convolution kernel by using the proposed normalization scheme (\ref{norma2}), where $\mathbf{t}$ will be solved by (\ref{norma}).
	\end{itemize}
	
	\begin{figure}[t]
		\centering
		\includegraphics*[width = \linewidth,height=3cm]{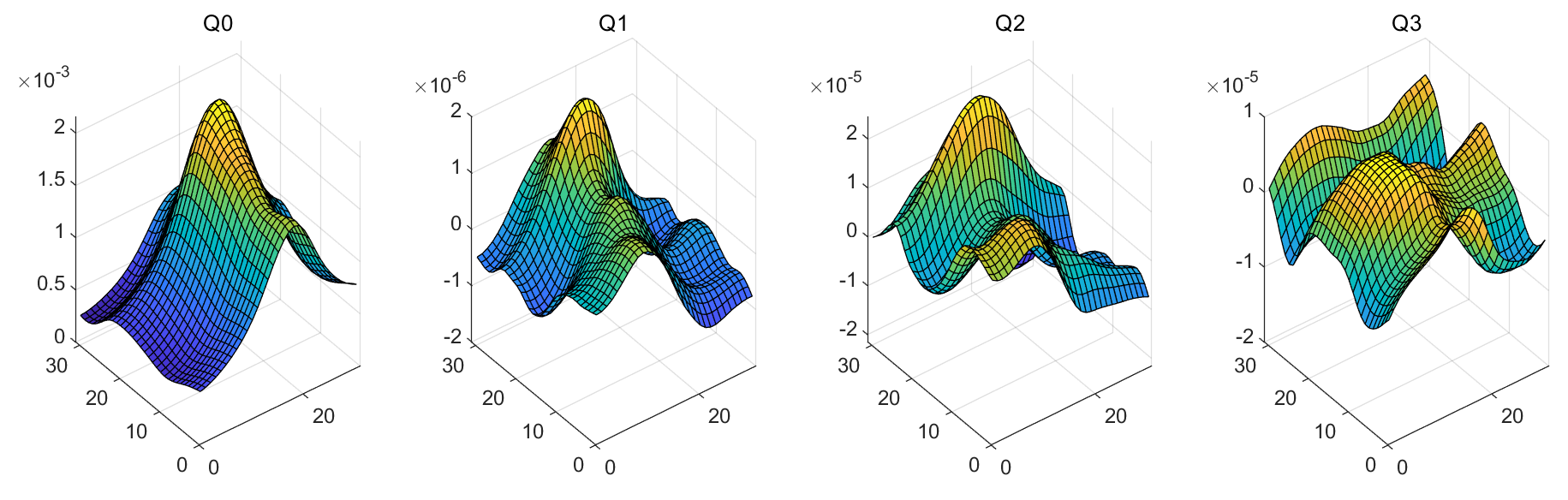}
		\includegraphics*[width = \linewidth,height=3cm]{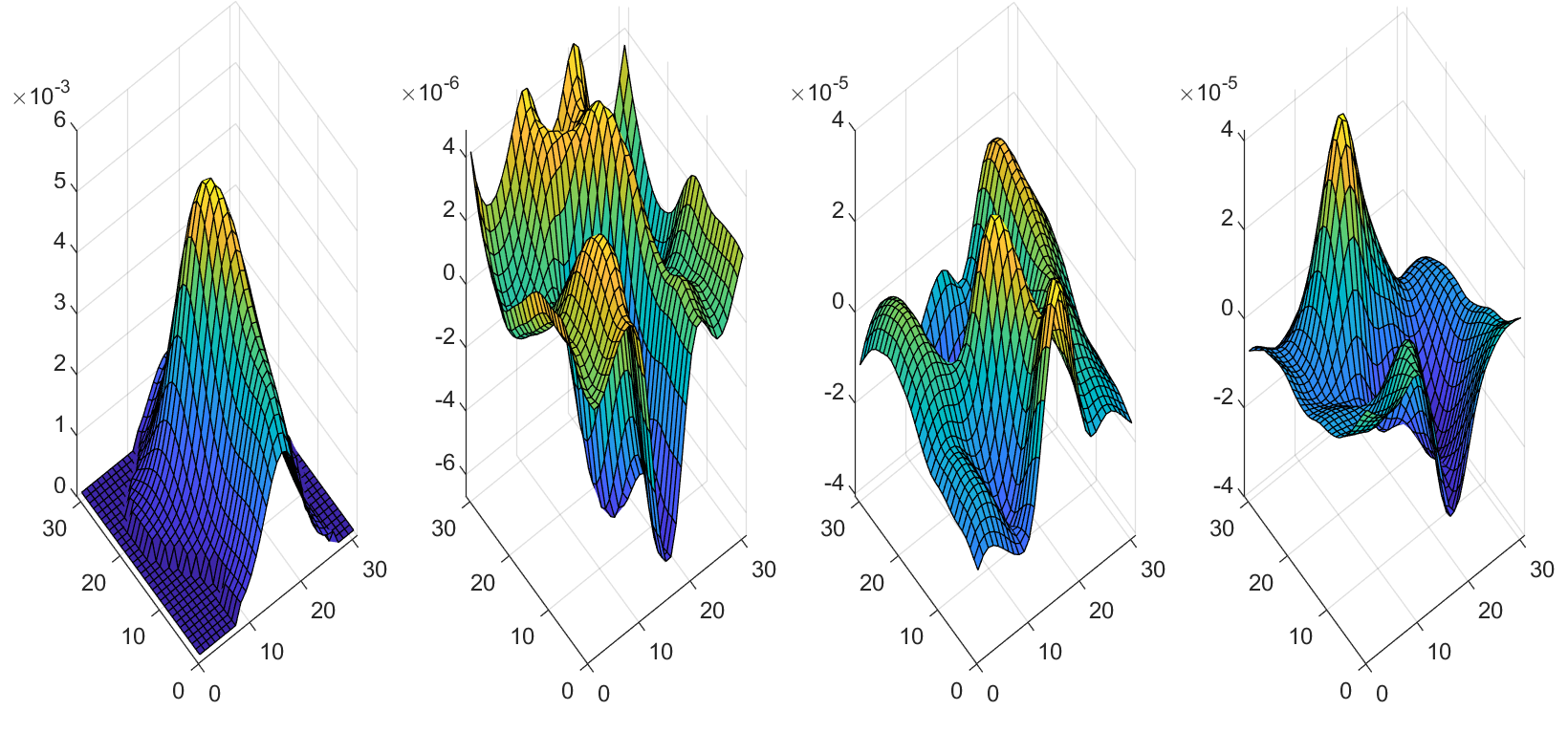}
		\caption{The reconstructed quaternion convolution kernels 
			$Q_0$, $Q_1$, $Q_2$ and $Q_3$ of the image in
			Figure 2 (top one).}
		\label{Q-SHOW}
	\end{figure}

	In the first experiment, we consider the image dataset proposed by K{\"o}hler et al.\cite{2012kholer} that captures real blurry images and records camera shake trajectories, which are then replayed via a robotic platform.	The dataset contains four sharp images and their blurred images with multiple motion trajectories (12 blurred images). To evaluate the accuracy of the obtained kernels, we reapply each estimate to the corresponding sharp image and synthesize blurred images for direct comparison with the real blurred images.
	In Figure \ref{fig-inverse}, we show two results obtained based on two testing pairs in the kernel experiment. 
	We display the spatial distributions of the S-CIELAB error, where green regions represent pixels that the color errors exceed 5 units. 
	In Table \ref{table-inverse}, we report the average PSNR, SSIM, and S-CIELAB color error values of the 12 deblurring results corresponding to each image, as well as the overall average values of all 48 deblurring results.	 
	We see from Figure \ref{fig-inverse} and Table \ref{table-inverse} that the proposed model by using normalized quaternion convolution kernels is always
	better than the traditional consistent convolution kernel model in terms of measure values.
	\begin{table}[t]
		\centering
		\caption{The average PSNR, SSIM and S-CIELAB color error values of four testing images in K{\"o}hler dataset.}
		\begin{tabular}{|c|c| m{1.3cm}<{\centering}|m{1.3cm}<{\centering}| m{1.3cm}<{\centering}|}
			\hline
			Test image               & Measure  & CCK with normalized $k$   & QCK with     $\|\mathbf{Q}\|_1=1$ & QCK with (\ref{norma2}) \\ \hline
			\multirow{3}{*}{\#1}    & PSNR     & 32.31  & 32.35                             & \textbf{32.57}                          \\ 
			& SSIM     & 0.9795 & 0.9798                            & \textbf{0.9800}                         \\ 
			& S-CIELAB & 77556  & 74371                             & \textbf{66665}                          \\ \hline
			\multirow{3}{*}{\#2}    & PSNR     & 31.68  & 31.73                             & \textbf{32.06}                          \\ 
			& SSIM     & 0.9779 & 0.9782                            & \textbf{0.9787}                         \\ 
			& S-CIELAB & 98232  & 94611                             & \textbf{87416}                          \\ \hline
			\multirow{3}{*}{\#3}    & PSNR     & 29.88  & 29.87                             & \textbf{29.93}                          \\ 
			& SSIM     & 0.9720 & 0.9721                            & \textbf{0.9724}                         \\ 
			& S-CIELAB & 112173 & 110518                            & \textbf{108896}                         \\ \hline
			\multirow{3}{*}{\#4}    & PSNR     & 30.93  & 30.98                             & \textbf{31.07}                          \\ 
			& SSIM     & 0.9777 & 0.9781                            & \textbf{0.9783}                         \\ 
			& S-CIELAB & 98591  & 95067                             & \textbf{88856}                          \\ \hline
			\multirow{3}{*}{Average} & PSNR     & 31.20  & 31.25                             & \textbf{31.39}                          \\ 
			& SSIM     & 0.9768 & 0.9771                            & \textbf{0.9773}                         \\ 
			& S-CIELAB & 96638  & 93236                             & \textbf{88364}                          \\ \hline
		\end{tabular}
		\label{table-inverse}
	\end{table}	
	In Figure \ref{Q-SHOW}, we display the recovered quaternion convolution kernels, and we find that $Q_0$ functions similarly to a non-negative consistent convolution kernel to capture the overall blur, and $Q_1$, $Q_2$, $Q_3$ are corresponding to red, green and blue channels respectively to model their unknown interdependencies. 
	Noting that the authors have observed in \cite{Kosik2025} that the green and blue blurs are almost identical, which is consistent with that the magnitude of $Q_1$ reflecting the blurring differences between green and blue channels, is noticeably smaller than that of $Q_2$ and $Q_3$. Meanwhile, we emphasize that $Q_1$, $Q_2$ and $Q_3$ contain negative entries to reflect the interdependencies among red, green and blue channels.
		\renewcommand{\dbltopfraction}{1}
	\begin{figure}[t]
		\centering
		\includegraphics*[width = 0.19\linewidth]{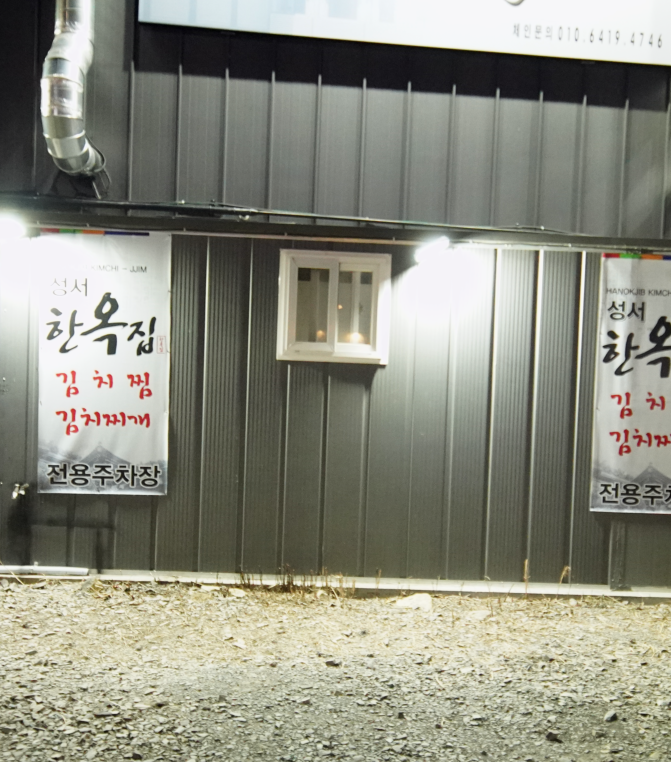}
		\includegraphics*[width = 0.19\linewidth]{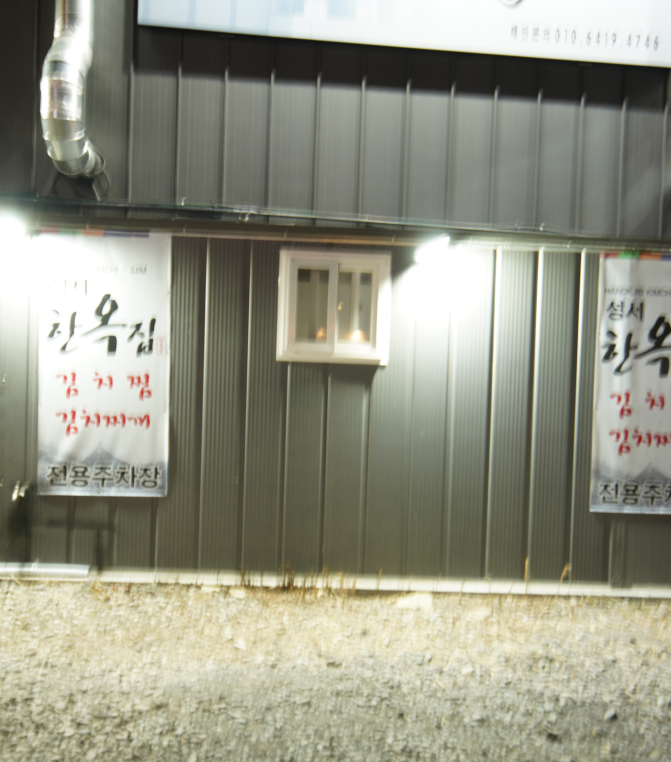}
		{\includegraphics*[width = 0.19\linewidth]{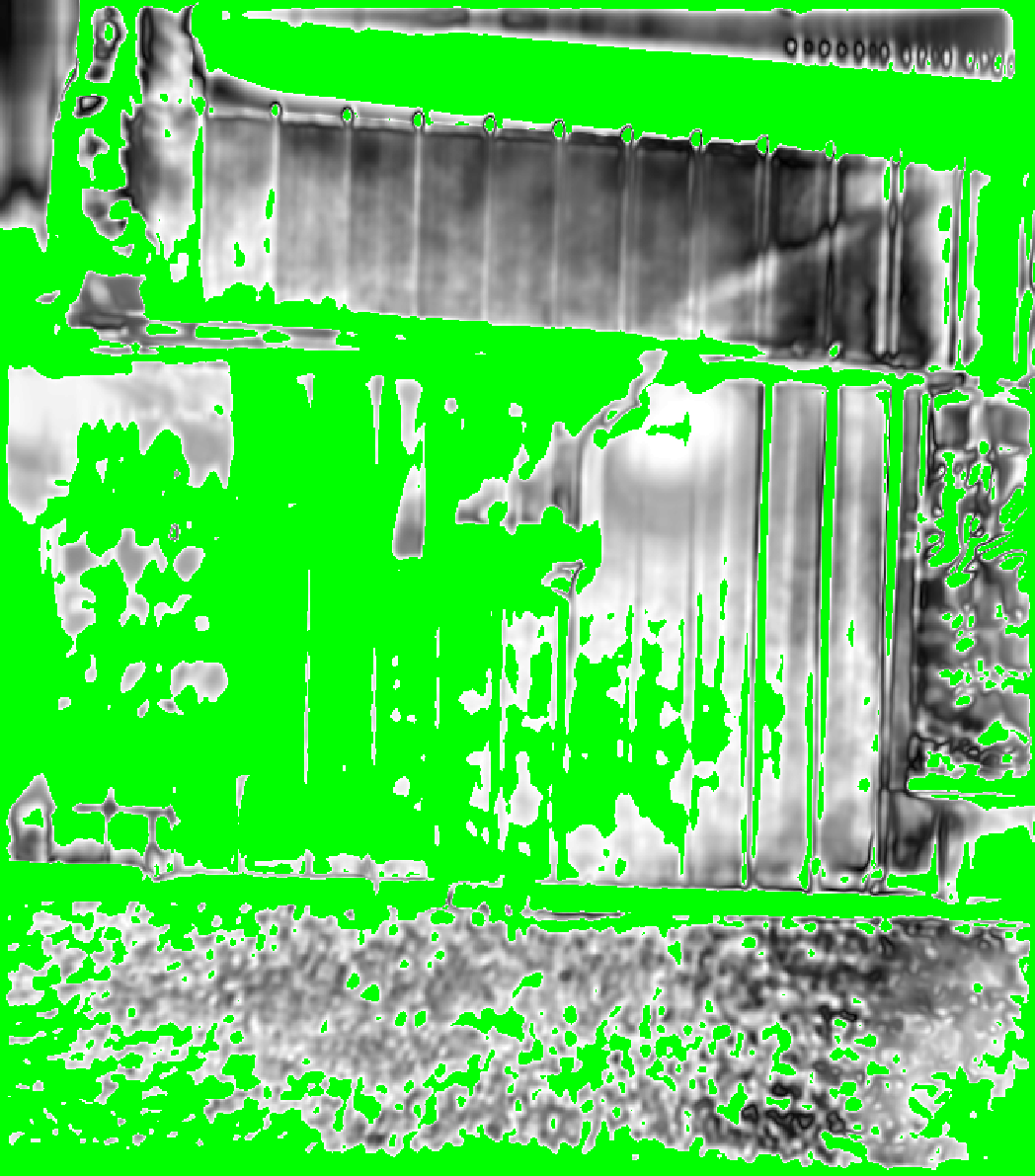}}
		{\includegraphics*[width = 0.19\linewidth]{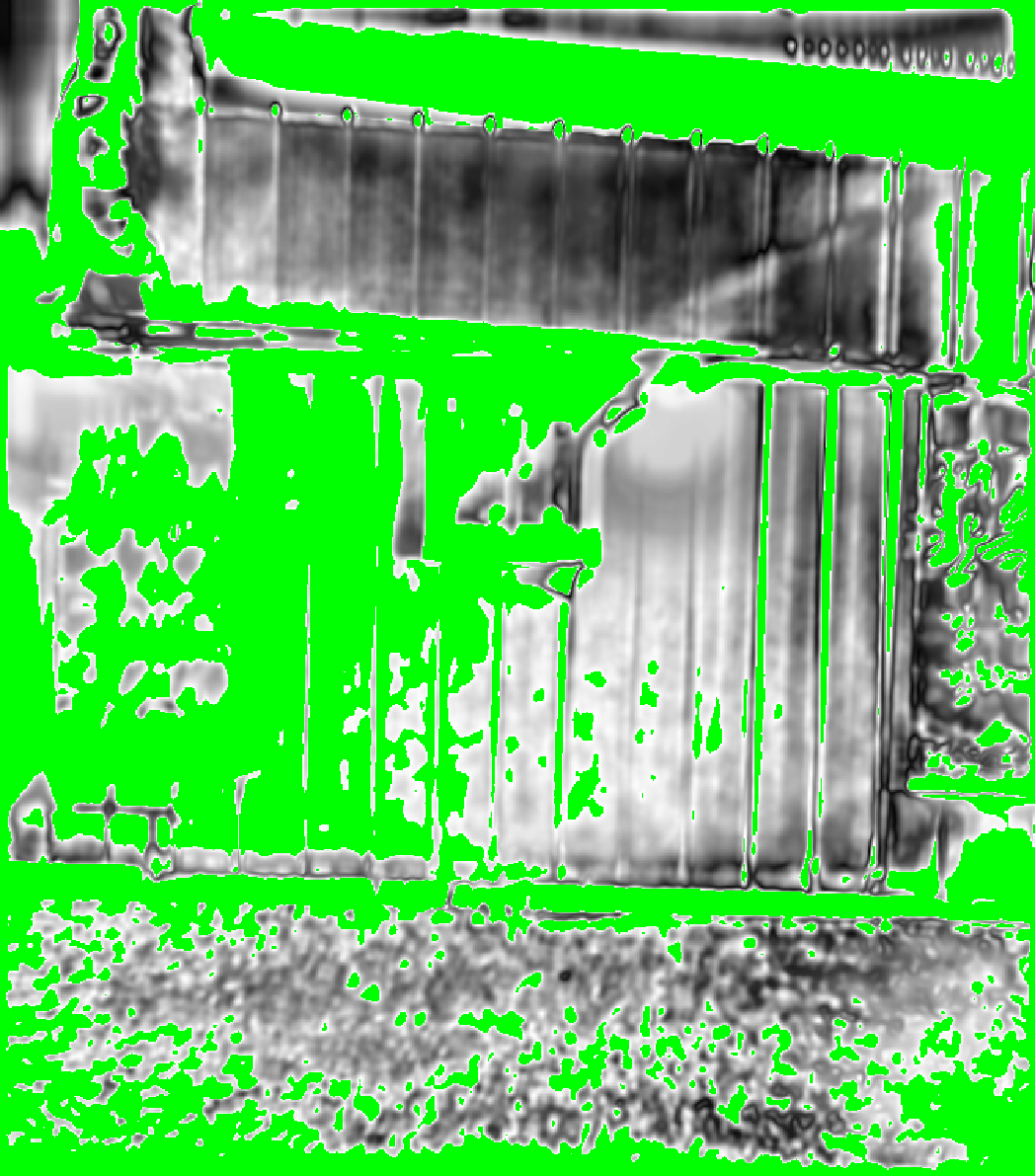}}
		\includegraphics*[width = 0.19\linewidth]{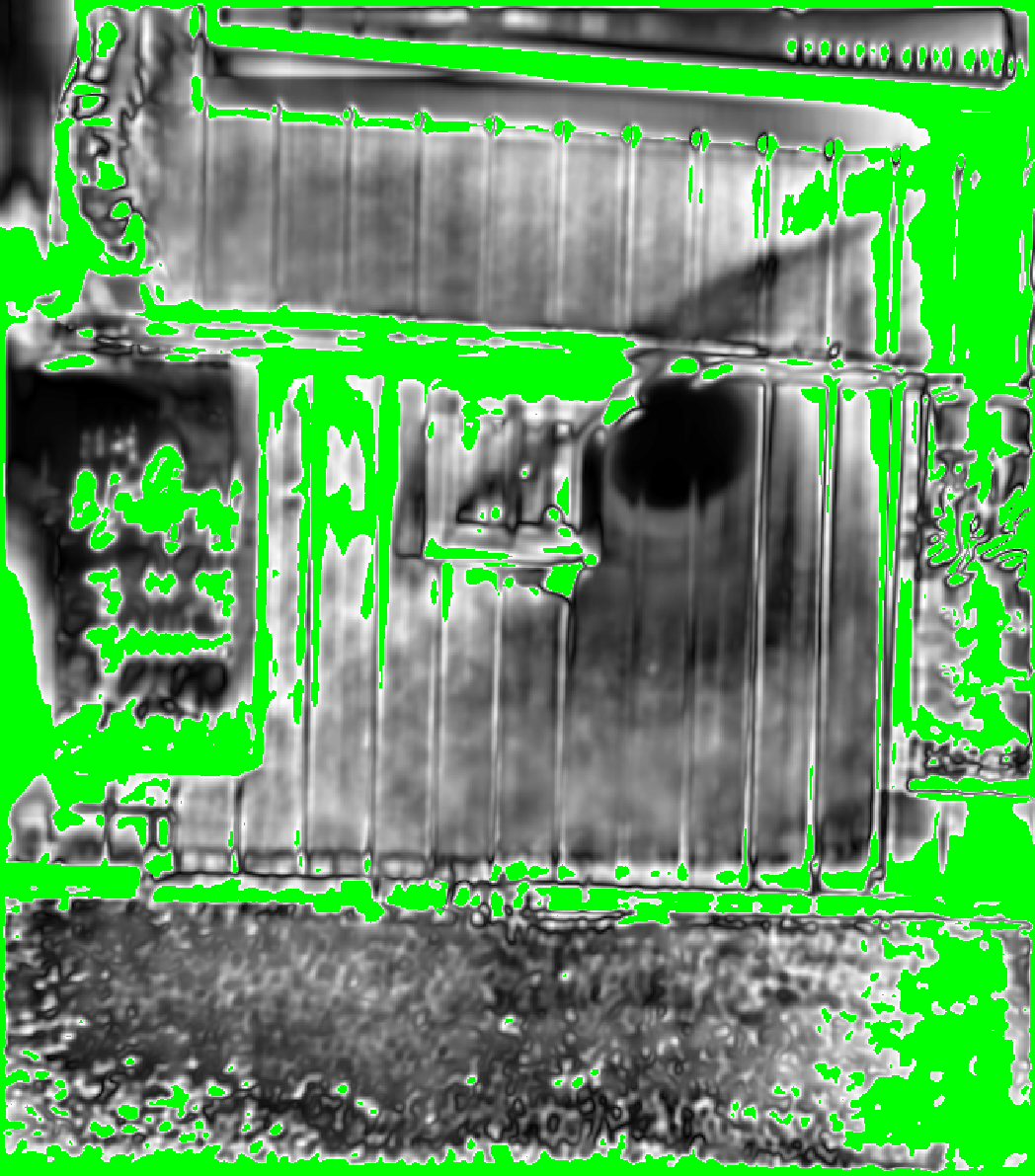}\\\vspace{1mm}
		\includegraphics*[width = 0.19\linewidth]{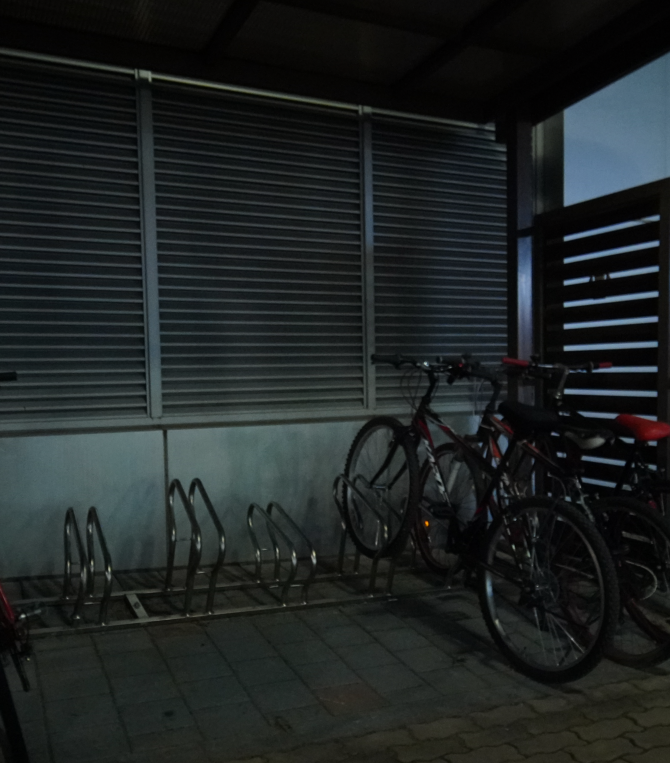}
		\includegraphics*[width = 0.19\linewidth]{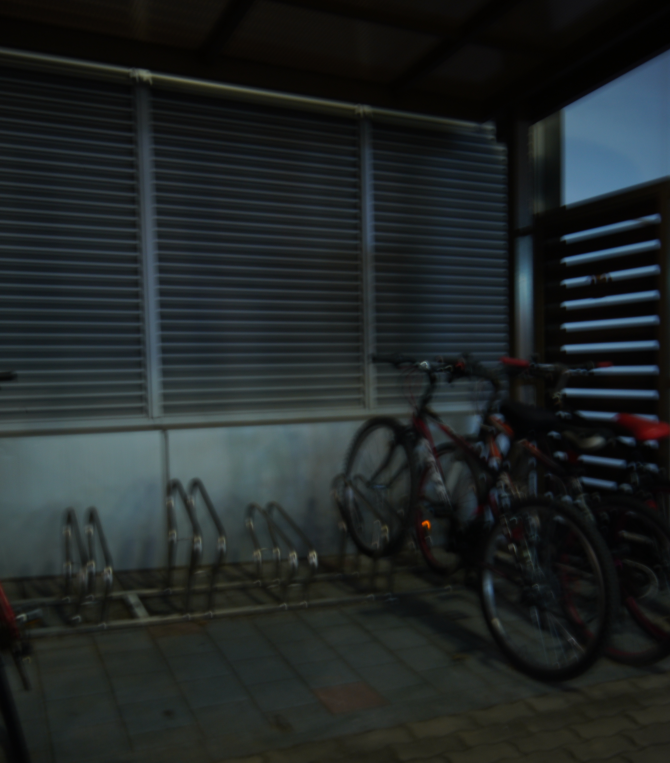}
		\includegraphics*[width = 0.19\linewidth]{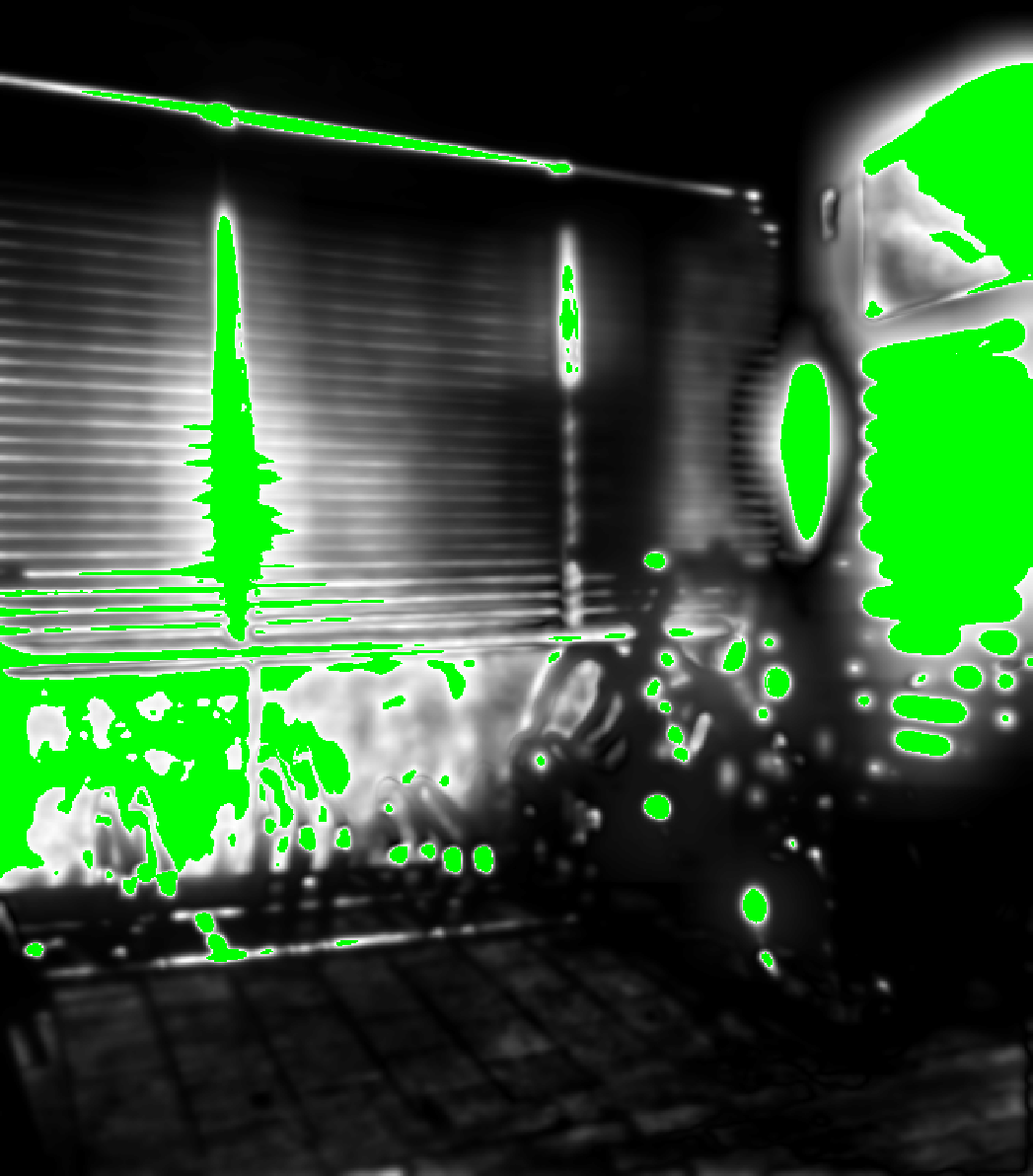}
		\includegraphics*[width = 0.19\linewidth]{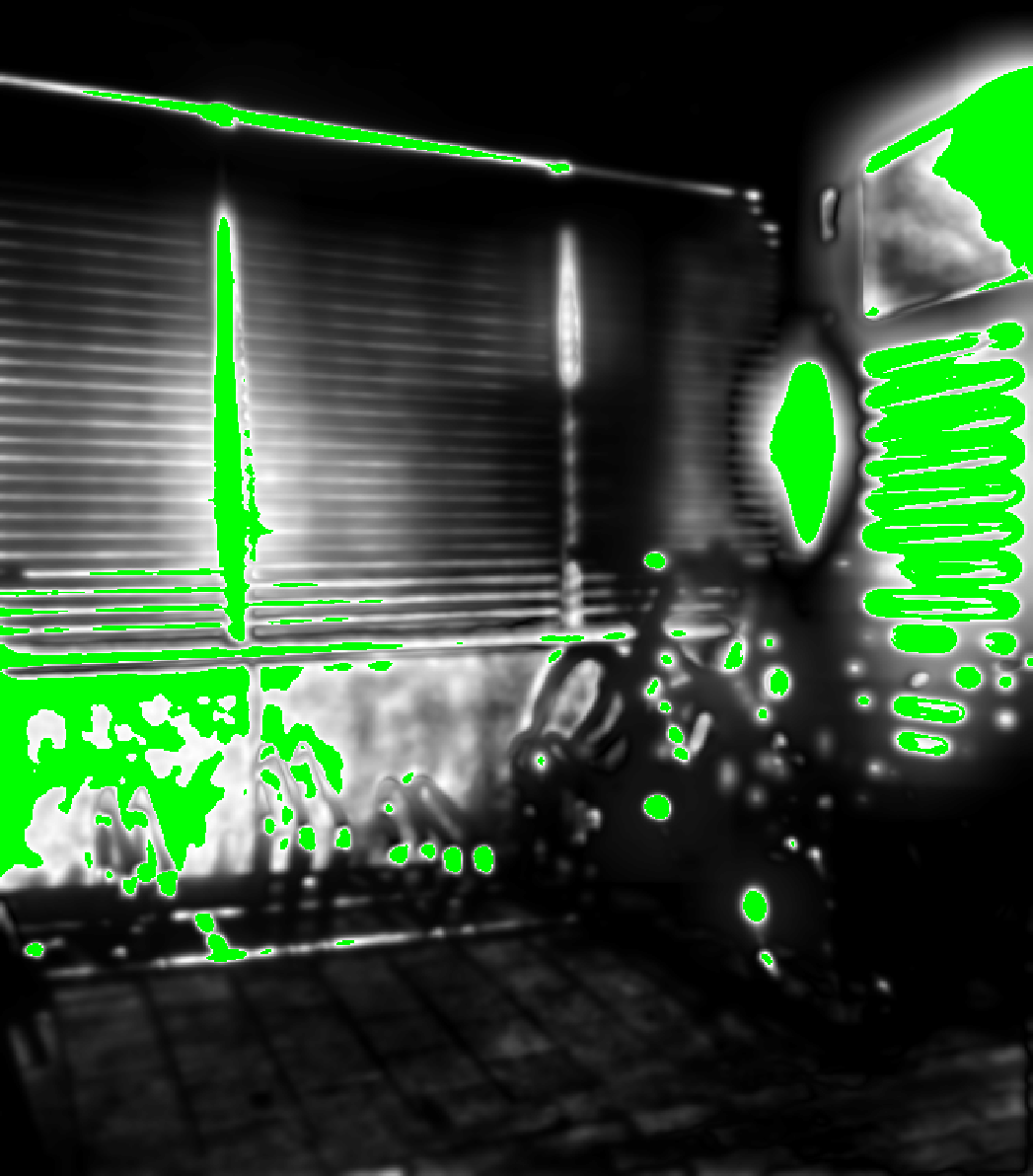}
		\includegraphics*[width = 0.19\linewidth]{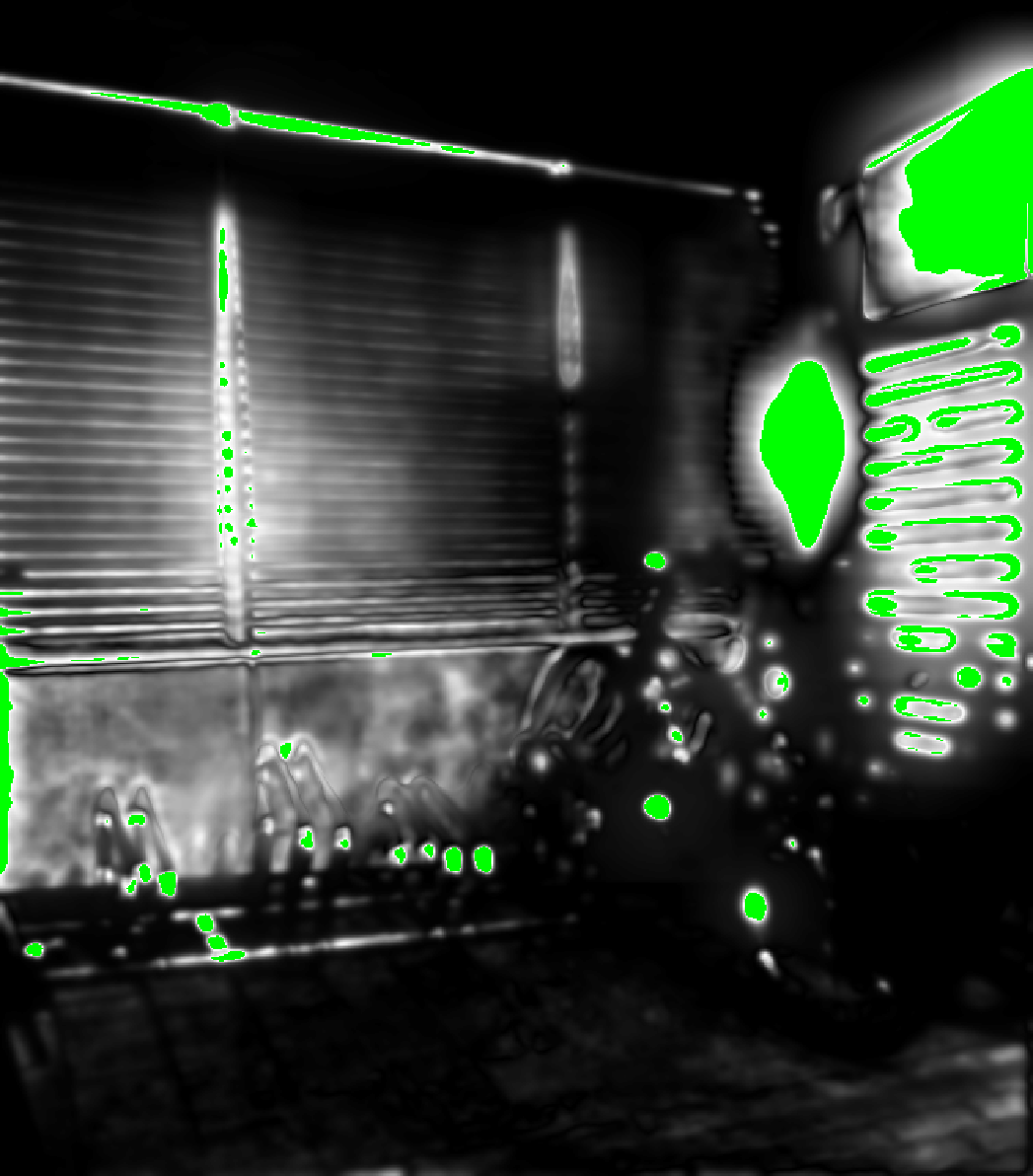}\\\vspace{1mm}
		\includegraphics*[width = 0.19\linewidth]{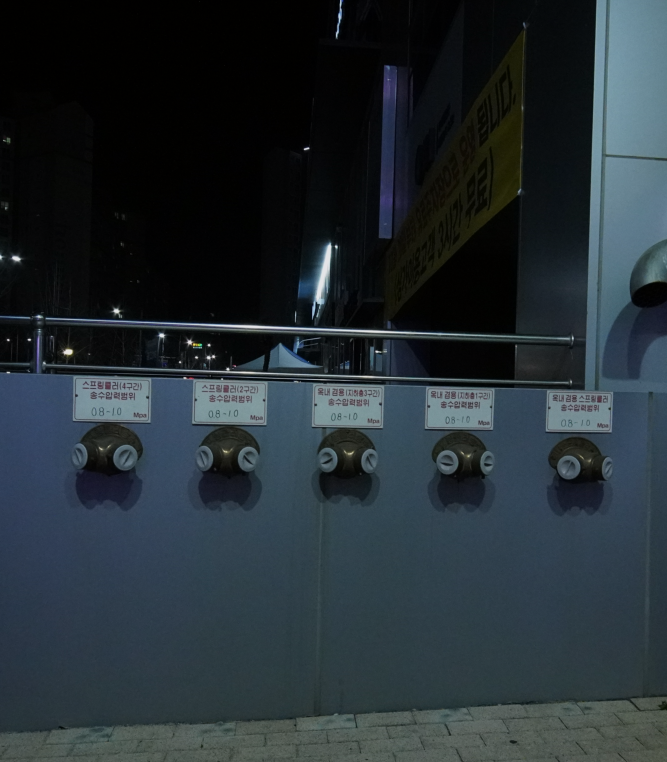}
		\includegraphics*[width = 0.19\linewidth]{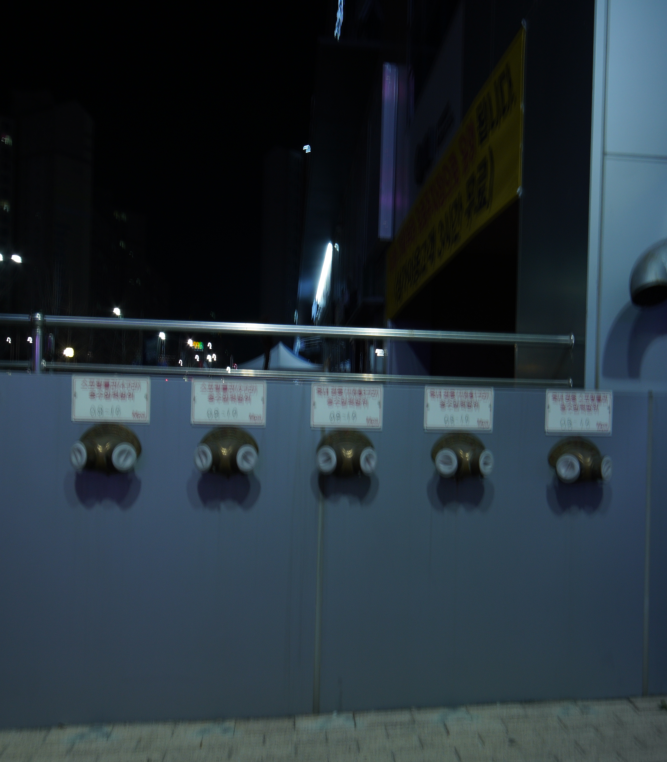}
		\includegraphics*[width = 0.19\linewidth]{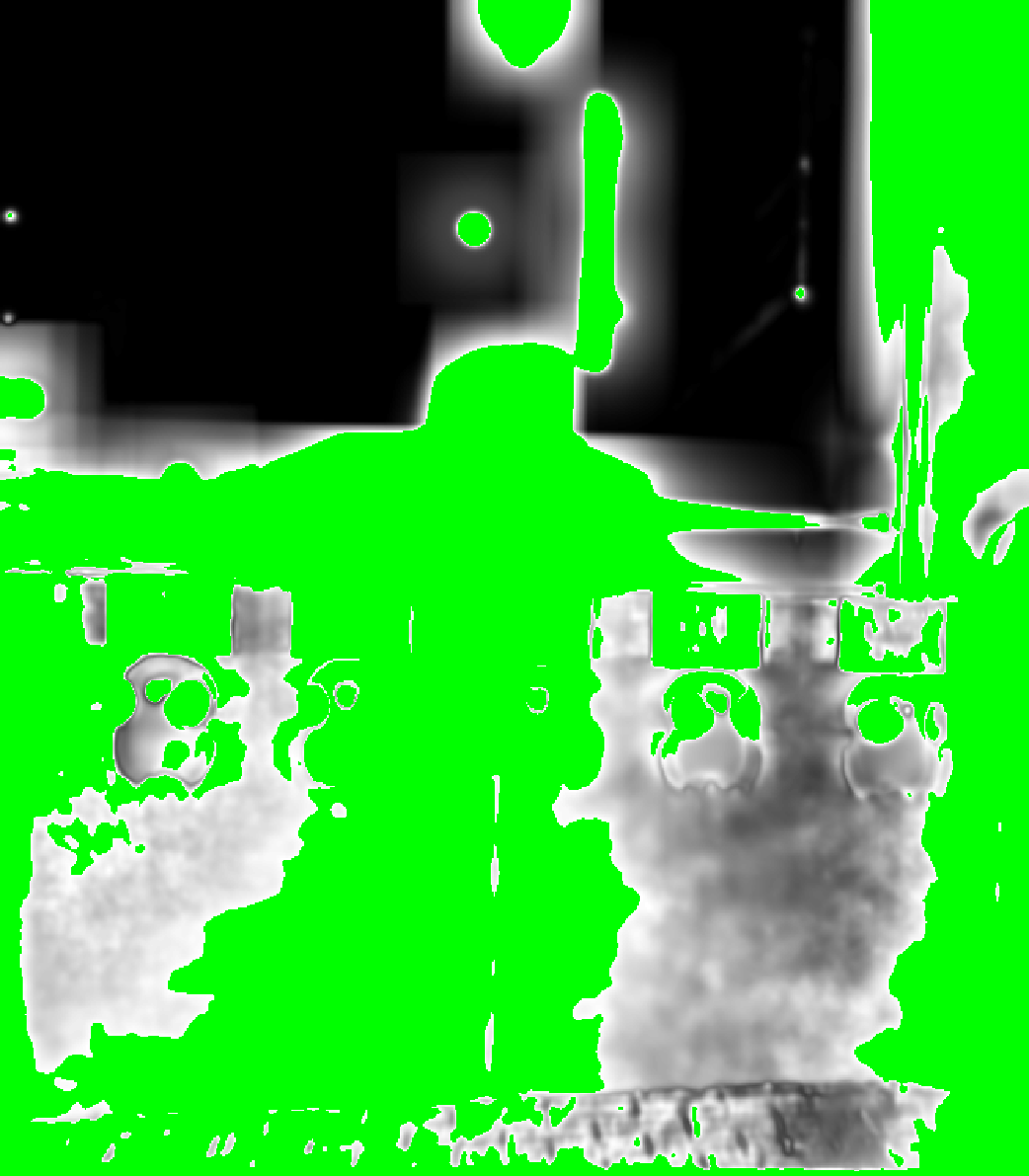}
		\includegraphics*[width = 0.19\linewidth]{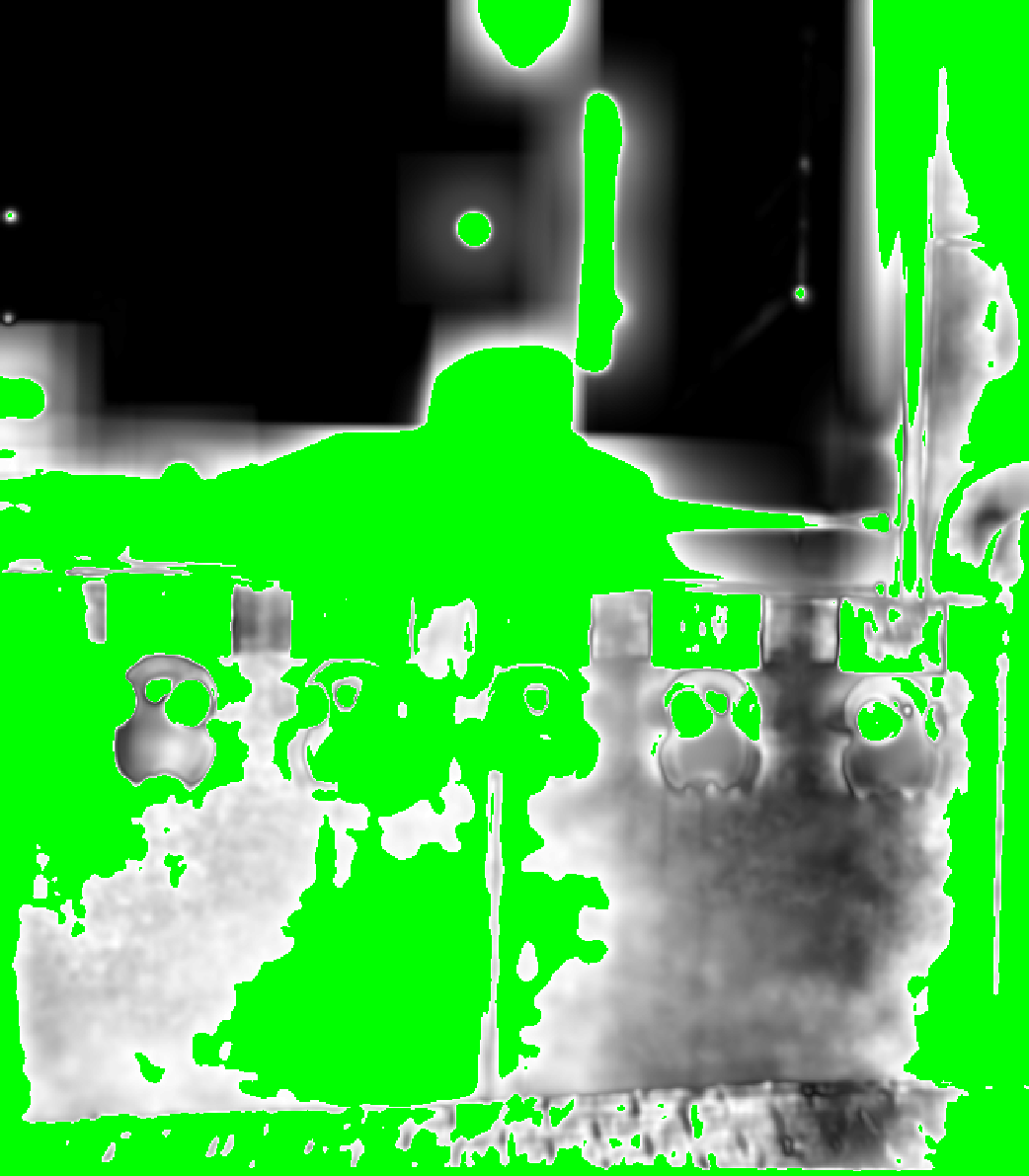}
		\includegraphics*[width = 0.19\linewidth]{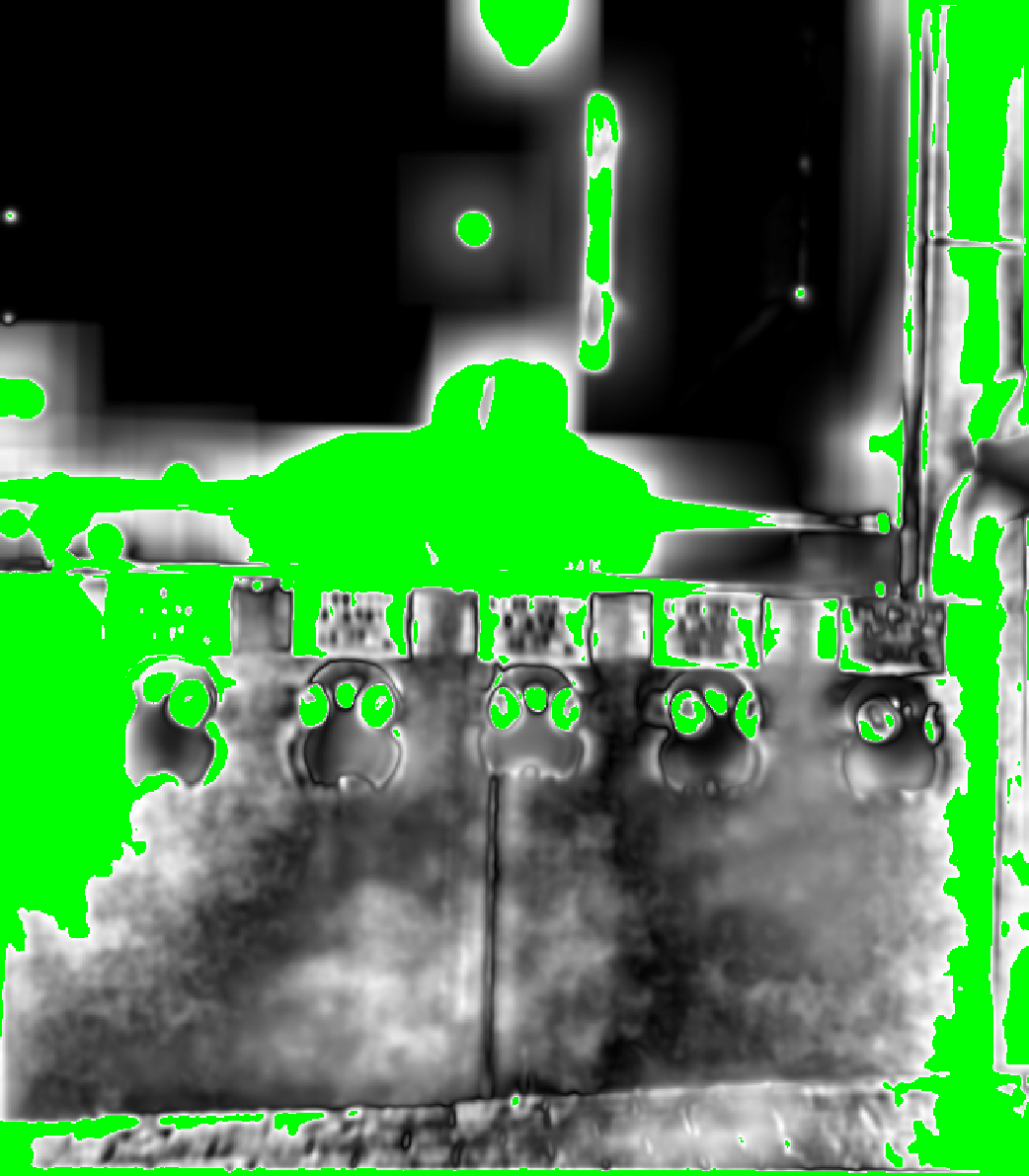}\\\vspace{1mm}
		{\includegraphics*[width = 0.19\linewidth]{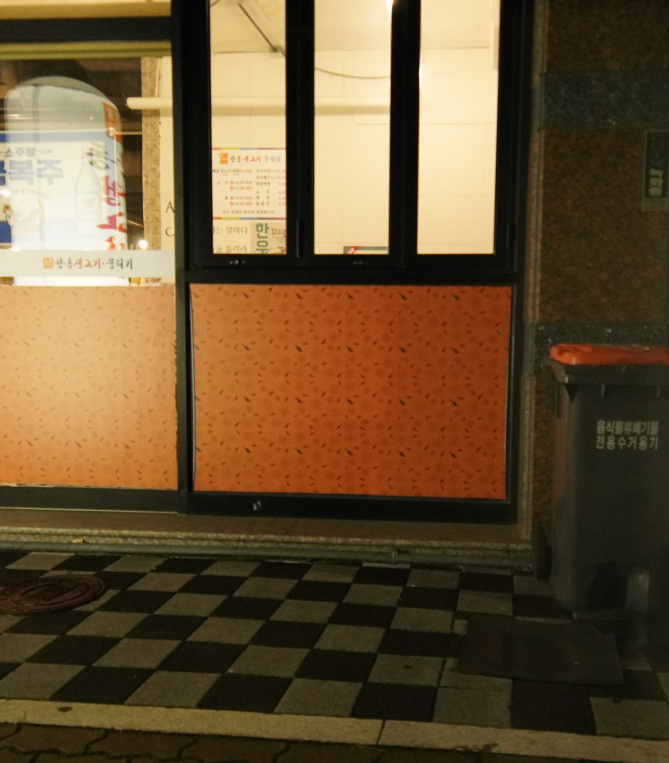}}
		{\includegraphics*[width = 0.19\linewidth]{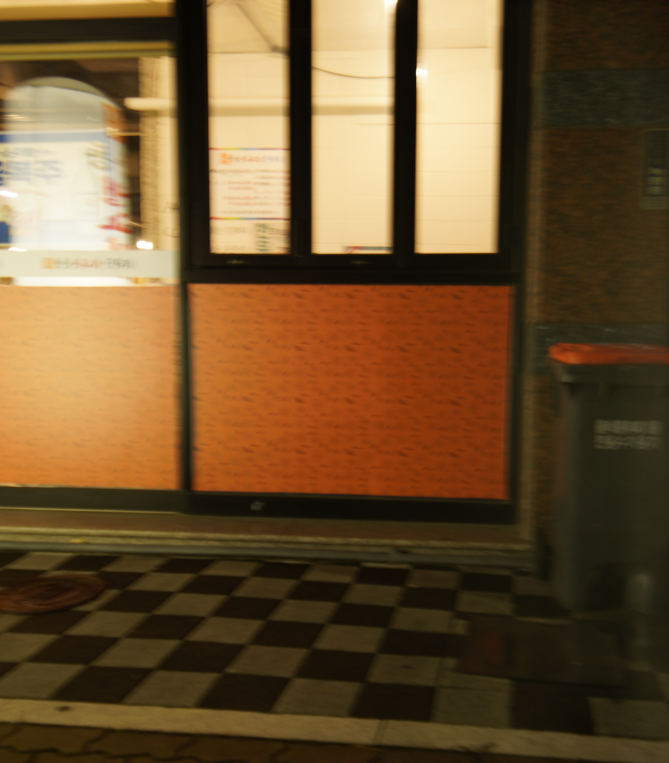}}
		{\includegraphics*[width = 0.19\linewidth]{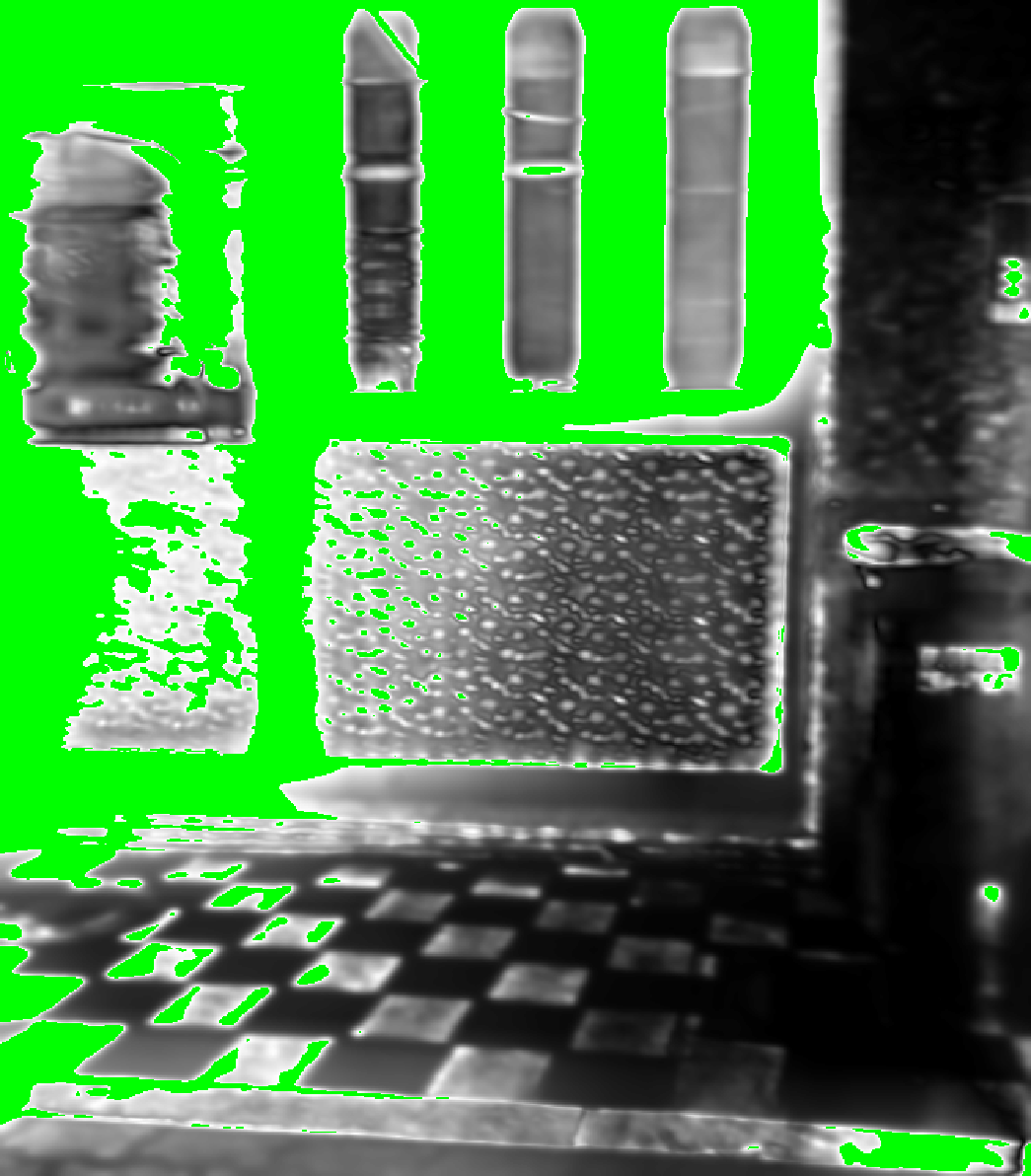}}
		{\includegraphics*[width = 0.19\linewidth]{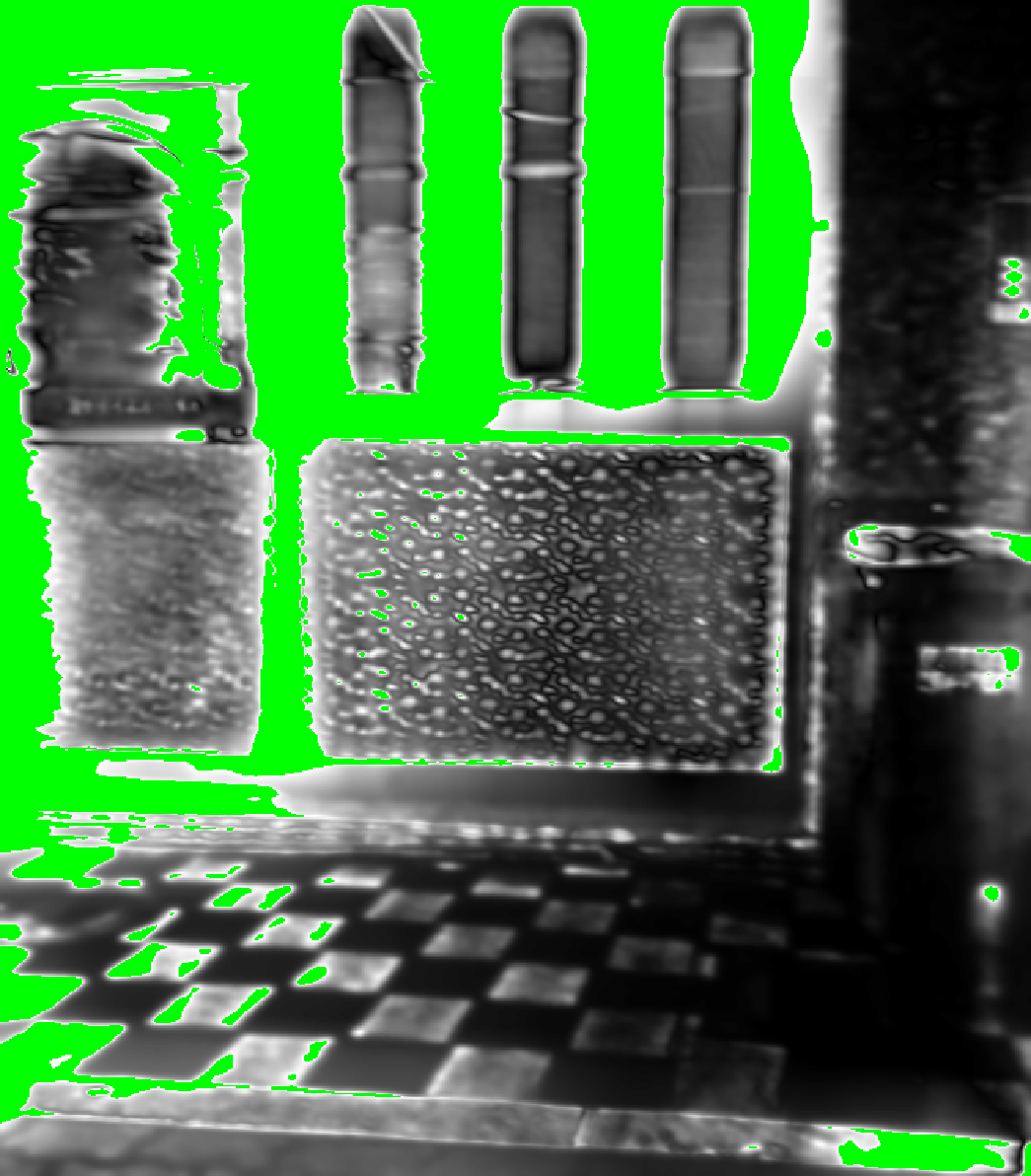}}
		{\includegraphics*[width = 0.19\linewidth]{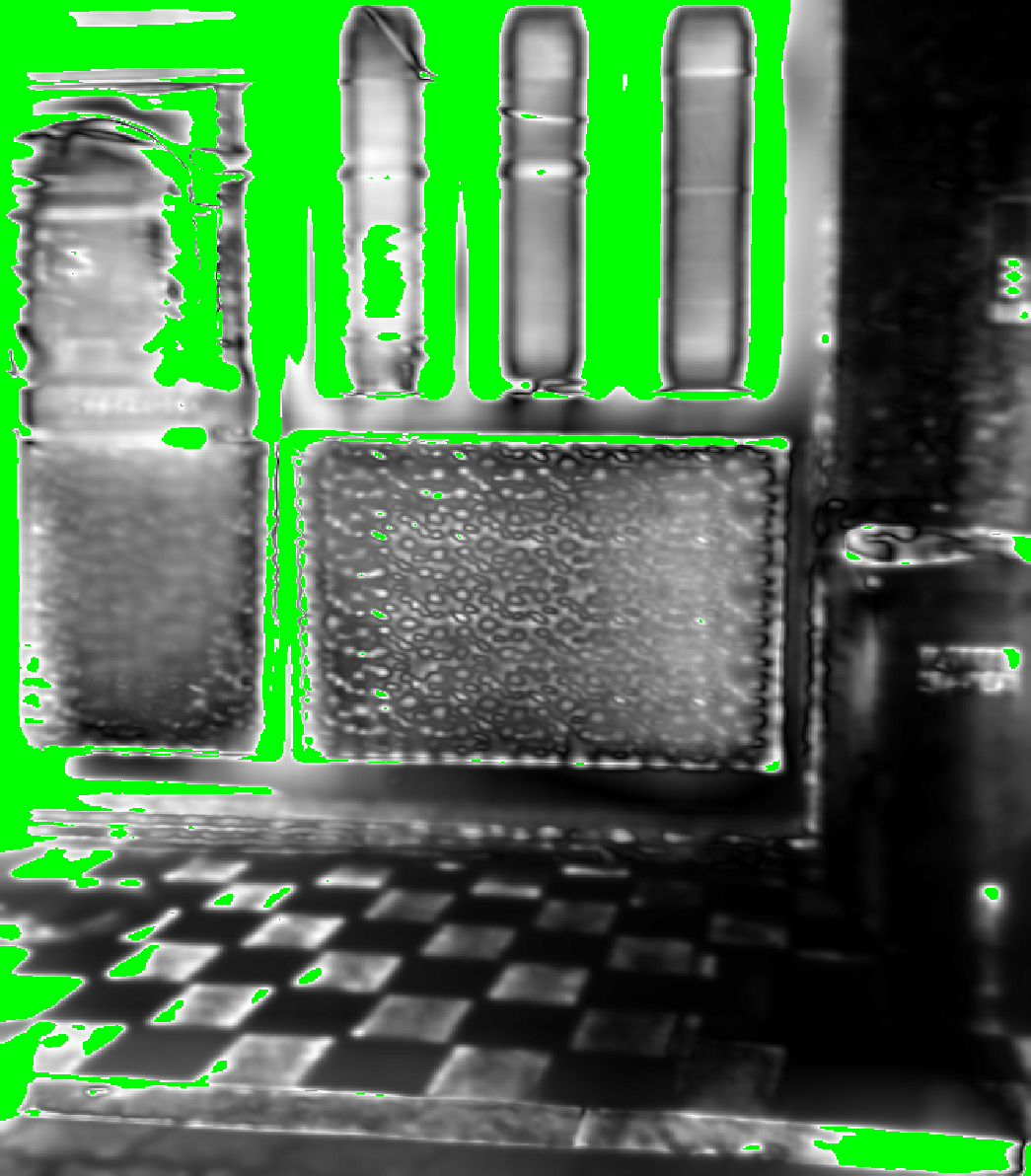}}\\\vspace{-4.5mm}
		\subfigure[sharp]{\includegraphics*[width = 0.19\linewidth,height = 0mm]{blank.png}}
		\subfigure[blurred]{\includegraphics*[width = 0.19\linewidth,height = 0mm]{blank.png}}
		\subfigure[CCK with normalized $k$]{\includegraphics*[width = 0.19\linewidth,height = 0mm]{blank.png}}
		\subfigure[QCK with     $\|\mathbf{Q}\|_1=1$]{\includegraphics*[width = 0.19\linewidth,height = 0mm]{blank.png}}
		\subfigure[QCK with (\ref{norma2})]{\includegraphics*[width = 0.19\linewidth,height = 0mm]{blank.png}}\\
		\caption{Four testing images from Rim's dataset and their spatial distributions of S-CIELAB error in the kernel experiments.
			The number of pixels with color errors exceeding 5 units are (233929, 204793, 118542), (65849, 50970, 23254), (231294, 188849, 101726), (142330, 110519, 84549) respectively from top to bottom.}
		\label{fig-inverse2}
	\end{figure}
	\begin{table}[t]
		\centering
		\caption{The average PSNR, SSIM and S-CIELAB color error values of 4738 pairs from Rim's dataset (second row), and the numbers of results achieving best measure values by using different settings (third row).}
		\resizebox{\linewidth}{!}{ 
			\begin{tabular}{|c| m{1.5cm}<{\centering}|m{1.5cm}<{\centering}| m{1.5cm}<{\centering}|}
				\hline
				Measure  & CCK  with normalized $k$  & QCK with     $\|\mathbf{Q}\|_1=1$ & QCK with (\ref{norma2}) \\ \hline
				PSNR     & 28.54  & 28.73                            & \textbf{29.72}                          \\ 
				SSIM     & 0.9200 & 0.9280                           & \textbf{0.9314}                         \\ 
				S-CIELAB & 159427  & 142183                            & \textbf{102707}                          \\ \hline
				PSNR     & 178 	&373 	&\textbf{4188} \\ 
				SSIM     & 231 	&798 	&\textbf{3710}    \\ 
				S-CIELAB & 204 	&537 	&\textbf{3997  }    \\ \hline
		\end{tabular}}
		\label{table-inverse2}
	\end{table}
		
	Since K{\"o}hler dataset was collected in a fully controlled laboratory environment, we employ the dataset introduced by Rim et al. \cite{real2020} which is composed of blurred images of the real world to further test the role of the proposed normalized quaternion convolution kernels. This dataset utilizes a beam splitter and two cameras with slow and high-speed shutters to simultaneously capture blurred and sharp image pairs, and contains 4738 blurry+real image pairs. In this experiment, we make use of the same strategy to conduct the kernel testing. 
		In Table \ref{table-inverse2}, we report the average PSNR, SSIM and S-CIELAB color error values corresponding to 4738 pairs in Rim's dataset, and 
	we also report the numbers of results achieving best measure values by using different settings.
	In Figure \ref{fig-inverse2}, we show four results obtained based on four testing pairs in Rim's dataset. We also display their spatial distributions of S-CIELAB error.
			The number of pixels with color errors exceeding 5 units are given in the caption of Figure \ref{fig-inverse2}.
	It is clear that the proposed model by using normalized quaternion convolution kernels 
	has the better ability to accurately characterize blurs of color images in blind deconvolution problem.

	\subsection{Testing experiments on K{\"o}hler dataset}	
	\begin{figure}[b]
		\centering
		\includegraphics*[width = 0.9\linewidth]{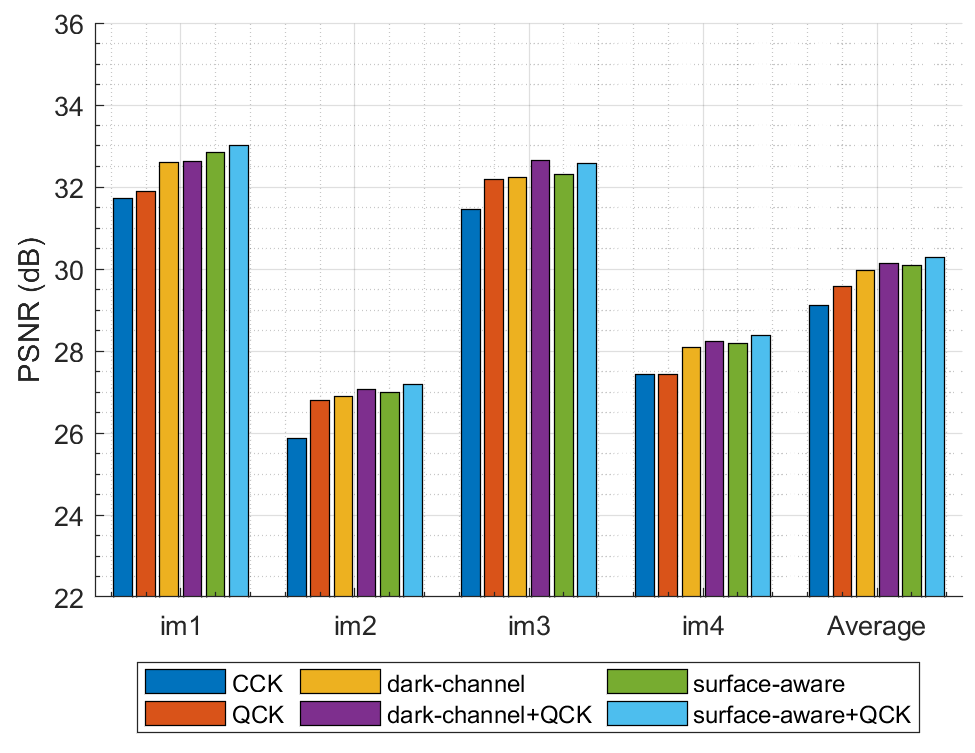}
		\caption{Quantitative evaluations on the testing images from K{\"o}hler dataset by using different methods.}
		\label{psnr-kohler}
	\end{figure}
\begin{figure}[t]
		\centering
		\subfigure[blurred image]{\includegraphics*[width = 0.4\linewidth]{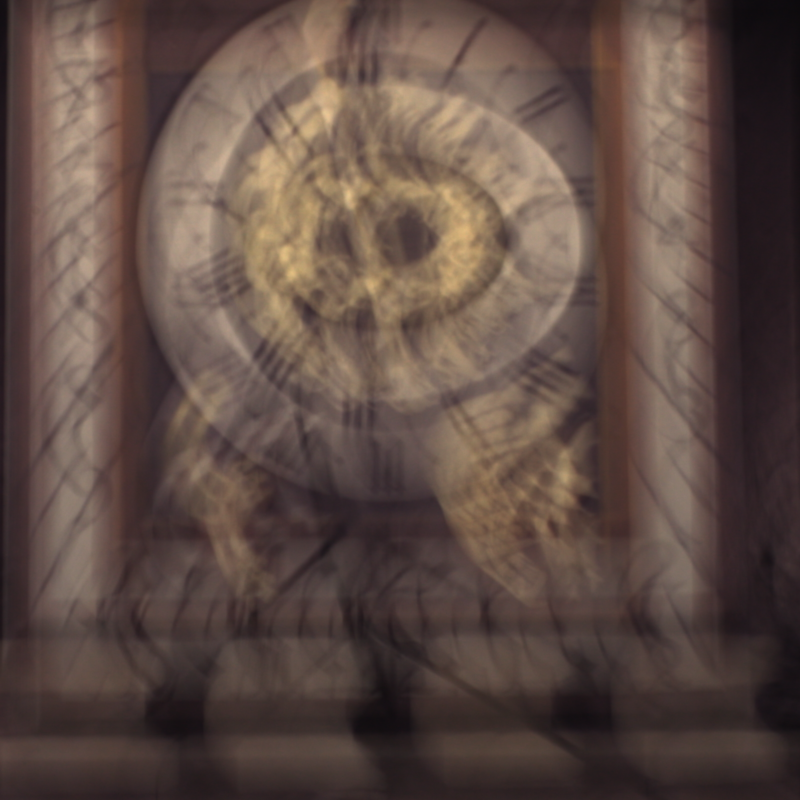}}
		\subfigure[sharp image]{\includegraphics*[width = 0.4\linewidth]{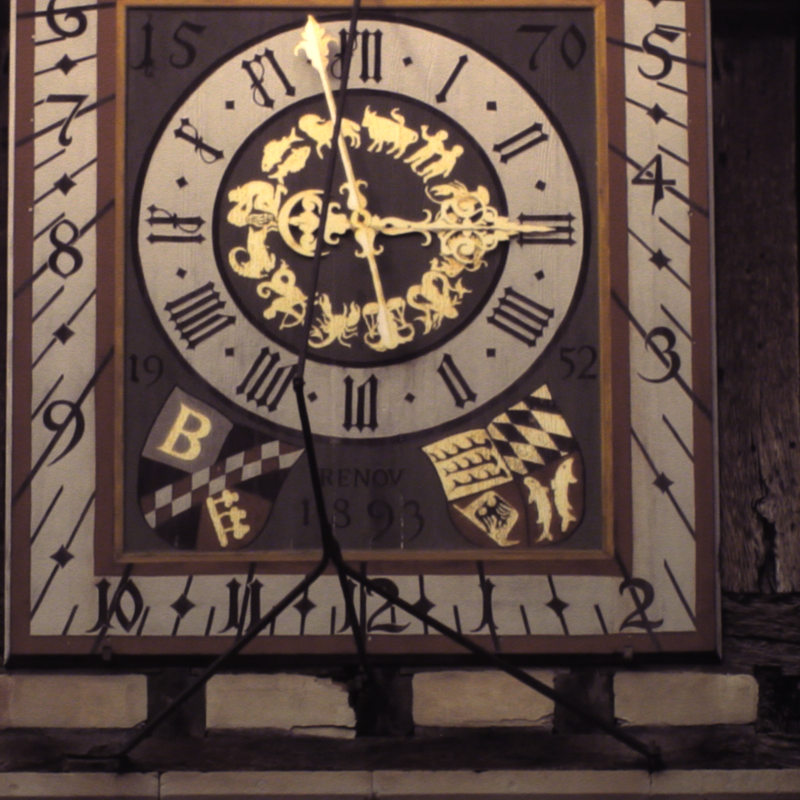}}\\
		\subfigure[CCK]{\includegraphics*[width = 0.4\linewidth]{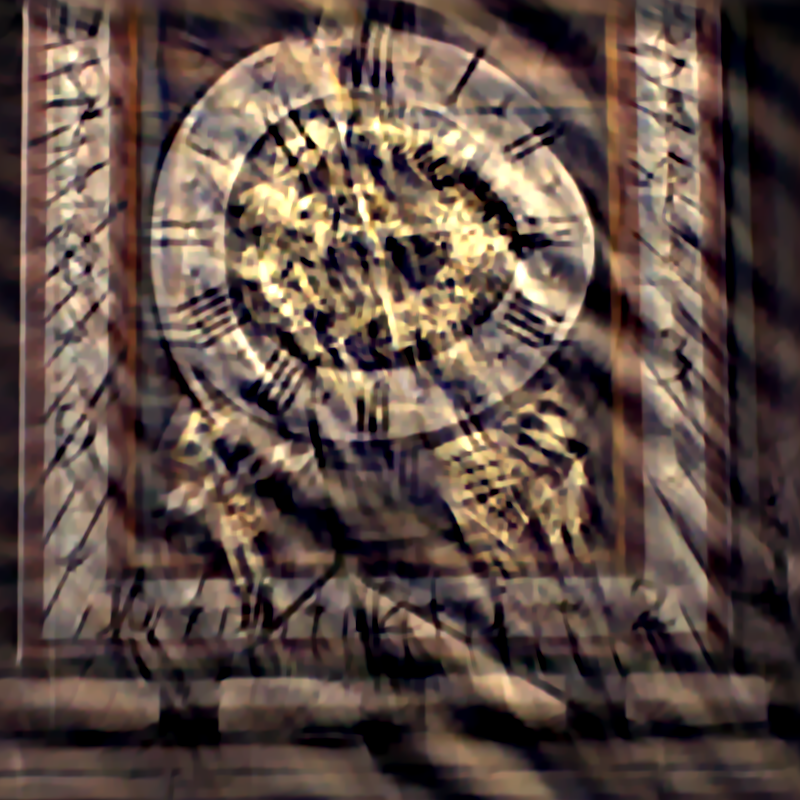}}
		\subfigure[QCK]{\includegraphics*[width = 0.4\linewidth]{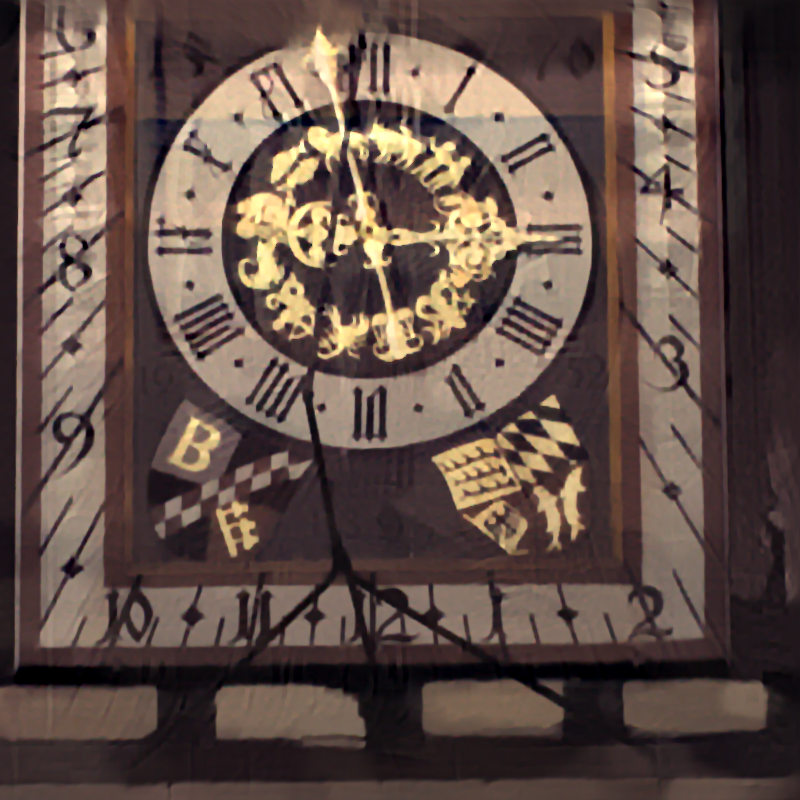}}\\
		\subfigure[dark-channel prior \cite{DCP2018}]{\includegraphics*[width = 0.4\linewidth]{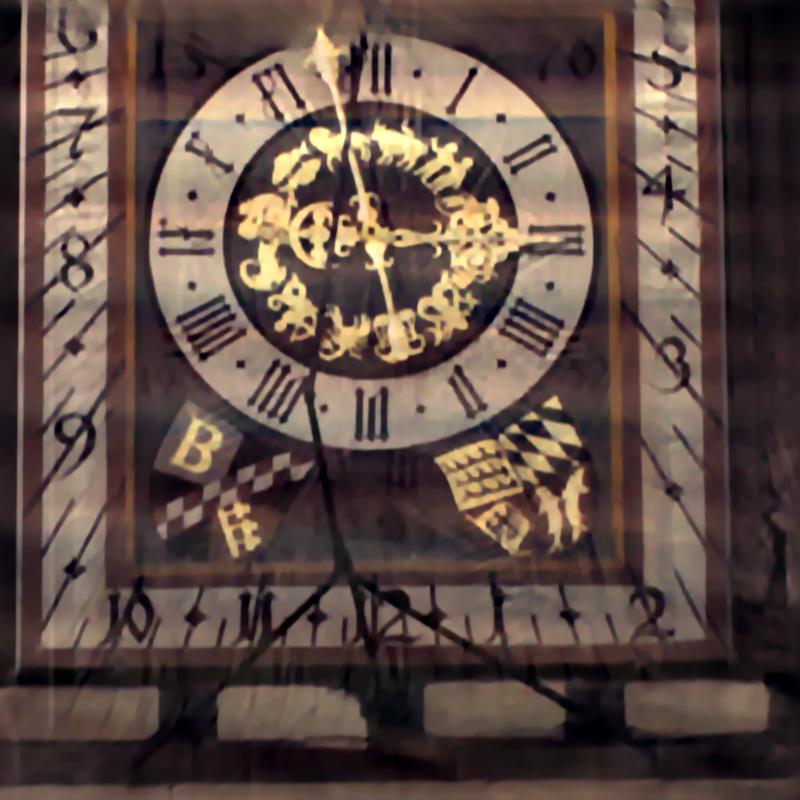}}
		\subfigure[dark-channel prior+QCK]{\includegraphics*[width = 0.4\linewidth]{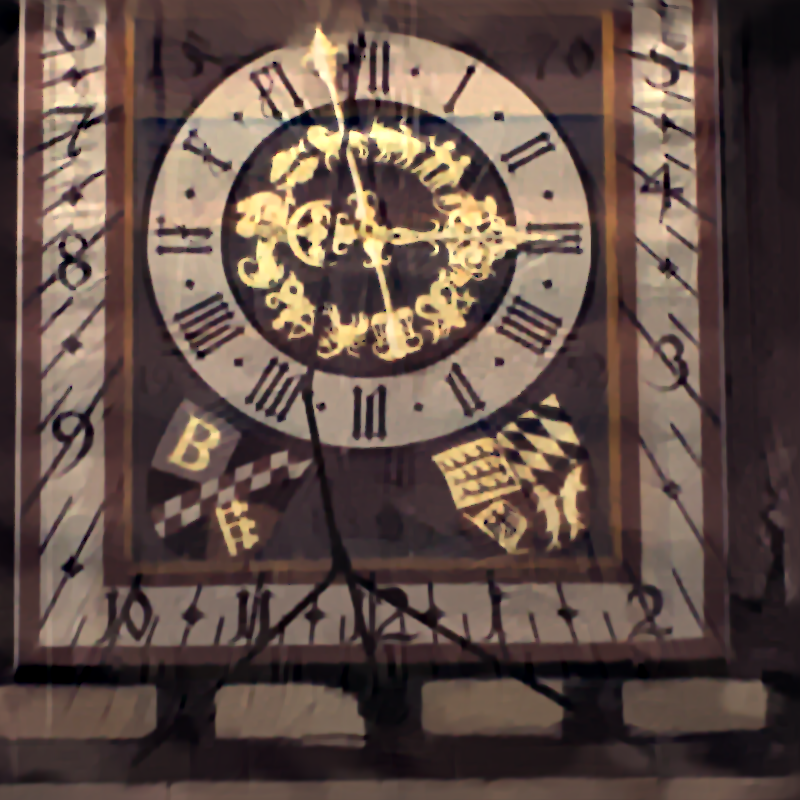}}
		\subfigure[surface-aware prior \cite{SA2021}]{\includegraphics*[width = 0.4\linewidth]{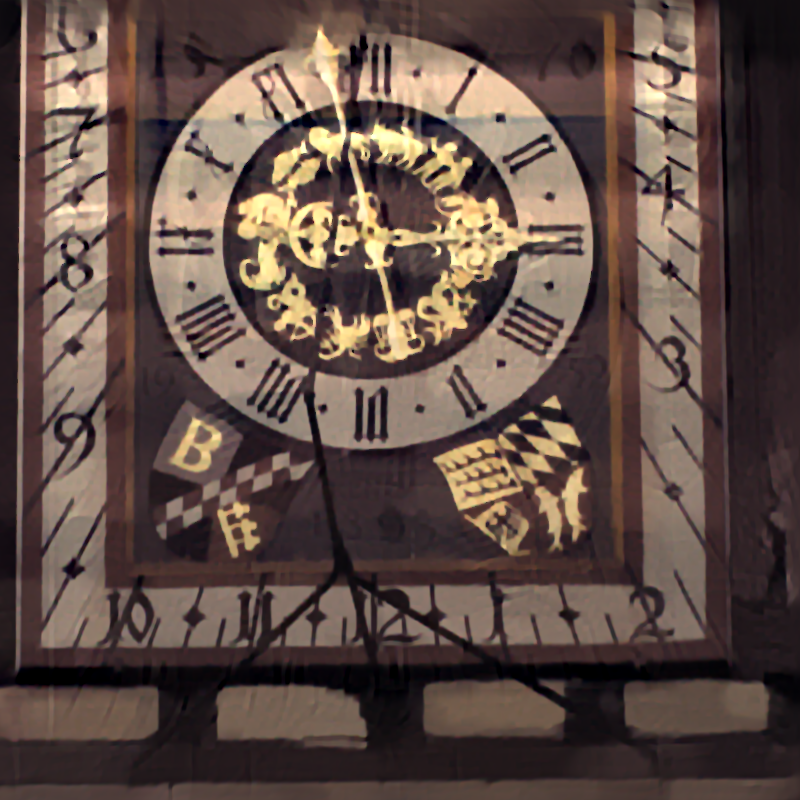}}
		\subfigure[surface-aware prior+QCK]{\includegraphics*[width = 0.4\linewidth]{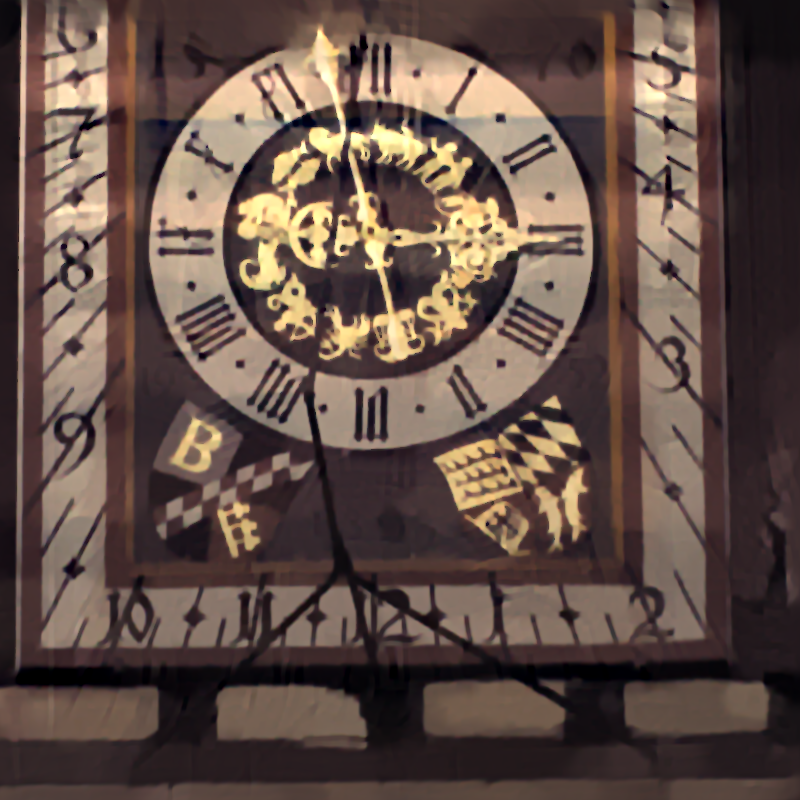}}\\
		\caption{A challenging example from K{\"o}hler dataset.
			(a) the blurred image; (b) the sharp image;
			(c) the restored result by using CCK;
			(d) the restored result by using QCK;
			(e) the restored result by using dark-channel prior;
			(f) the restored result by using dark-channel prior+QCK;			       					       
			(g) the restored result by using surface-aware prior;
			(h) the restored result by using surface-aware prior+QCK.}
		\label{fig-kohler1}
	\end{figure}
\begin{figure}[t]
		\centering
		\subfigure[blurred image]{\includegraphics*[width = 0.4\linewidth]{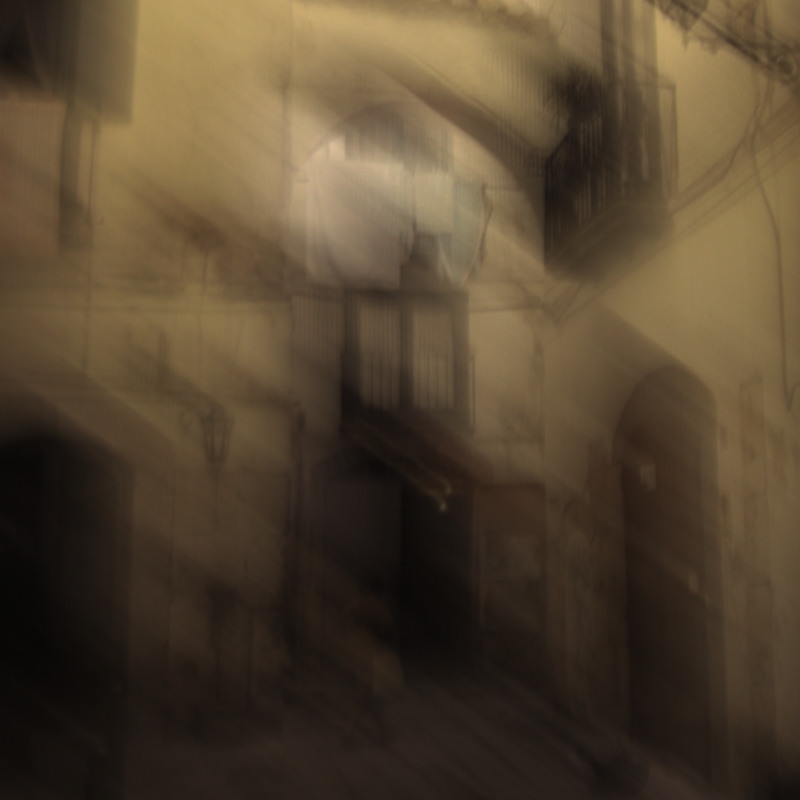}}
		\subfigure[sharp image]{\includegraphics*[width = 0.4\linewidth]{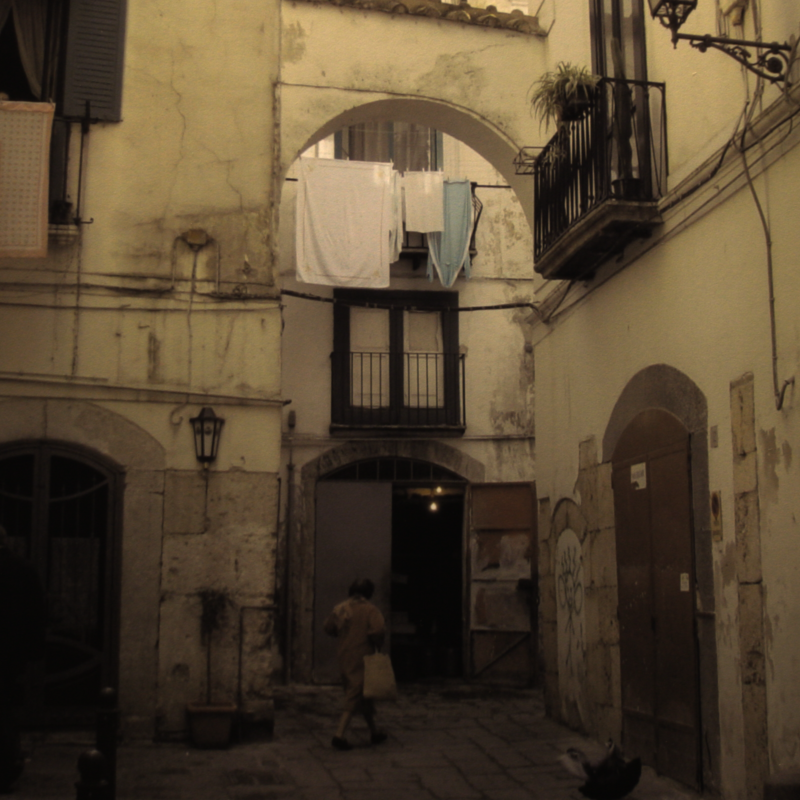}}\\
		\subfigure[CCK]{\includegraphics*[width = 0.4\linewidth]{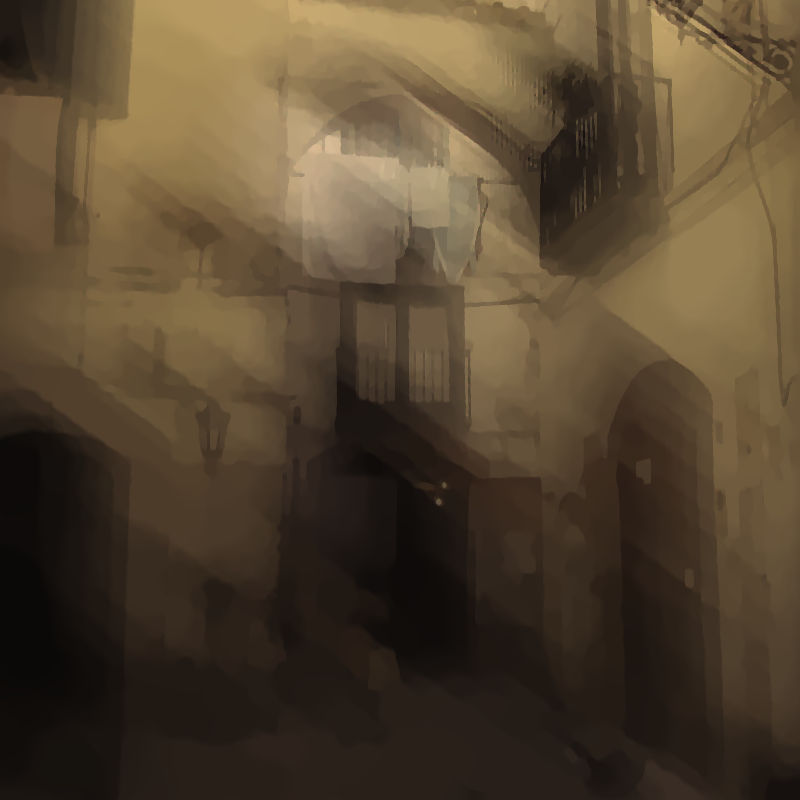}}
		\subfigure[QCK]{\includegraphics*[width = 0.4\linewidth]{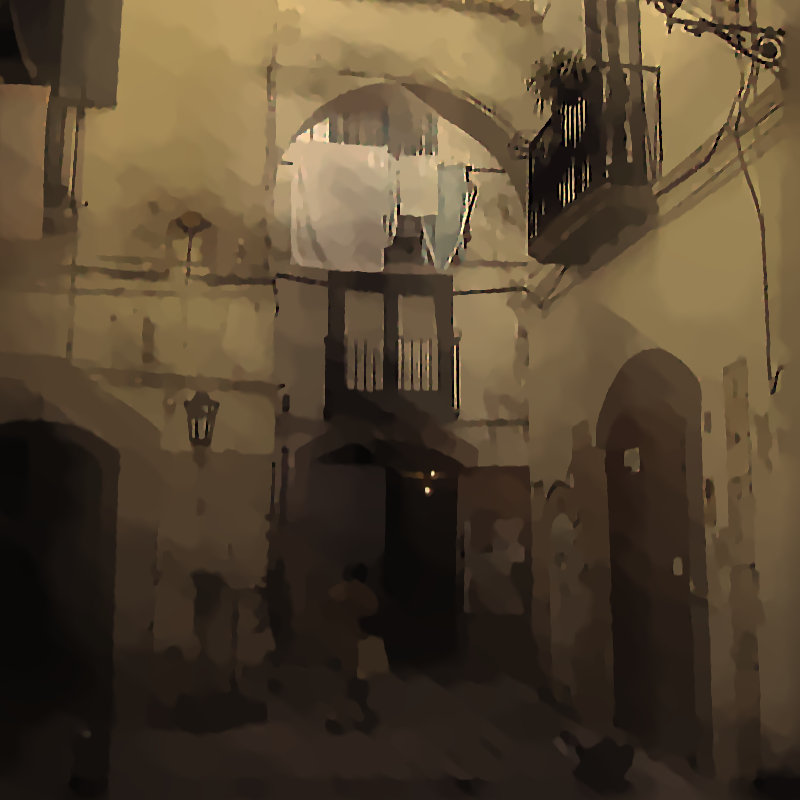}}\\
		\subfigure[dark-channel prior \cite{DCP2018}]{\includegraphics*[width = 0.4\linewidth]{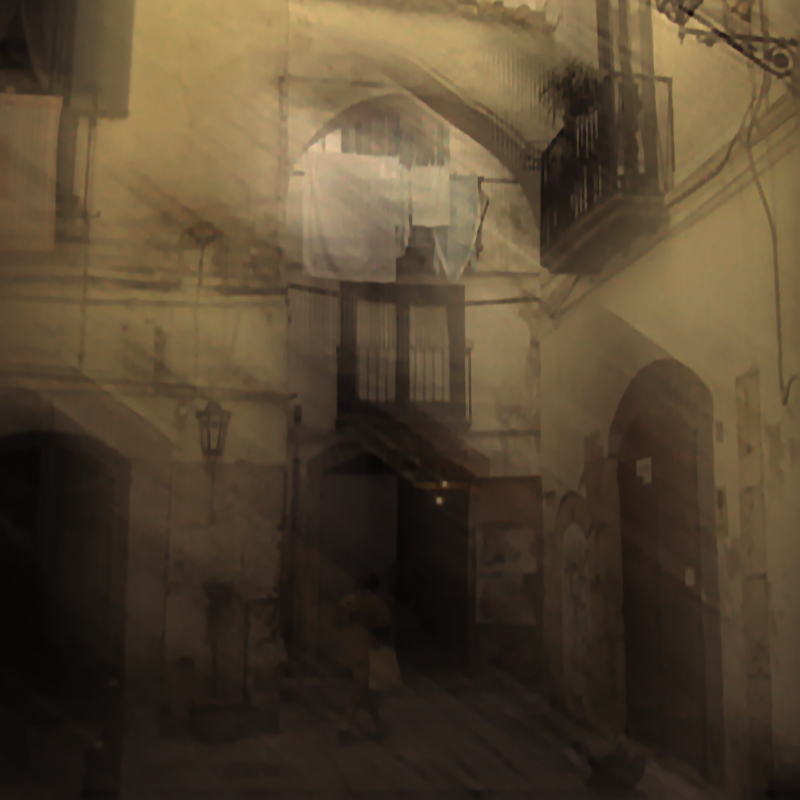}}
		\subfigure[dark-channel prior+QCK]{\includegraphics*[width = 0.4\linewidth]{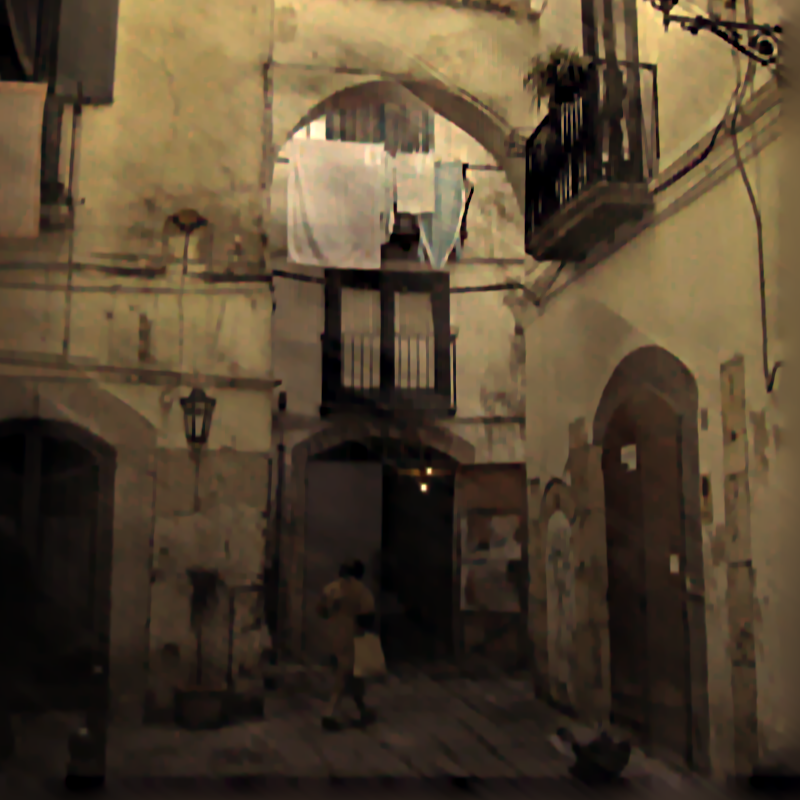}}
		\subfigure[surface-aware prior \cite{SA2021}]{\includegraphics*[width = 0.4\linewidth]{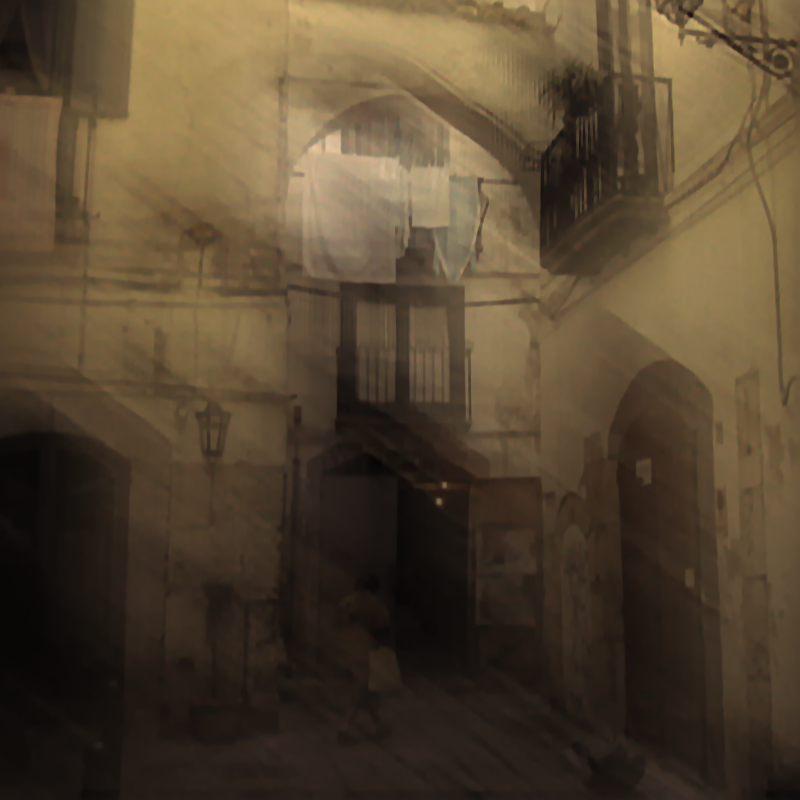}}
		\subfigure[surface-aware prior+QCK]{\includegraphics*[width = 0.4\linewidth]{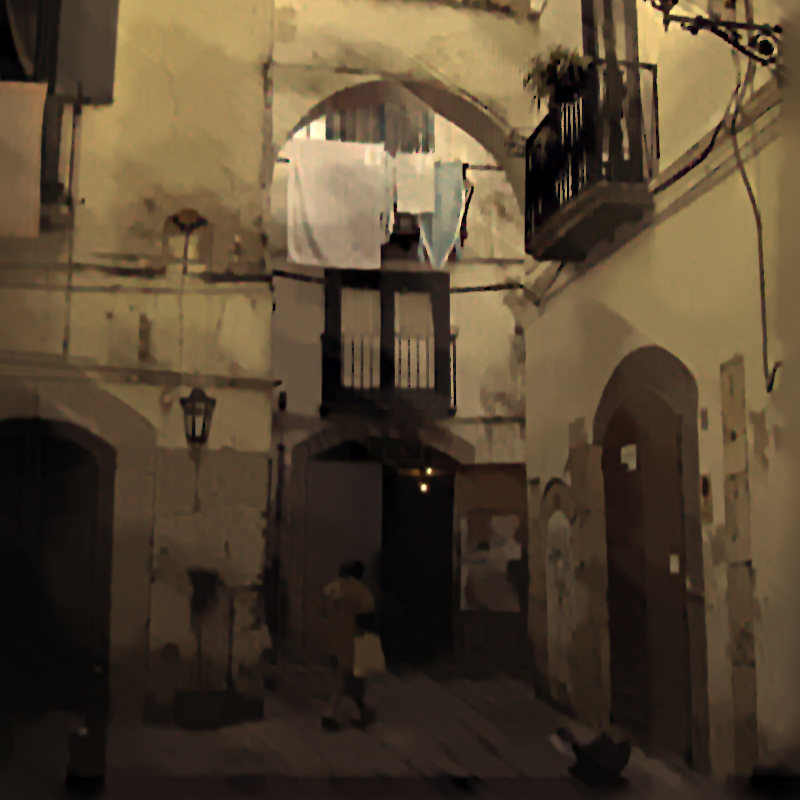}}\\
		\caption{A challenging example from K{\"o}hler dataset.
			(a) the blurred image; (b) the sharp image;
			(c) the restored result by using CCK;
			(d) the restored result by using QCK;
			(e) the restored result by using dark-channel prior;
			(f) the restored result by using dark-channel prior+QCK;			       					       
			(g) the restored result by using surface-aware prior;
			(h) the restored result by using surface-aware prior+QCK.}
		\label{fig-kohler2}
	\end{figure}

	Noting that non-blind model usually recovers finer details and yields superior quantitative scores, we derive the restored results
	by feeding the kernel estimated by the blind deconvolution model into the non-blind model in this experiment. We still consider K{\"o}hler dataset \cite{2012kholer} in this subsection.
	In Figure \ref{fig-kohler1} and Figure \ref{fig-kohler2}, we present two challenging examples with significant blurs in K{\"o}hler dataset. 
	We see from the figures that the results by using CCK fail to deblur the inputs, while the results by using QCK achieve very good deblurring effect. 
	In Figure \ref{fig-kohler2}, the original sharp image is degraded by a long and unidirectional motion blur. The small-magnitude tails are effectively suppressed and thus lost in the CCK model.
	In contrast, the QCK model preserves these weak components, resulting in more accurate restoration.	
	In Figure \ref{psnr-kohler}, we show the average PSNR values by using different methods of the 12 deblurring results corresponding to each image, also the overall average values of all 48 deblurring results.		
	It can be seen that the PSNR values are improved due to the proposed normalized quaternion convolution kernel.

	\begin{figure*}[t]
		\centering
		\includegraphics*[width = 0.137\linewidth]{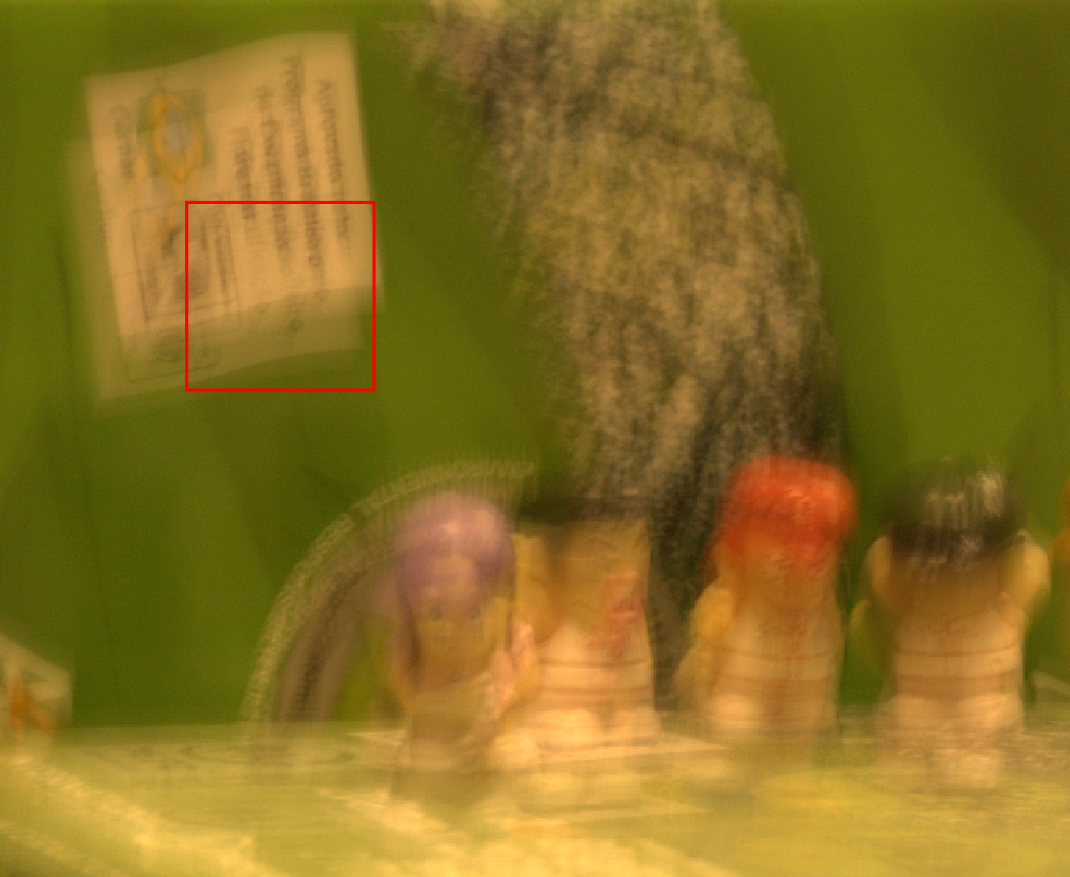}
		\includegraphics*[width = 0.137\linewidth]{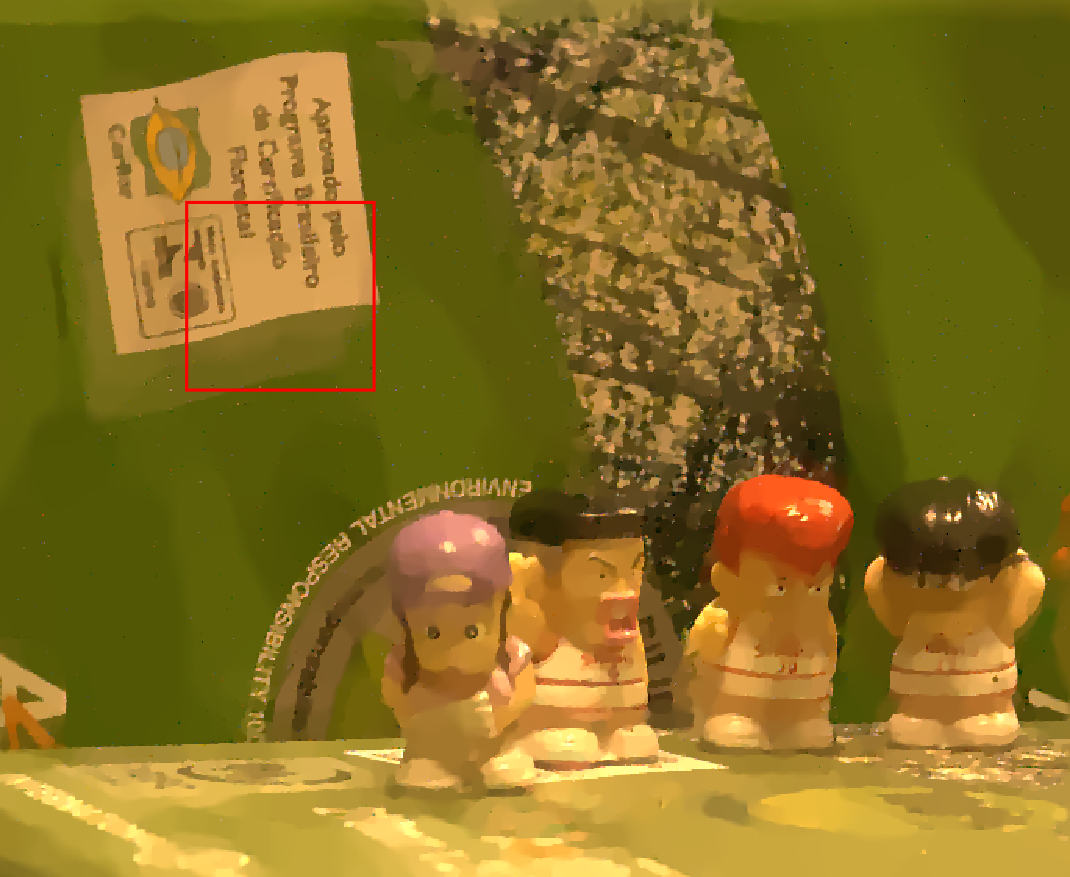}
		\includegraphics*[width = 0.137\linewidth]{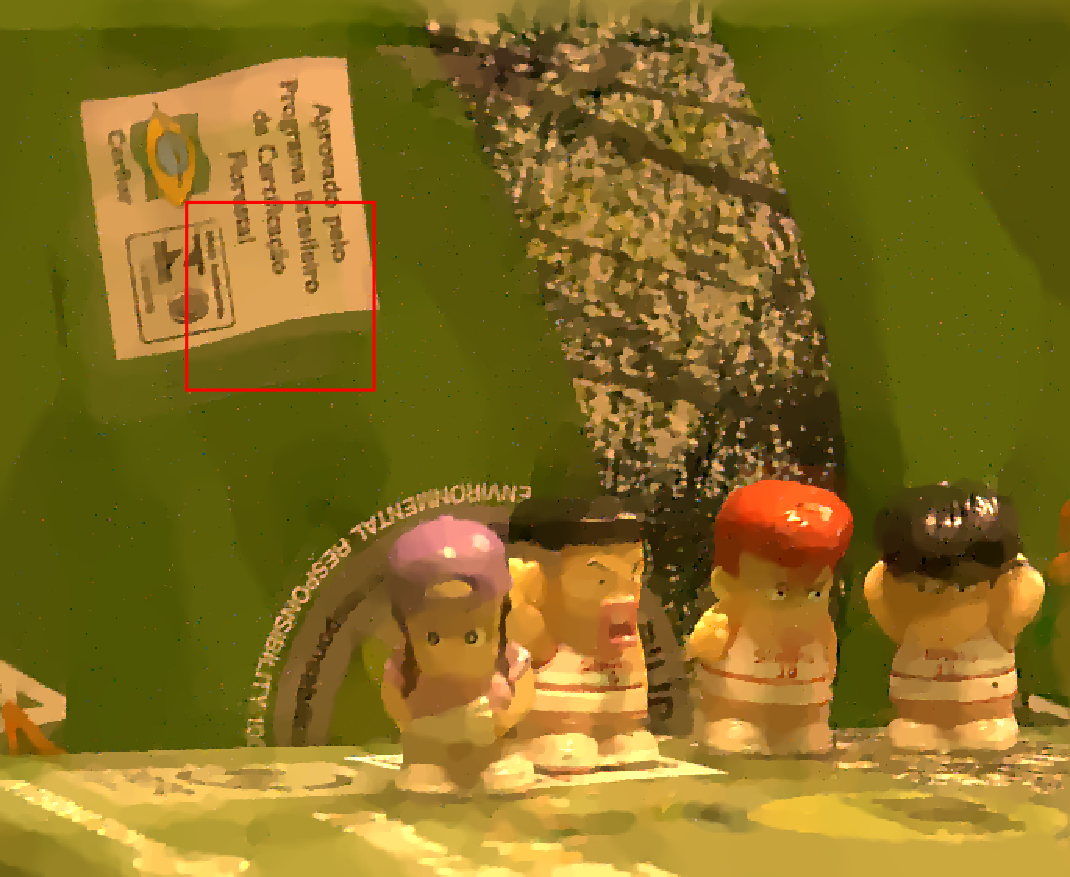}
		\includegraphics*[width = 0.137\linewidth]{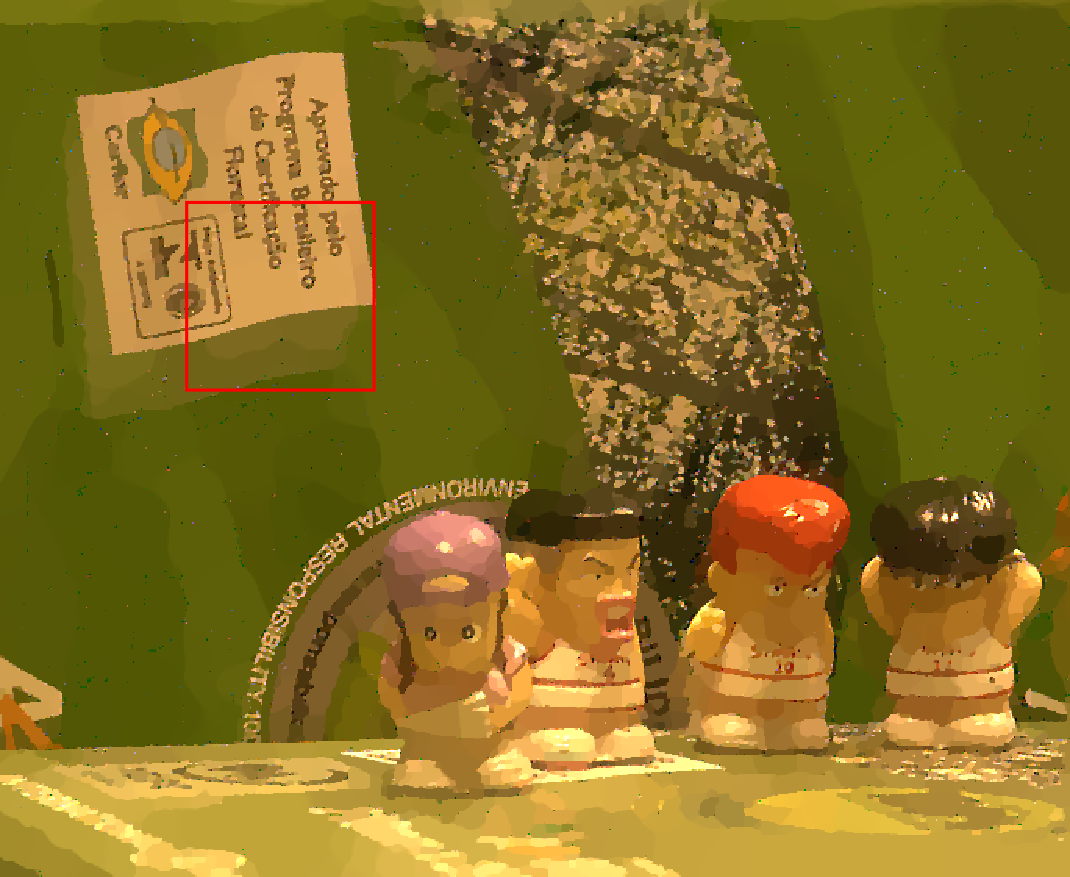}
		\includegraphics*[width = 0.137\linewidth]{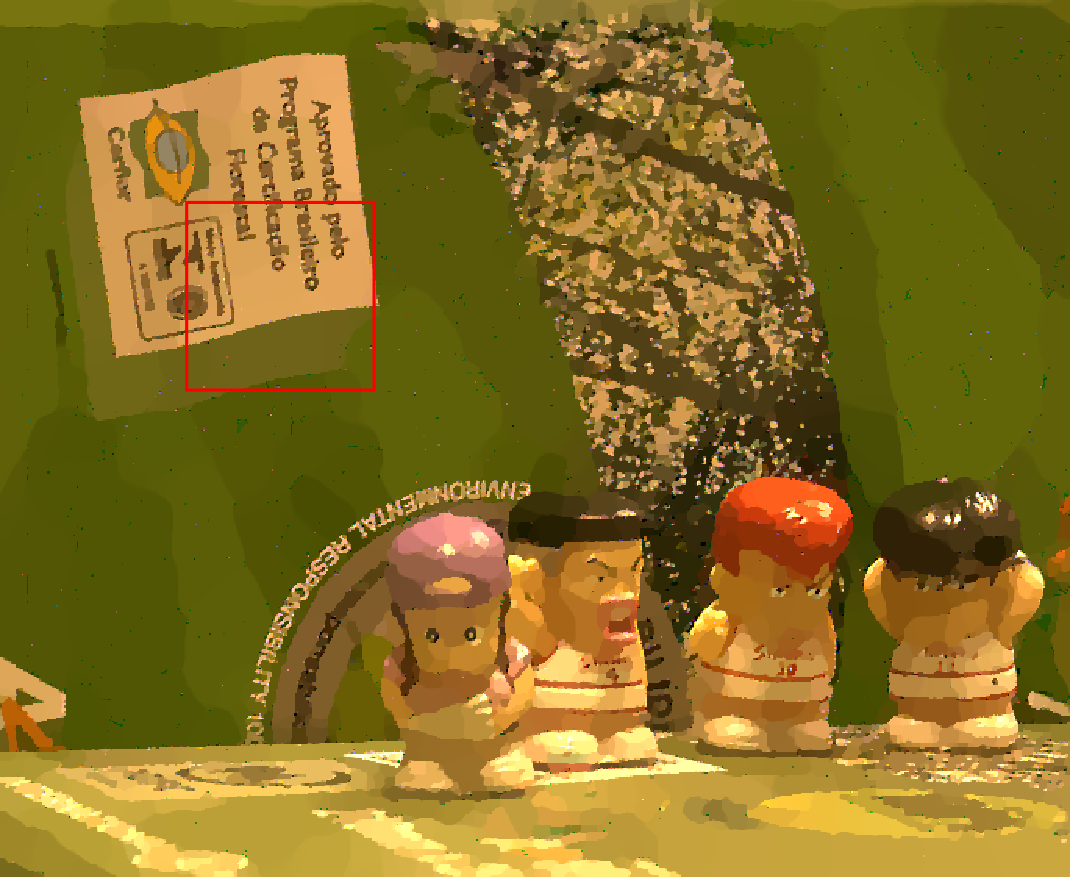}
		\includegraphics*[width = 0.137\linewidth]{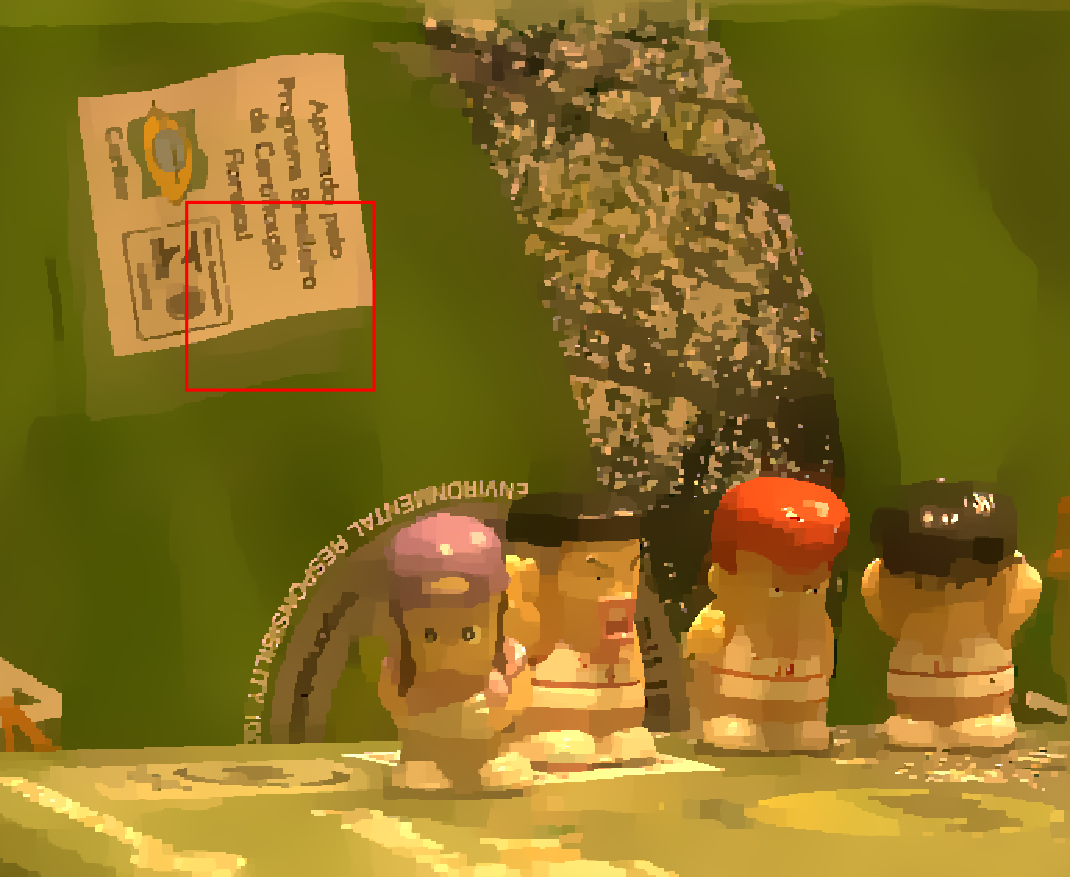}
		\includegraphics*[width = 0.137\linewidth]{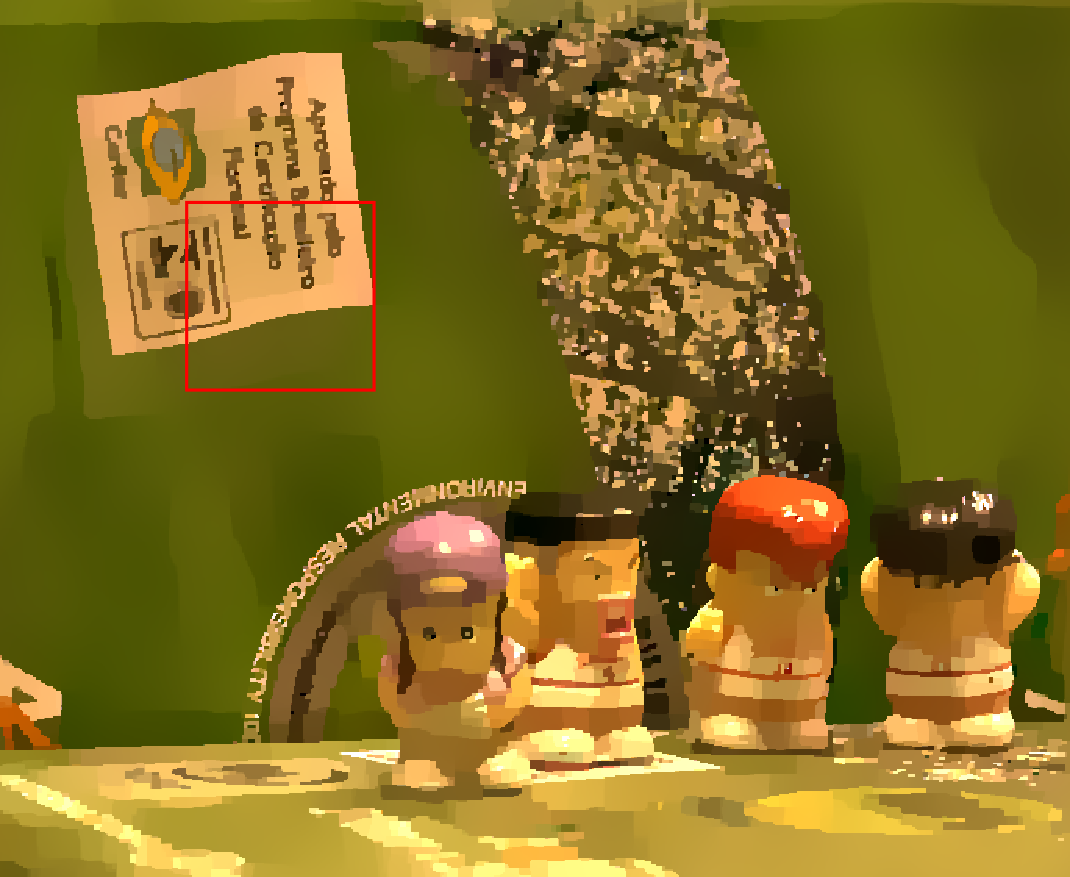}
		\\ \vspace{2mm}
		\includegraphics*[width = 0.137\linewidth]{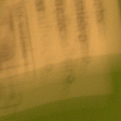}
		\includegraphics*[width = 0.137\linewidth]{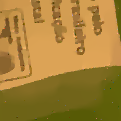}
		\includegraphics*[width = 0.137\linewidth]{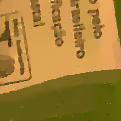}
		\includegraphics*[width = 0.137\linewidth]{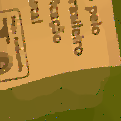}
		\includegraphics*[width = 0.137\linewidth]{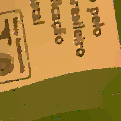}
		\includegraphics*[width = 0.137\linewidth]{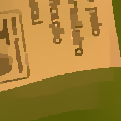}
		\includegraphics*[width = 0.137\linewidth]{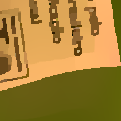}
		\\ \vspace{2mm}
		\includegraphics*[width = 0.137\linewidth]{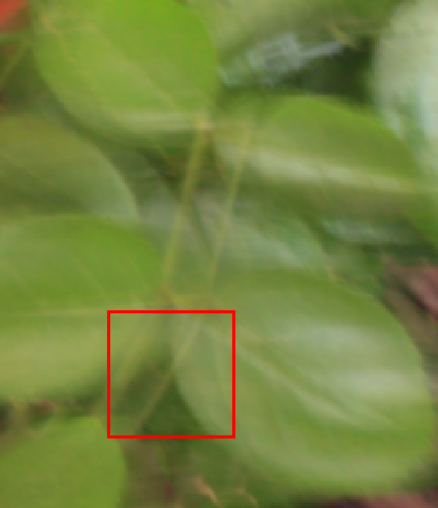}
		\includegraphics*[width = 0.137\linewidth]{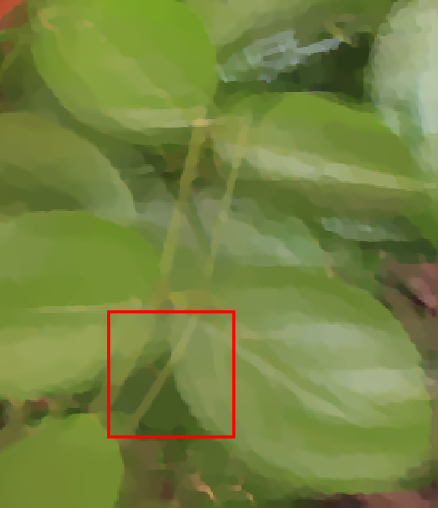}
		\includegraphics*[width = 0.137\linewidth]{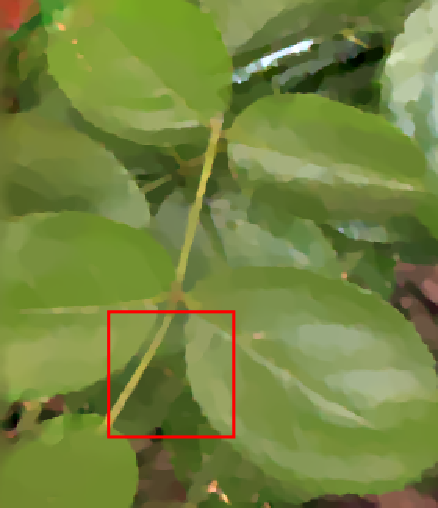}
		\includegraphics*[width = 0.137\linewidth]{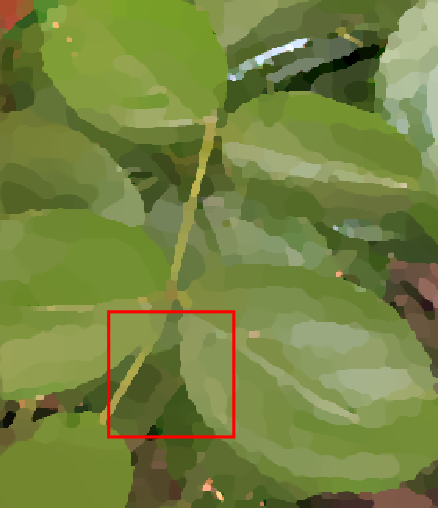}
		\includegraphics*[width = 0.137\linewidth]{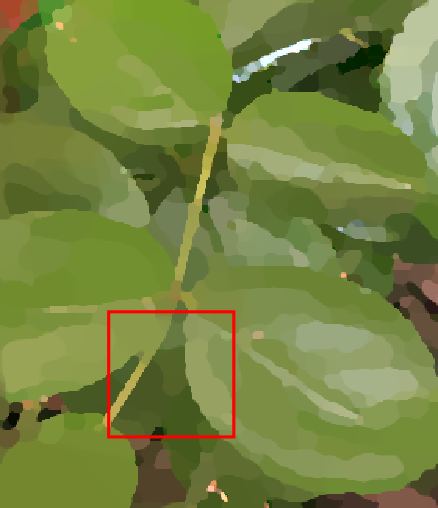}
		\includegraphics*[width = 0.137\linewidth]{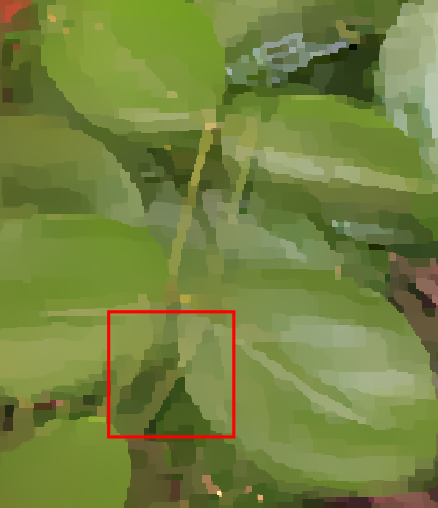}
		\includegraphics*[width = 0.137\linewidth]{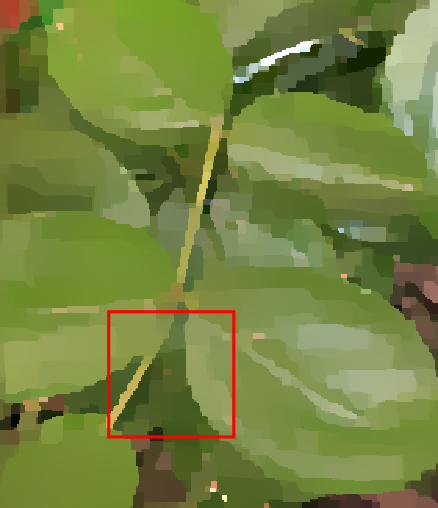}
		\\
		\subfigure[blurred image]{\includegraphics*[width = 0.137\linewidth]{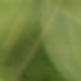}}
		\subfigure[CCK]{\includegraphics*[width = 0.137\linewidth]{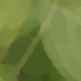}}
		\subfigure[QCK]{\includegraphics*[width = 0.137\linewidth]{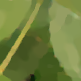}}
		\subfigure[{\scriptsize dark-channel \cite{DCP2018}}]{\includegraphics*[width = 0.137\linewidth]{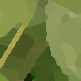}}
		\subfigure[{\scriptsize dark-channel+QCK}]{\includegraphics*[width = 0.137\linewidth]{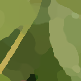}}
		\subfigure[{\scriptsize surface-aware \cite{SA2021}}]{\includegraphics*[width = 0.137\linewidth]{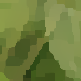}}
		\subfigure[{\scriptsize surface-aware+QCK}]{\includegraphics*[width = 0.137\linewidth]{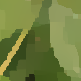}}\\
		\caption{(a) the blurred image; 
			(b) the restored result by using CCK;
			(c) the restored result by using QCK;
			(d) the restored result by using dark-channel prior;
			(e) the restored result by using dark-channel prior+QCK;			       					       
			(f) the restored result by using surface-aware prior;
			(g) the restored result by using surface-aware prior+QCK.
		}
		\label{fig-blind}
	\end{figure*}

	\subsection{Testing experiments on real blurred color images}

		\begin{figure*}[!t]
		\centering
		\includegraphics*[width = 0.137\linewidth]{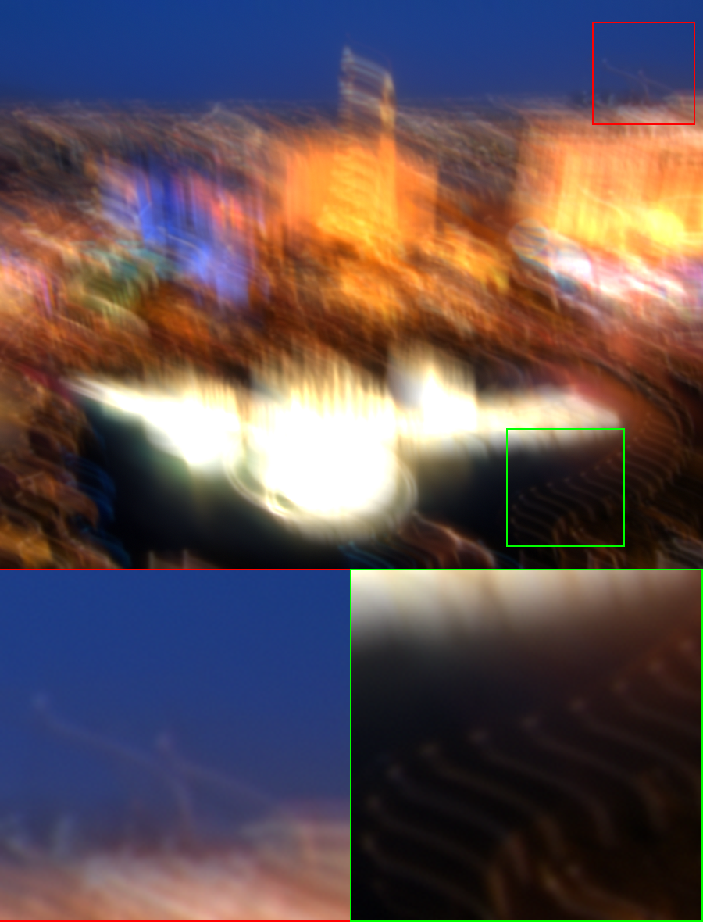}
		\includegraphics*[width = 0.137\linewidth]{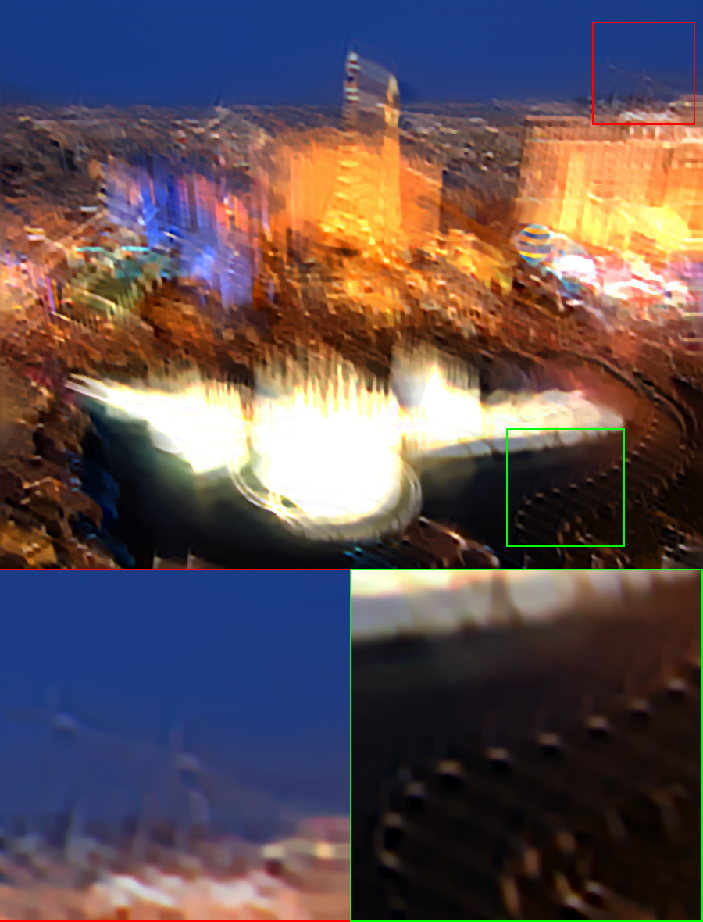}
		\includegraphics*[width = 0.137\linewidth]{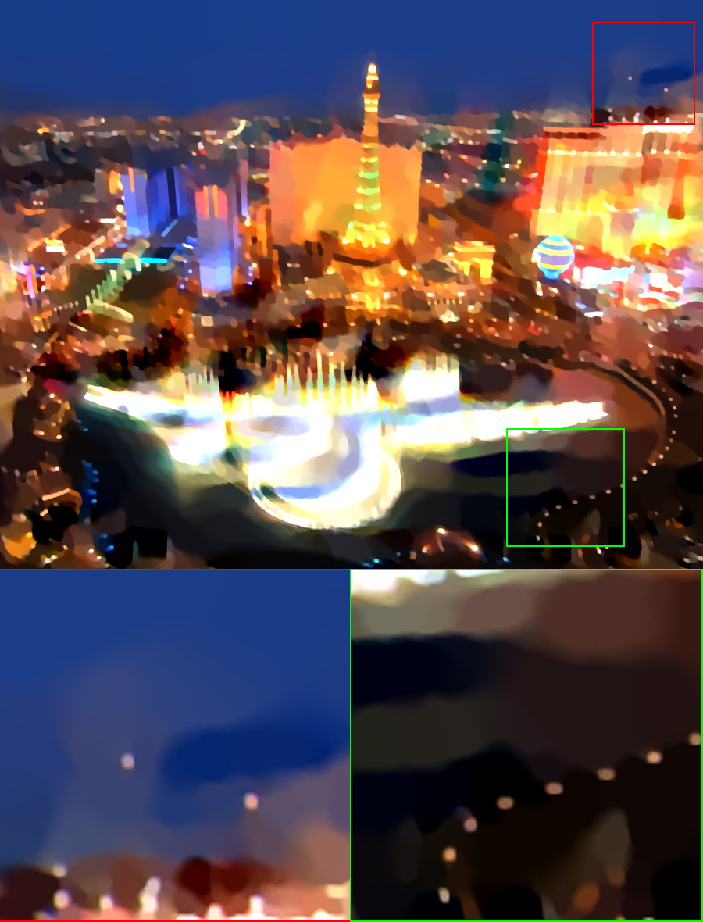}
		\includegraphics*[width = 0.137\linewidth]{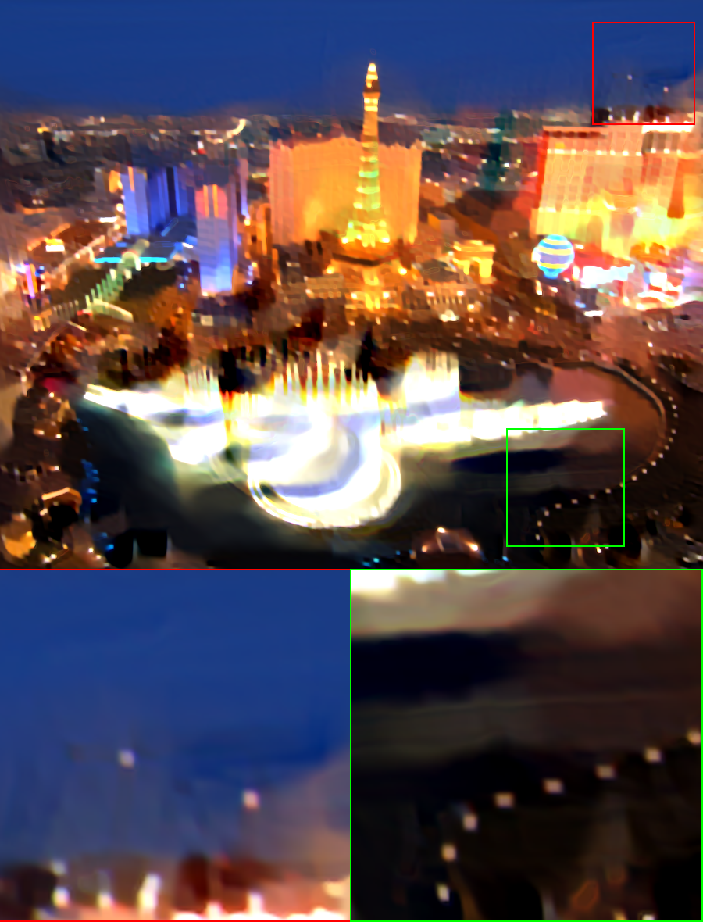}
		\includegraphics*[width = 0.137\linewidth]{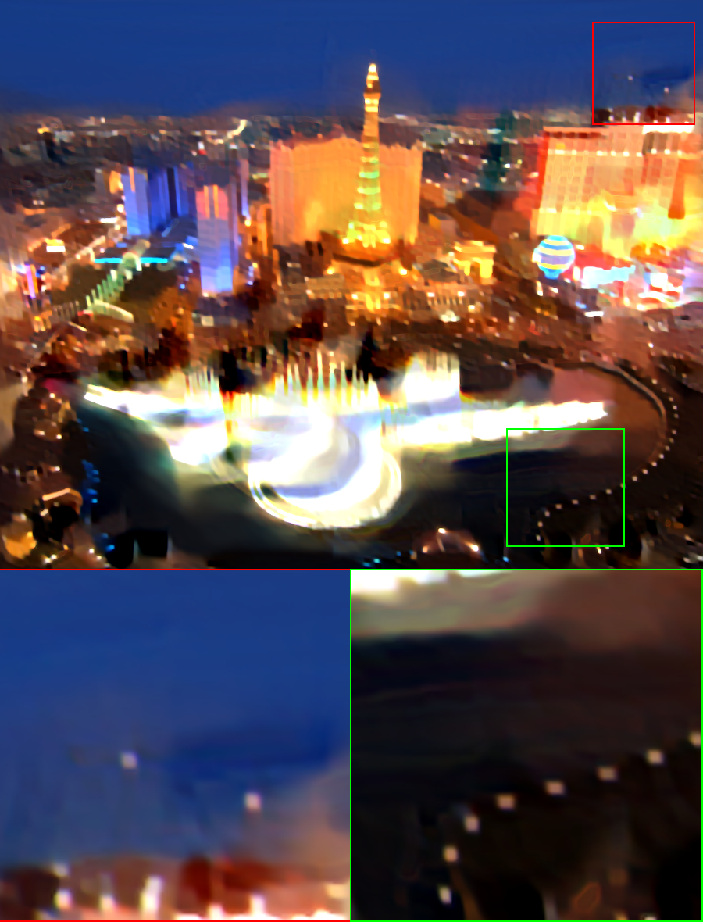}
		\includegraphics*[width = 0.137\linewidth]{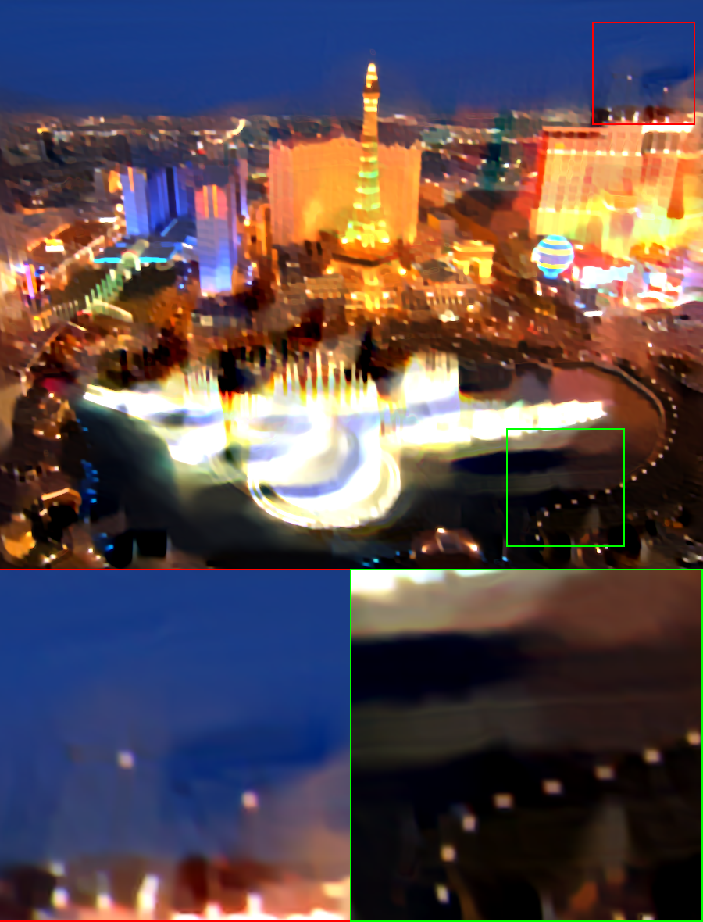}
		\includegraphics*[width = 0.137\linewidth]{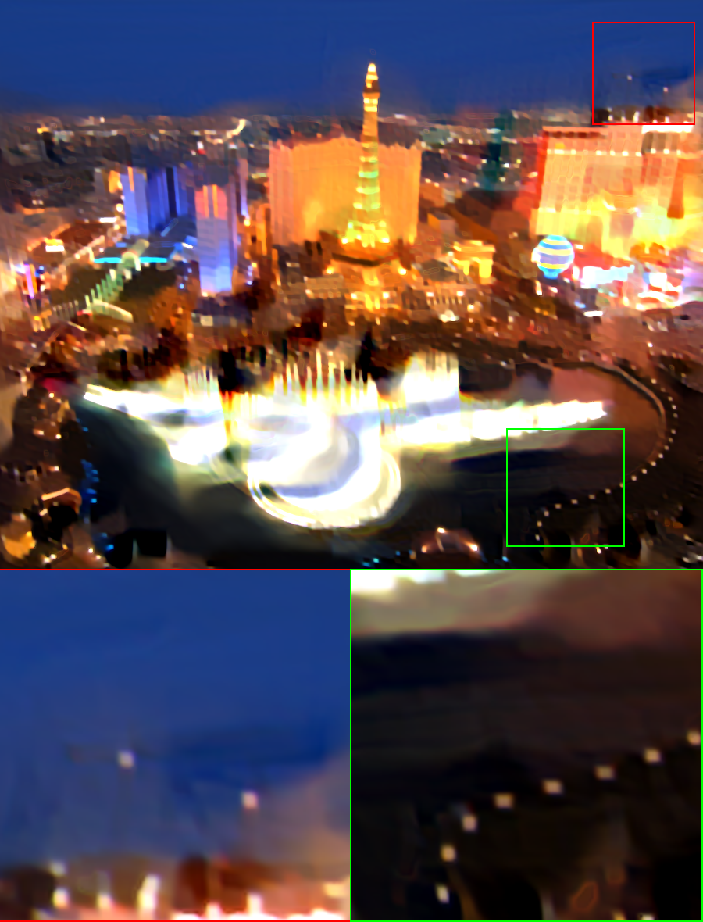}
		\\
		\subfigure[blurred image]{\includegraphics*[width = 0.137\linewidth]{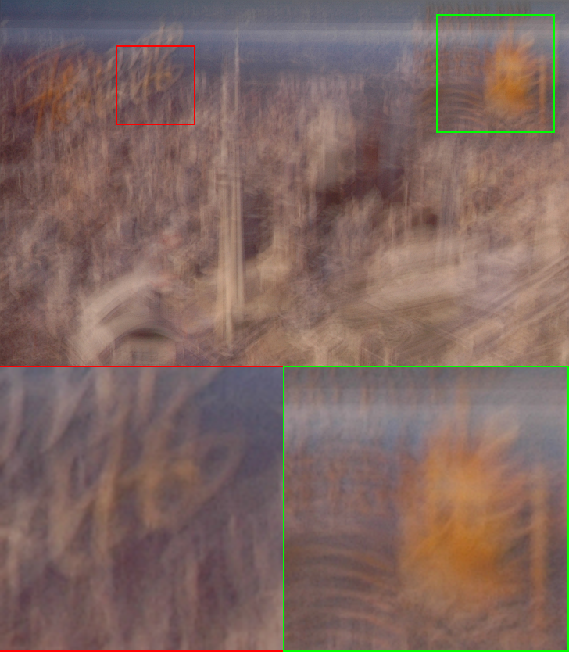}}
		\subfigure[CCK]{\includegraphics*[width = 0.137\linewidth]{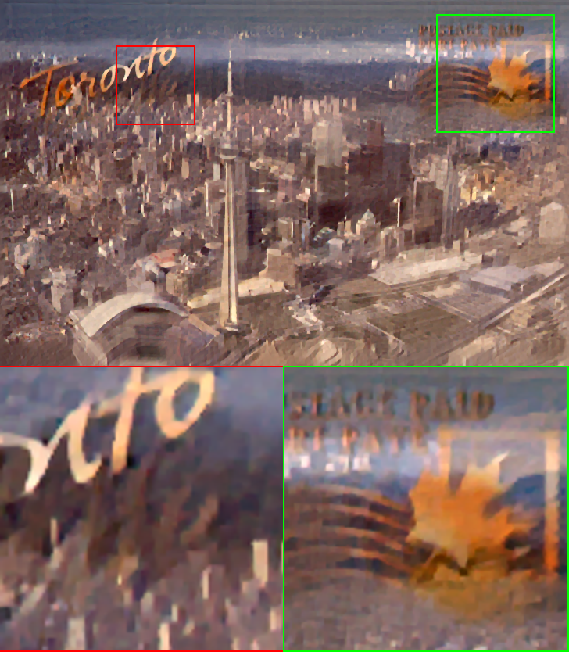}}
		\subfigure[QCK]{\includegraphics*[width = 0.137\linewidth]{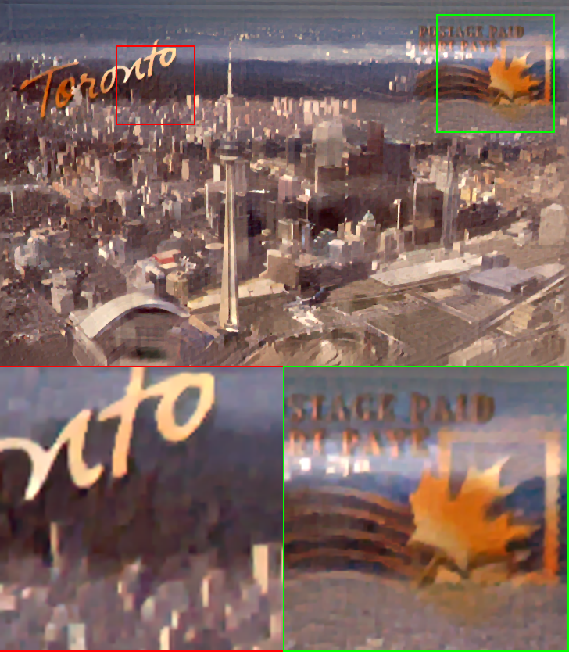}}
		\subfigure[dark-channel\cite{DCP2018}]{\includegraphics*[width = 0.137\linewidth]{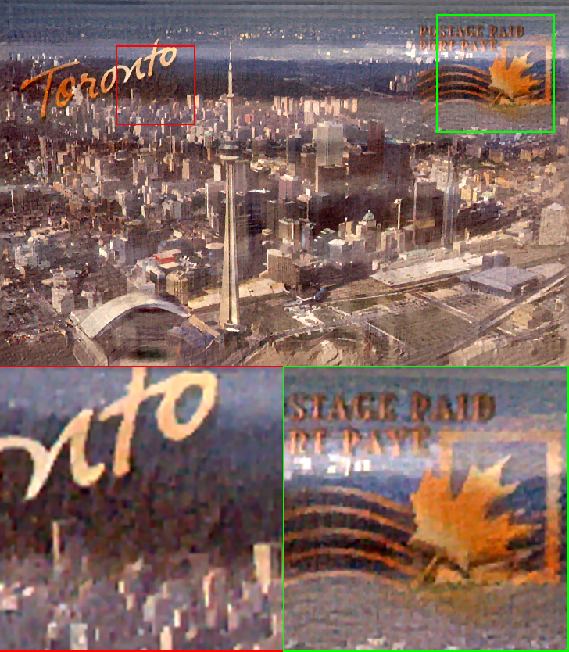}}
		\subfigure[{\scriptsize dark-channel+QCK}]{\includegraphics*[width = 0.137\linewidth]{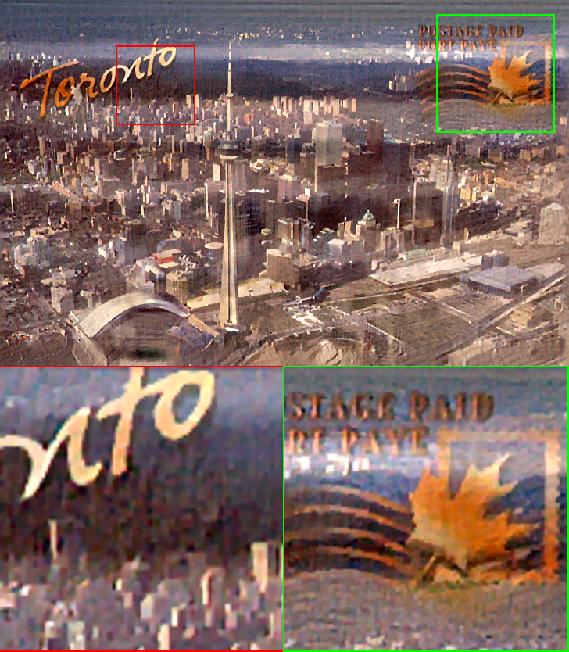}}
		\subfigure[{\scriptsize surface-aware\cite{SA2021}}]{\includegraphics*[width = 0.137\linewidth]{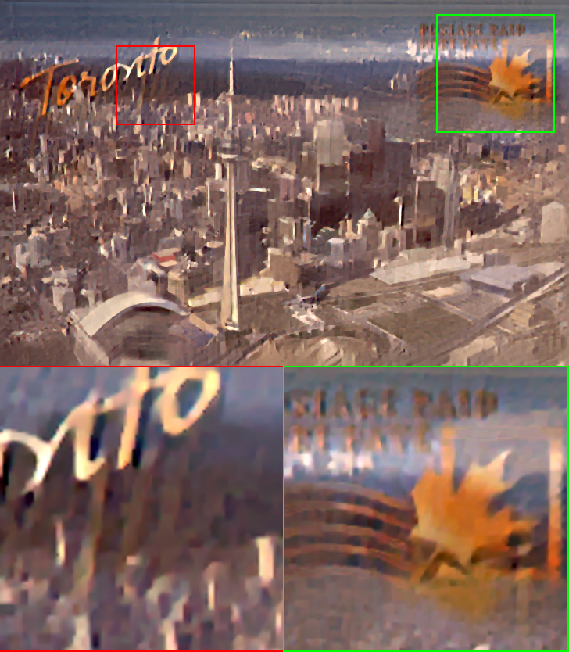}}
		\subfigure[{\scriptsize surface-aware+QCK}]{\includegraphics*[width = 0.137\linewidth]{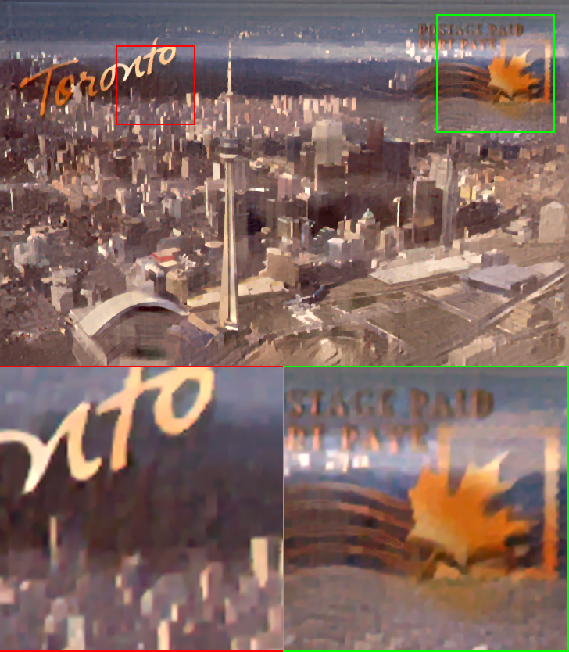}}
		\caption{
			(a) the blurred image; 
			(b) the restored result by using CCK;
			(c) the restored result by using QCK;
					       (d) the restored result by using dark-channel prior;
					       (e) the restored result by using dark-channel prior+QCK;			       					       
			(f) the restored result by using surface-aware prior;
			(g) the restored result by using surface-aware prior+QCK.}
		\label{fig-nonblind}
	\end{figure*}
	
	\begin{figure*}[!t]
		\centering
		\includegraphics*[width = 0.137\linewidth]{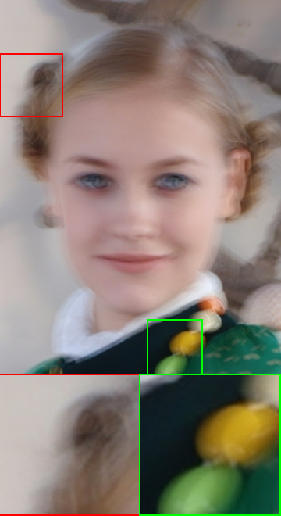}
		\includegraphics*[width = 0.137\linewidth]{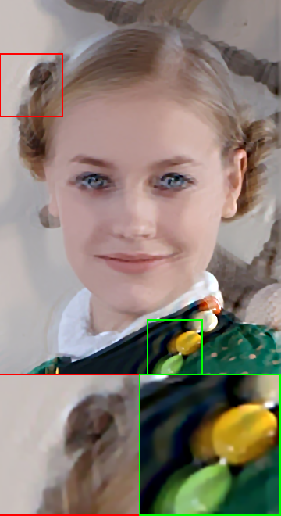}
		\includegraphics*[width = 0.137\linewidth]{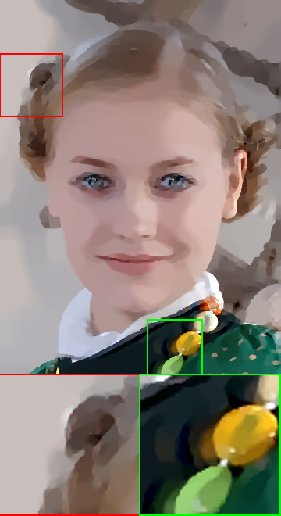}
		\includegraphics*[width = 0.137\linewidth]{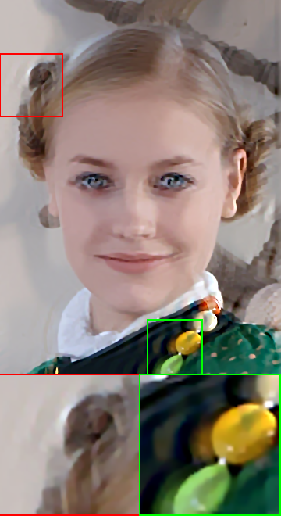}
		\includegraphics*[width = 0.137\linewidth]{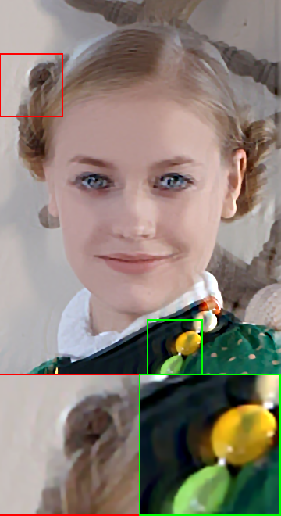}
		\includegraphics*[width = 0.137\linewidth]{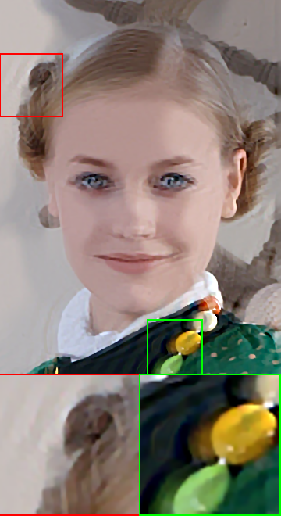}
		\includegraphics*[width = 0.137\linewidth]{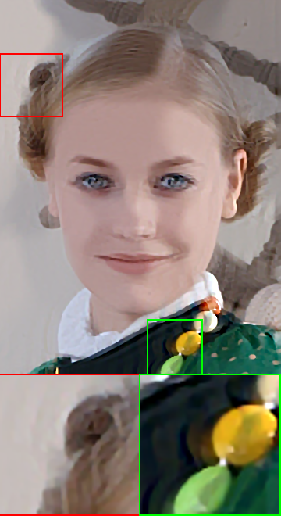}
		\\
		\subfigure[blurred image]{\includegraphics*[width = 0.137\linewidth]{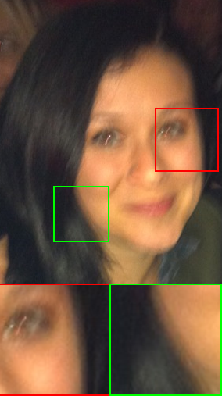}}
		\subfigure[CCK]{\includegraphics*[width = 0.137\linewidth]{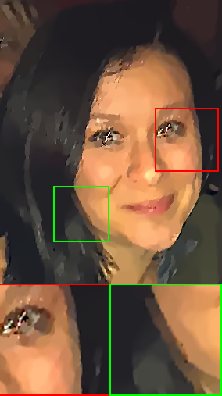}}
		\subfigure[QCK]{\includegraphics*[width = 0.137\linewidth]{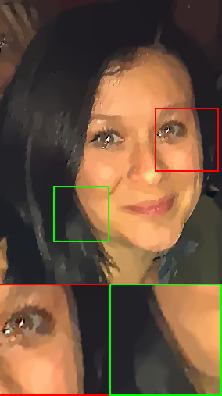}}
		\subfigure[dark-channel\cite{DCP2018}]{\includegraphics*[width = 0.137\linewidth]{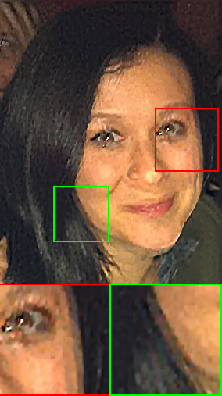}}
		\subfigure[{\scriptsize dark-channel+QCK}]{\includegraphics*[width = 0.137\linewidth]{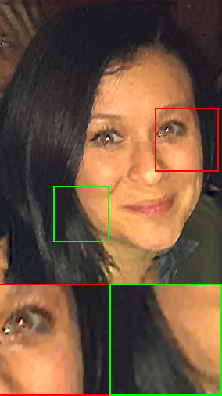}}
		\subfigure[{\scriptsize surface-aware\cite{SA2021}}]{\includegraphics*[width = 0.137\linewidth]{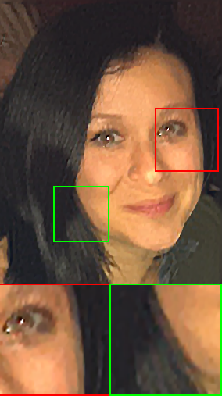}}
		\subfigure[{\scriptsize surface-aware+QCK}]{\includegraphics*[width = 0.137\linewidth]{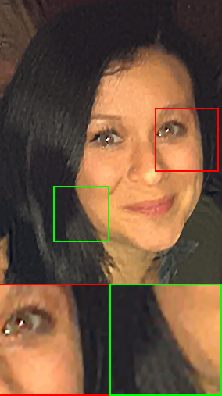}}
		\caption{(a) the blurred image; 
			(b) the restored result by using CCK;
			(c) the restored result by using QCK;
					       (d) the restored result by using dark-channel prior;
					       (e) the restored result by using dark-channel prior+QCK;			       					       
			(f) the restored result by using surface-aware prior;
			(g) the restored result by using surface-aware prior+QCK.}
		\label{fig-face}
	\end{figure*}
	\begin{figure*}[t]
		\centering
		\subfigure[blurred image]{\includegraphics*[width = 0.137\linewidth]{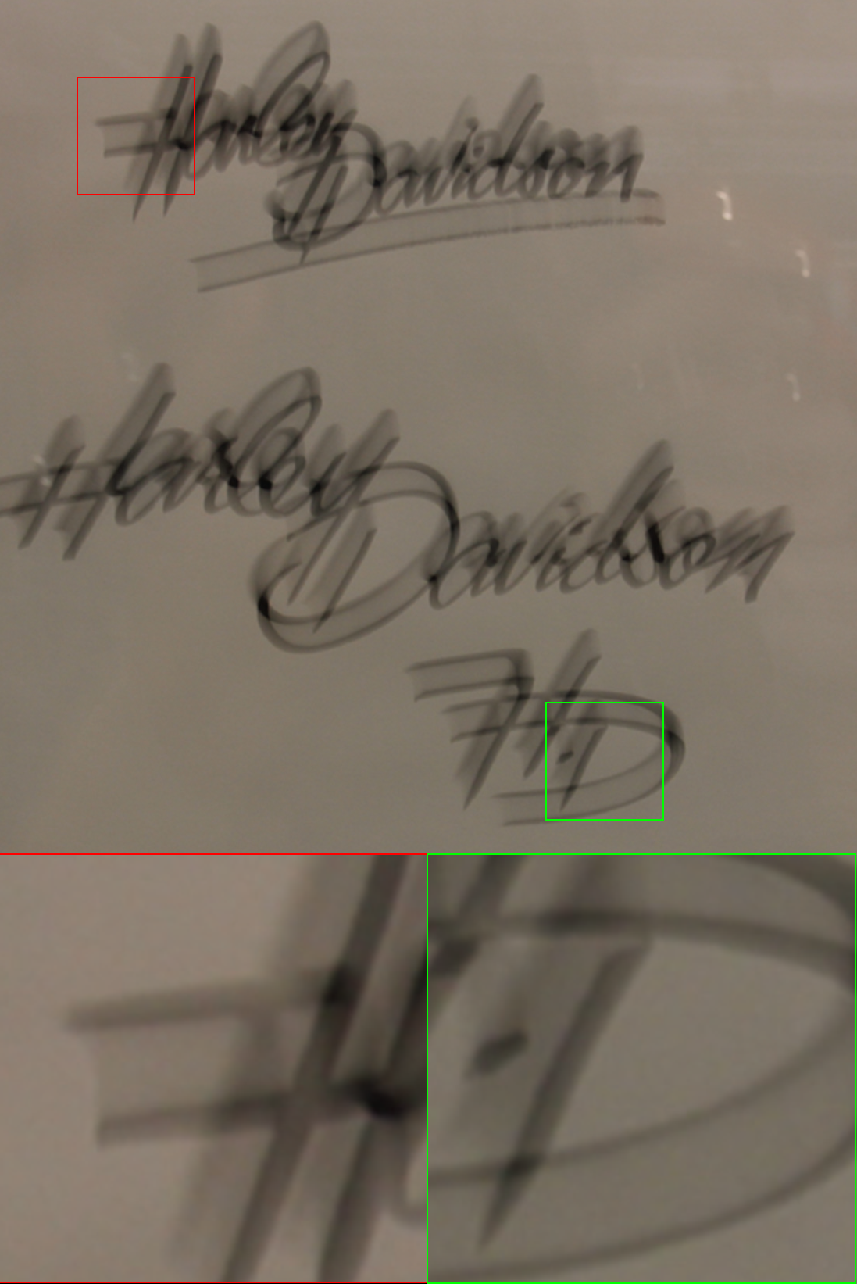}}
		\subfigure[CCK]{\includegraphics*[width = 0.137\linewidth]{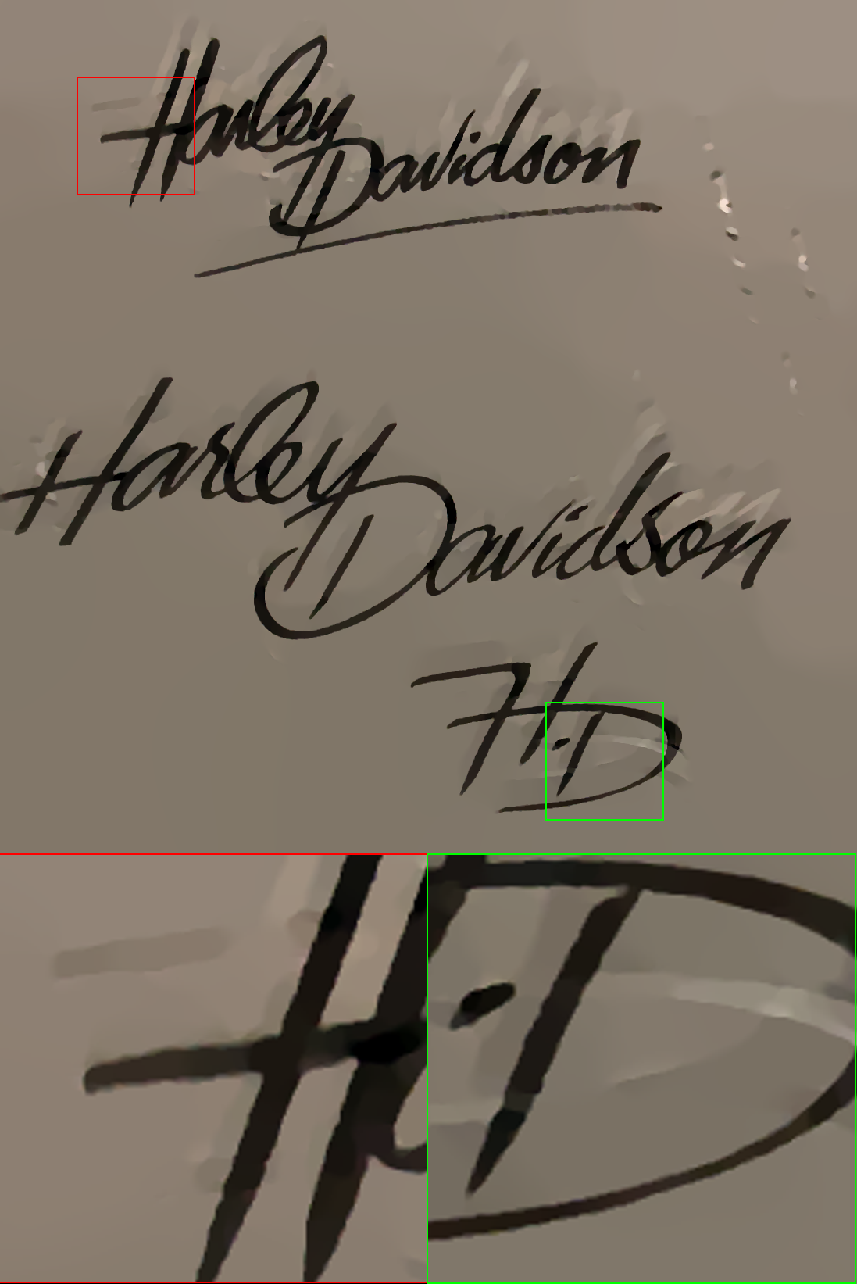}}
		\subfigure[QCK]{\includegraphics*[width = 0.137\linewidth]{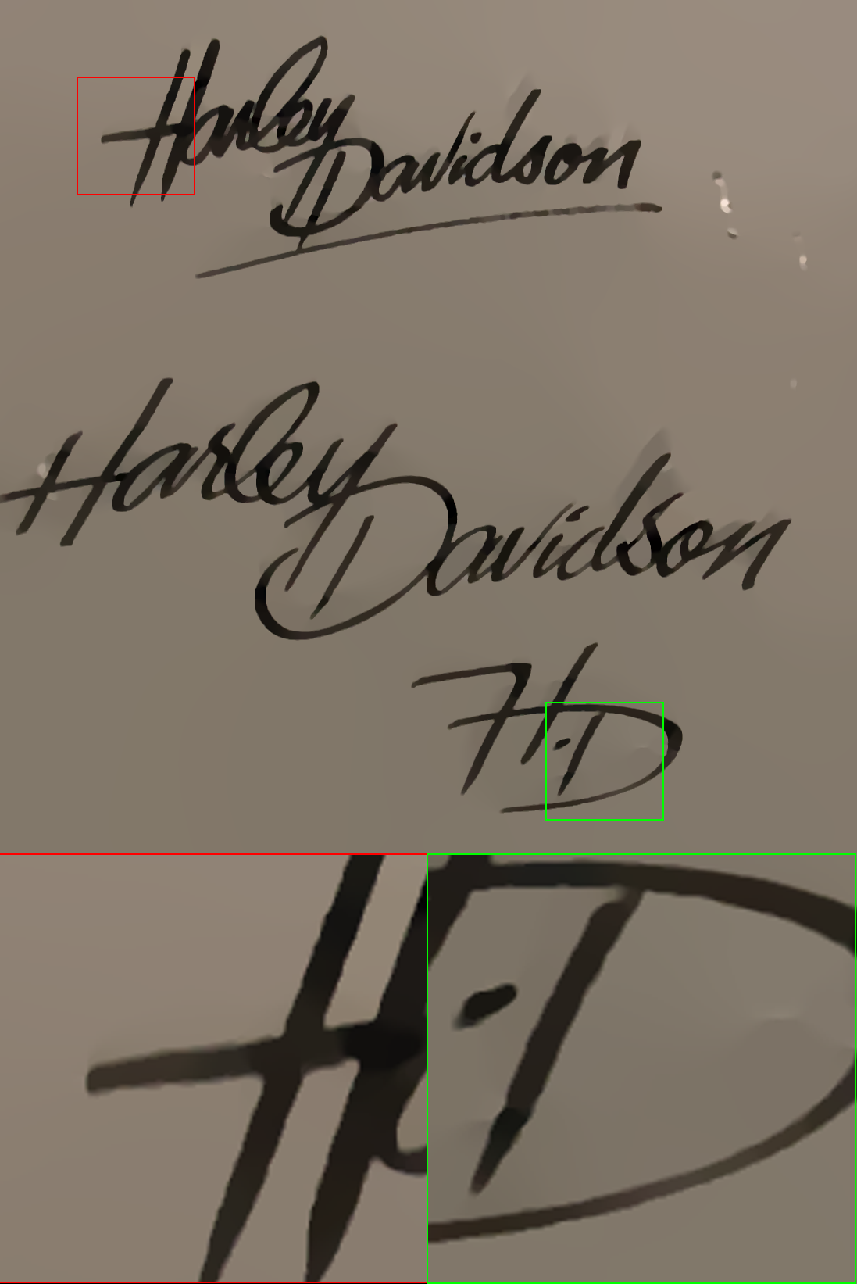}}
		\subfigure[dark-channel\cite{DCP2018}]{\includegraphics*[width = 0.137\linewidth]{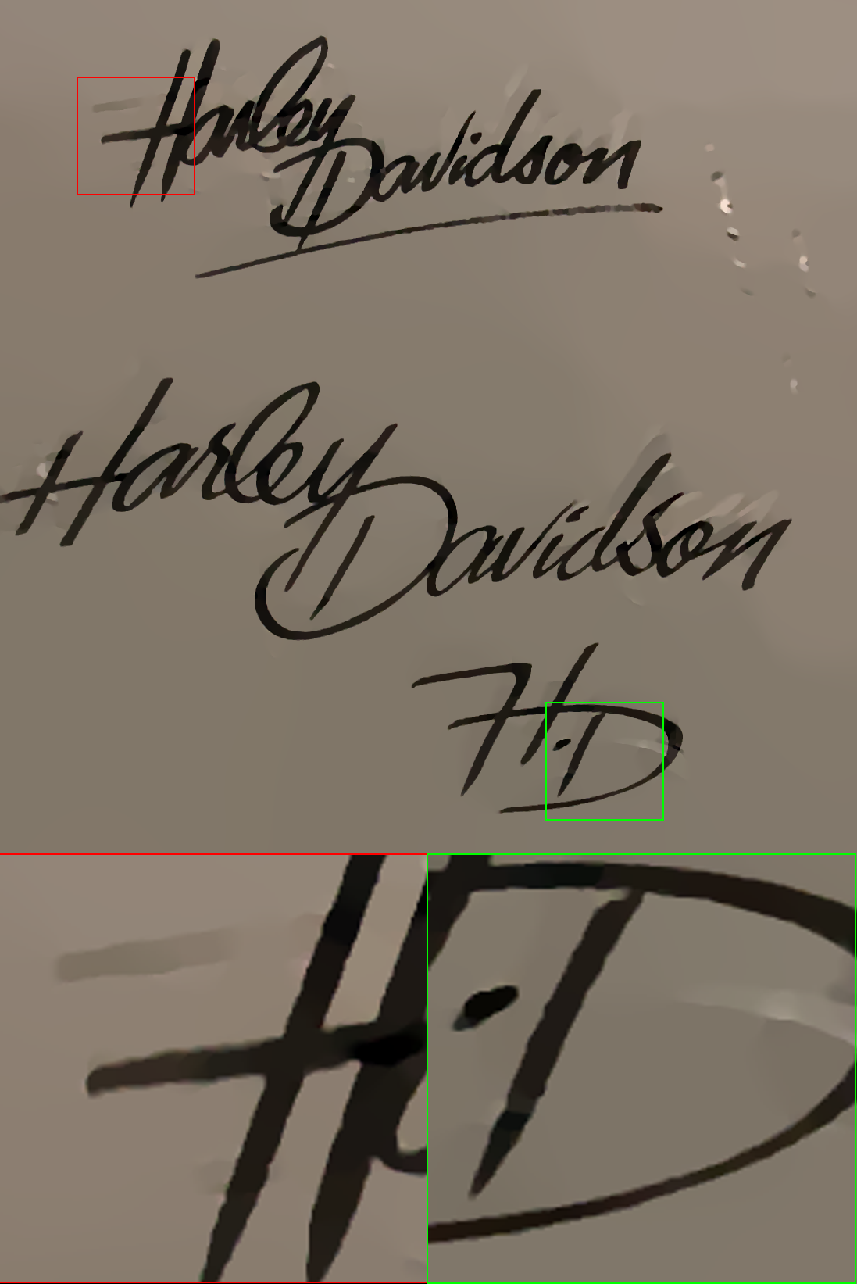}}
		\subfigure[{\scriptsize dark-channel+QCK}]{\includegraphics*[width = 0.137\linewidth]{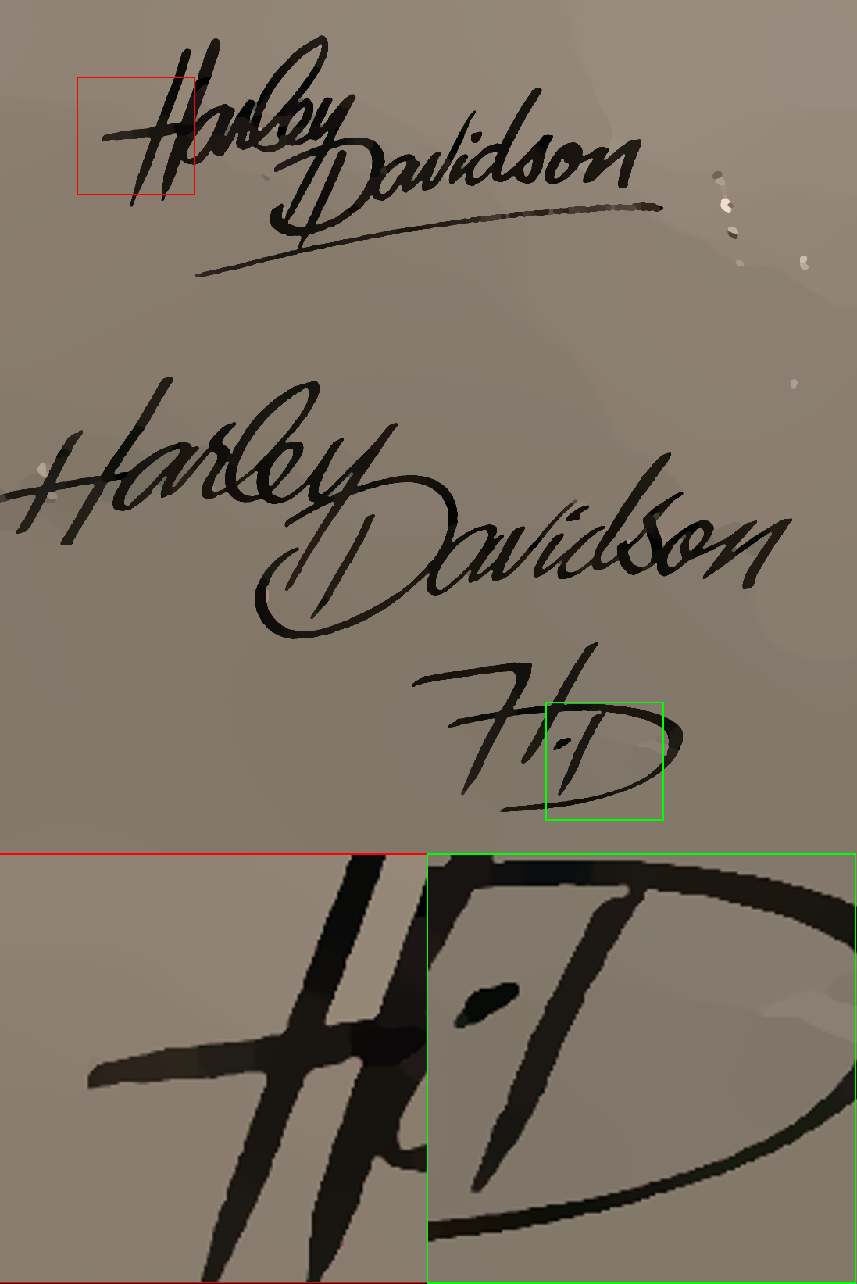}}
		\subfigure[{\scriptsize surface-aware\cite{SA2021}}]{\includegraphics*[width = 0.137\linewidth]{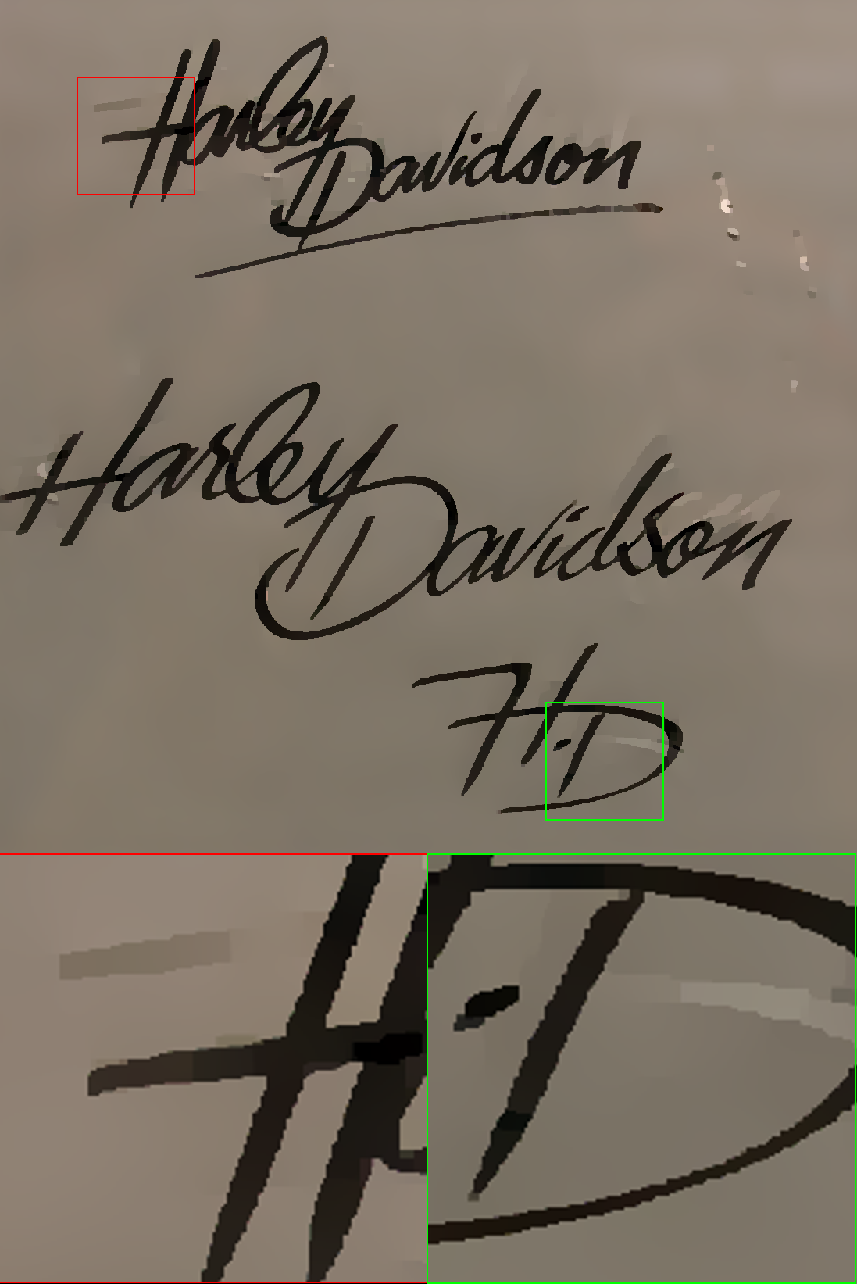}}
		\subfigure[{\scriptsize surface-aware+QCK}]{\includegraphics*[width = 0.137\linewidth]{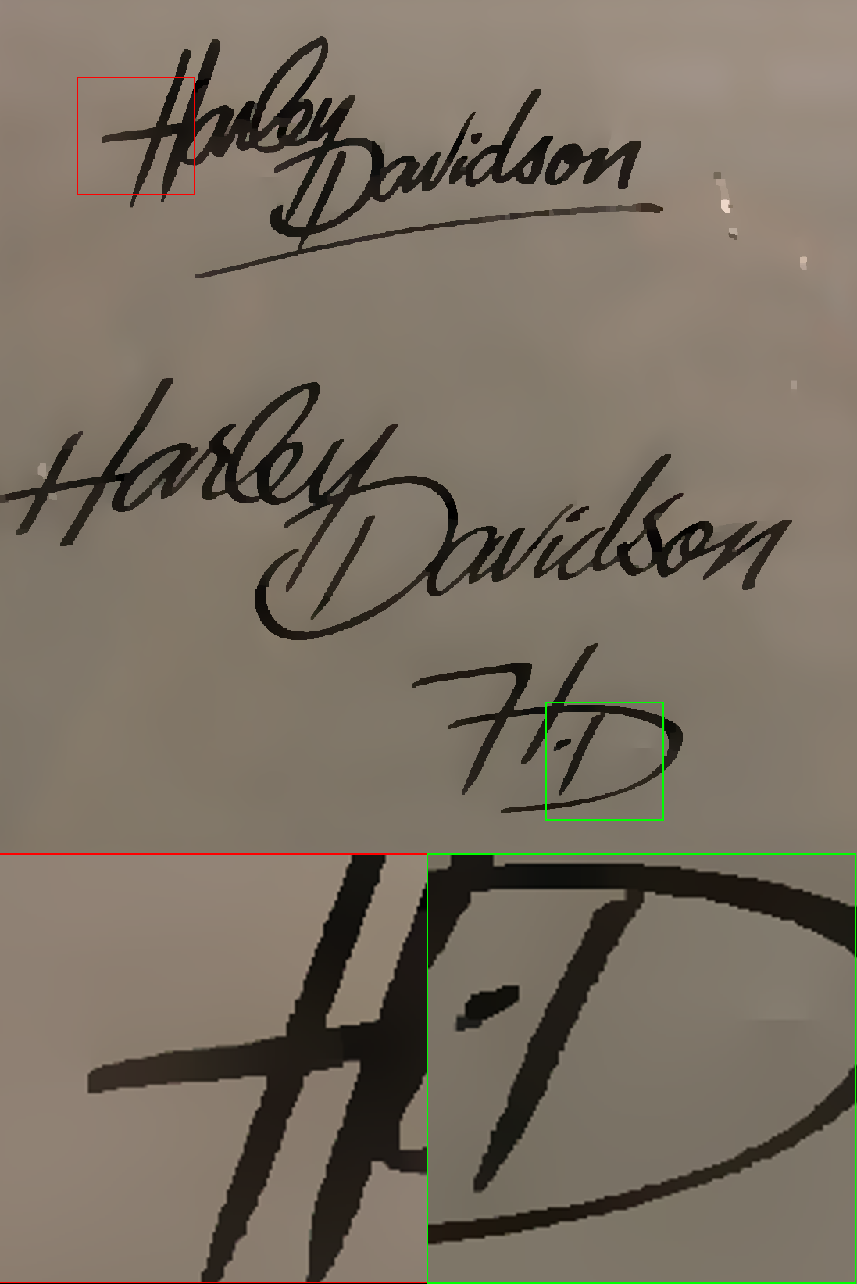}}
		\caption{(a) the blurred image; 
			(b) the restored result by using CCK;
			(c) the restored result by using QCK;
					       (d) the restored result by using dark-channel prior;
					       (e) the restored result by using dark-channel prior+QCK;			       					       
			(f) the restored result by using surface-aware prior;
			(g) the restored result by using surface-aware prior+QCK.}
		\label{fig-text}
	\end{figure*}
	
	In this subsection, we further test the proposed quaternion deconvolution model on real blurred image (without the underlying 
	sharp image) from the project in \cite{DCP2018} by using different methods. We show the blind deconvolution results in the Figures \ref{fig-blind}-\ref{fig-text} and evaluate the results visually. 
	
	First, we compare the latent images obtained directly from the blind-deconvolution models. 
	As shown in Figure \ref{fig-blind}, the proposed model with normalized quaternion convolution kernel demonstrates superior performance in artifacts reduction in complex natural scenes. 
	Compared with the blind deconvolution results obtained by traditional methods, the proposed quaternion model reveals more subtle textures and structures.
	In the first image of Figure \ref{fig-blind}, we focus on the lower edge of the label. The proposed quaternion deconvolution model significantly suppresses artifacts in this region. 
	The artifacts appear more regular by using dark channel + QCK, and are almost completely eliminated by using surface-aware + QCK.	
	Similarly, In the second image of Figure \ref{fig-blind}, we see that the proposed quaternion deconvolution model significantly reduces artifacts in the plant branches, demonstrating its effectiveness.	

	In Figure \ref{fig-nonblind}, we show the restored results produced by feeding the kernels obtained from the blind deconvolution model into the non-blind deconvolution model as discussed in Section V. B. We remark that the non-blind models use the same parameters by using the same image prior.
	The first image in Figure \ref{fig-nonblind} demonstrates that the proposed quaternion deconvolution model is effective in suppressing artifacts even in the presence of high saturation, while the second image highlights its advantage in handling complex blur.
	
	Furthermore, the proposed quaternion deconvolution model also works well on human face images. 
	As shown in Figure \ref{fig-face}, the proposed quaternion deconvolution model effectively eliminates artifacts, such as those in the chin and eye areas.
	In addition, we evaluate the effectiveness of the proposed quaternion deconvolution model for deblurring text images.
	In Figure \ref{fig-text}, we present the text deblurring results by using different methods, showing that the quaternion deconvolution model significantly removes artifacts and sharpens the text compared to traditional methods.		 
	
	In summary, the proposed normalized quaternion convolution kernel plays an important role in real-world image deconvolution problem: it consistently produces results with fewer artifacts and significantly better visual quality in deblurring natural scenes, faces, and text images.

	\section{Conclusion}
	
	In this paper, we have proposed a novel quaternion deconvolution model specifically designed for color image deblurring. 
	Based on a quaternion convolution kernel which consists of four components: one analogous to a single real blur kernel and three others to account for the interdependencies among RGB channels, we formulate a novel quaternion data fidelity term applied to color image blind deconvolution.
	In order to preserve image intensity, we propose to use the normalized quaternion convolution kernel in the blind deconvolution process. 
	Extensive experiments on synthetic and real-world datasets demonstrate the effectiveness of the proposed quaternion deconvolution model in reducing artifacts and the ability to achieve impressive results as a powerful tool for color image deconvolution.
	We emphasize that the proposed quaternion convolution kernel can efficiently combine or generalize other methods with various priors to improve the deblurring effect.

	In future work, we will focus on the quaternion regularization technique for color image processing. We note that most of current regularization approaches still treat three color channels as statistically equivalent, which ignores the complex interactions and correlations between color channels.
	By introducing a quaternion-based regularization term, we expect to achieve more balanced and accurate results, even when details in some channels are less pronounced or partially missing.

	\bibliographystyle{IEEEtran}
	\bibliography{REF_Q.bib}

\end{document}